\PassOptionsToPackage{shortlabels}{enumitem}
\documentclass[11pt, a4paper]{include/gdm_format}
\usepackage[authoryear, sort&compress, round]{natbib}
\usepackage{xcolor}
\usepackage{subcaption}
\usepackage{graphicx}
\usepackage{xr}
\usepackage{marvosym}
\usepackage{makecell}
\usepackage{fvextra}
\usepackage{tablefootnote}
\usepackage{pdfpages}
\usepackage{float}
\usepackage{etoc}
\usepackage{hyperref}
\usepackage{url}
\usepackage{xurl}
\usepackage[parfill]{parskip}
\usepackage{amsmath}
\usepackage{xspace}
\usepackage{multirow}
\usepackage{paracol}
\usepackage{booktabs}
\usepackage{threeparttable}
\usepackage{color}
\usepackage{todonotes}
\usepackage{colortbl}
\usepackage{cleveref}
\usepackage{algorithm}
\usepackage{algorithmicx}
\usepackage{algpseudocode}
\usepackage{wrapfig}
\usepackage{sidecap}
\usepackage{soul}
\usepackage{multicol}
\usepackage{pifont}
\usepackage[normalem]{ulem}
\usepackage{listings}
\usepackage{changes}
\usepackage{titletoc}
\sidecaptionvpos{figure}{t}
\usepackage{adjustbox}
\usepackage{pifont}
\usepackage{tabularx}
\usepackage{siunitx}
\usepackage{lipsum}   % in the preamble

\DeclareUnicodeCharacter{2014}{---}
\DeclareUnicodeCharacter{2192}{\ensuremath{\rightarrow}}
\DeclareUnicodeCharacter{2264}{\ensuremath{\leq}}
\DeclareUnicodeCharacter{26A0}{\textbf{WARNING}}
\DeclareUnicodeCharacter{03C4}{\ensuremath{\tau}}
\DeclareUnicodeCharacter{223C}{\ensuremath{\sim}}

\usepackage[most]{tcolorbox}

\tcbset{
  base/.style={
    arc=0mm,
    bottomtitle=0.5mm,
    boxrule=0mm,
    colbacktitle=black!10!white,
    coltitle=black,
    fonttitle=\bfseries,
    left=2.5mm,
    leftrule=1mm,
    right=3.5mm,
    title={#1},
    toptitle=0.75mm,
  },
}

\newcommand{\appendixfileinput}[1]{%
  \VerbatimInput[
    fontsize={\fontsize{7pt}{8pt}\selectfont},
    breaklines=true,
    breakanywhere=true,
    breaksymbolleft={},
    breaksymbolright={}
  ]{#1}%
}

\newtcolorbox{promptbox}[1]{
  base={#1},
  breakable,
  colback=violet!5!white,
  colframe=violet!75!black,
}

\newtcolorbox{taskinstructionsbox}[1]{
  base={Initial Task Instructions for #1 Topic Direction},
  breakable,
  colback=blue!5!white,
  colframe=blue!75!black,
}

\newtcolorbox{reviewbox}{
  base={Review},
  breakable,
  colback=green!5!white,
  colframe=green!75!black,
}

\usepackage{pgffor}
\tcbuselibrary{skins}
\newcount\embedpdfpagecount
\newlength{\embedpdfmaxheight}
\newlength{\embedpdfavail}
\newcommand{\embedpdfframecolor}{black!35}
\newcommand{\embedpdfslack}{3\baselineskip+18pt}

\newcommand{\embedpdfpage}[3]{%
	\clearpage
	\setlength{\embedpdfmaxheight}{\dimexpr\textheight-(\embedpdfslack)\relax}%
	\embedpdfplace{#1}{#2}{#3}%
}

\newcommand{\embedpdffirstpage}[3]{%
	\par\penalty0\relax
	\setlength{\embedpdfmaxheight}{\dimexpr\textheight-(\embedpdfslack)\relax}%
	\ifdim\pagegoal<\maxdimen
		\setlength{\embedpdfavail}{\dimexpr\pagegoal-\pagetotal-(\embedpdfslack)\relax}%
		\ifdim\embedpdfavail<\embedpdfmaxheight
			\ifdim\embedpdfavail>0.6\textheight
				\setlength{\embedpdfmaxheight}{\embedpdfavail}%
			\else
				\clearpage
			\fi
		\fi
	\fi
	\embedpdfplace{#1}{#2}{#3}%
}

\newcommand{\embedpdfplace}[3]{%
	\begingroup
	\centering
	\begin{tcolorbox}[colback=white, colframe=\embedpdfframecolor,
			boxrule=0.4pt, arc=3pt, boxsep=0pt, width=\dimexpr\linewidth-5pt\relax,
			left=2pt, right=2pt, top=2pt, bottom=2pt, halign=center,
			drop fuzzy shadow=black!45]
		\includegraphics[page=#2, width=\linewidth,
			height=\embedpdfmaxheight, keepaspectratio]{#1}%
	\end{tcolorbox}
	\par\vspace{4pt}
	{\footnotesize #3\par}%
	\endgroup
}

\newcommand{\embedpdfpaper}[2]{%
	\pdfximage{#1}%
	\embedpdfpagecount=\pdflastximagepages\relax
	\foreach \embedpdfp in {1,...,\the\embedpdfpagecount}{%
		\ifnum\embedpdfp=1\relax
			\embedpdffirstpage{#1}{\embedpdfp}%
			{\itshape #2\ \upshape(page~\embedpdfp\ of~\the\embedpdfpagecount)}%
		\else
			\embedpdfpage{#1}{\embedpdfp}%
			{\itshape #2\ \upshape(page~\embedpdfp\ of~\the\embedpdfpagecount)}%
		\fi
	}%
	\clearpage
}

\newlength{\embedpdfcellwidth}
\newlength{\embedpdfcellheight}

\newcommand{\embedpdfcell}[2]{%
	\begin{minipage}[c]{\embedpdfcellwidth}%
		\centering
		\ifnum#2=0\relax
			\rule{0pt}{0pt}%
		\else
			\begin{tcolorbox}[enhanced, colback=white, colframe=\embedpdfframecolor,
					boxrule=0pt, arc=0pt, boxsep=0pt, width=\linewidth,
					left=1pt, right=1pt, top=1pt, bottom=1pt, halign=center,
					borderline={0.4pt}{0pt}{densely dotted, \embedpdfframecolor}]
				\includegraphics[page=#2, width=\linewidth,
					height=\embedpdfcellheight, keepaspectratio]{#1}%
			\end{tcolorbox}%
		\fi
	\end{minipage}%
}

\newcommand{\embedpdffourbody}[6]{%
	\begingroup
	\centering
	\setlength{\lineskip}{0pt}%
	\embedpdfcell{#1}{#2}\embedpdfcell{#1}{#3}\par
	\embedpdfcell{#1}{#4}\embedpdfcell{#1}{#5}\par
	\vspace{6pt}
	{\footnotesize #6\par}%
	\endgroup
}

\newcommand{\embedpdffourplace}[6]{%
	\clearpage
	\setlength{\embedpdfcellwidth}{0.5\linewidth}%
	\setlength{\embedpdfcellheight}{\dimexpr0.5\textheight-1.5\baselineskip\relax}%
	\embedpdffourbody{#1}{#2}{#3}{#4}{#5}{#6}%
}

\newcommand{\embedpdffourfirstplace}[6]{%
	\par\penalty0\relax
	\setlength{\embedpdfcellwidth}{0.5\linewidth}%
	\setlength{\embedpdfcellheight}{\dimexpr0.5\textheight-1.5\baselineskip\relax}%
	\ifdim\pagegoal<\maxdimen
		\setlength{\embedpdfavail}{\dimexpr\pagegoal-\pagetotal\relax}%
		\ifdim\embedpdfavail<\textheight
			\ifdim\embedpdfavail>0.6\textheight
				\setlength{\embedpdfcellheight}{\dimexpr0.5\embedpdfavail-1.5\baselineskip\relax}%
			\else
				\clearpage
			\fi
		\fi
	\fi
	\embedpdffourbody{#1}{#2}{#3}{#4}{#5}{#6}%
}

\newcommand{\embedpdffourchunk}[5]{%
	\edef\embedpdfpageA{\the\numexpr#3\relax}%
	\edef\embedpdfpageB{\the\numexpr#3+1\relax}%
	\edef\embedpdfpageC{\the\numexpr#3+2\relax}%
	\edef\embedpdfpageD{\the\numexpr#3+3\relax}%
	\edef\embedpdfpageEnd{\the\numexpr#3+3\relax}%
	\ifnum\embedpdfpageEnd>#4\relax \edef\embedpdfpageEnd{\the\numexpr#4\relax}\fi
	\ifnum\embedpdfpageB>#4\relax \def\embedpdfpageB{0}\fi
	\ifnum\embedpdfpageC>#4\relax \def\embedpdfpageC{0}\fi
	\ifnum\embedpdfpageD>#4\relax \def\embedpdfpageD{0}\fi
	\ifnum\embedpdfpageA=\embedpdfpageEnd\relax
		\def\embedpdffourcap{\itshape #5\ \upshape(page~\embedpdfpageA\ of~#4)}%
	\else
		\def\embedpdffourcap{\itshape #5\ \upshape(pages~\embedpdfpageA--\embedpdfpageEnd\ of~#4)}%
	\fi
	#1{#2}{\embedpdfpageA}{\embedpdfpageB}{\embedpdfpageC}{\embedpdfpageD}{\embedpdffourcap}%
}

\newcount\embedpdffours
\newcommand{\embedpdfpaperfour}[2]{%
	\pdfximage{#1}%
	\embedpdfpagecount=\pdflastximagepages\relax
	\embedpdffours=1\relax
	\loop
		\ifnum\embedpdffours=1\relax
			\embedpdffourchunk{\embedpdffourfirstplace}{#1}{\the\embedpdffours}{\the\embedpdfpagecount}{#2}%
		\else
			\embedpdffourchunk{\embedpdffourplace}{#1}{\the\embedpdffours}{\the\embedpdfpagecount}{#2}%
		\fi
		\advance\embedpdffours by 4\relax
	\ifnum\embedpdffours<\numexpr\embedpdfpagecount+1\relax\repeat
	\clearpage
}

\newcommand{\cmark}{\ding{51}} % checkmark
\newcommand{\xmark}{\ding{55}} % cross

\newif\ifcomments
\commentsfalse % replace with \commentstrue to enable all comments

\newif\ifcommenterrors
\commenterrorsfalse % replace with \commenterrorstrue to raise an error for every unresolved comment

\title{The Agentic Scientist: Orchestrating Coding Agents for Automated Scientific (Co-)Discovery}

\correspondingauthor{Robert Tjarko Lange (robert@sakana.ai). Full author contributions list in \Cref{appsec:contribs}.}

\author[1,*]{Robert Tjarko Lange}
\author[1,*]{Luca Grillotti}
\author[1,*]{Yutaro Yamada}
\author[1]{Rujikorn Charakorn}
\author[1]{David Ha}

\affil[1]{Sakana AI}
\affil[*]{Equal Contribution.}

\newcommand{\agentscientist}{\textsc{The Agentic Scientist}\xspace}
\newcommand{\agentscientistnothe}{\textsc{Agentic Scientist}\xspace}
\newcommand{\aiscientistfirst}{\textsc{AI Scientist-v1}\xspace}
\newcommand{\aiscientistsecond}{\textsc{AI Scientist-v2}\xspace}
\newcommand{\autoillustrate}{\textsc{autoillustrate}\xspace}

\begin{document}

\vspace{-5em}
\begin{abstract}
Modern coding agents are fundamentally reshaping the role of human input for computational research: they write code, launch in silico experiments, analyze their results, and autonomously operate with ever-increasing time horizons. 
Despite these developments, fully autonomous research agents suffer from a limited `ideation radius' and myopic hill-climbing search dynamics.
To bridge this gap, we introduce \agentscientist, an end-to-end distributed pipeline for human researchers to conduct computational research alongside modern coding agents.
Building on its predecessor \citep[\textsc{The AI Scientist},][]{lu2026towards}, our system decomposes research into task drafting, idea generation, planning, sandboxed experimentation, result aggregation, baseline and ablation construction, post-hoc analysis, and report writing.
Importantly, each step in the pipeline is powered by the next generation of headless coding agents and leverages agentic self-verification.
The overall research process is driven by a meta-controller that manages parallel research agents and enforces rubric verification at each stage.
This design enables the same orchestration layer to be reused across heterogeneous domains with different execution environments, agentic coding harnesses, and evaluation metrics. Our system can be used across a spectrum of autonomy: either as a co-scientist tool, or to perform end-to-end scientific investigation.
Within the co-scientist setting, we leveraged a development version of our system to submit a single research paper, which got accepted for publication at ALIFE 2026 \citep{yamada2026diffusing}.
We further apply \agentscientist fully autonomously to open problem-solving tasks in machine learning, and demonstrate the system's capability of scientific discovery across different domains of computational research, including: a hypernetwork-based inductive bias for rapid Transformer adaptation and a meta-trained evolution strategy that matches or exceeds CMA-ES on 5 of 7 unseen Brax control tasks.
Our ablations suggest that reliable scientific autonomy depends not only on base model capabilities, but also on the cross-agent orchestration layer.
We open-source \agentscientist and its figure generation tool, \autoillustrate, to aid collective development of automated (co-)scientist systems:

\vspace{0.5em}
\begin{center}
\begin{tabular}{rcl}
\raisebox{-1.5pt}{\includegraphics[height=1.05em]{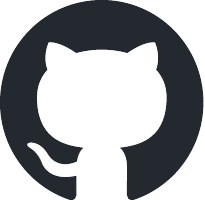}} & \textsc{Agentic Scientist}\xspace & \href{https://github.com/SakanaAI/AI-Scientist-v3}{\path{https://github.com/SakanaAI/AI-Scientist-v3}} \\
\raisebox{-1.5pt}{\includegraphics[height=1.05em]{figures/github-logo.pdf}} & \autoillustrate & \href{https://github.com/SakanaAI/autoillustrate}{\path{https://github.com/SakanaAI/Autoillustrate}} \\
\end{tabular}
\end{center}

\end{abstract}

\maketitle
% \vspace{-0.35cm}
% \setcounter{tocdepth}{1}
% \etocdepthtag.toc{mtchapter}
% \etocsettagdepth{mtchapter}{subsection}
% \etocsettagdepth{mtappendix}{none}
% \tableofcontents
% \clearpage
\vspace{-0.5em}
\begin{figure}[H]
    \centering
    \includegraphics[width=0.875\linewidth]{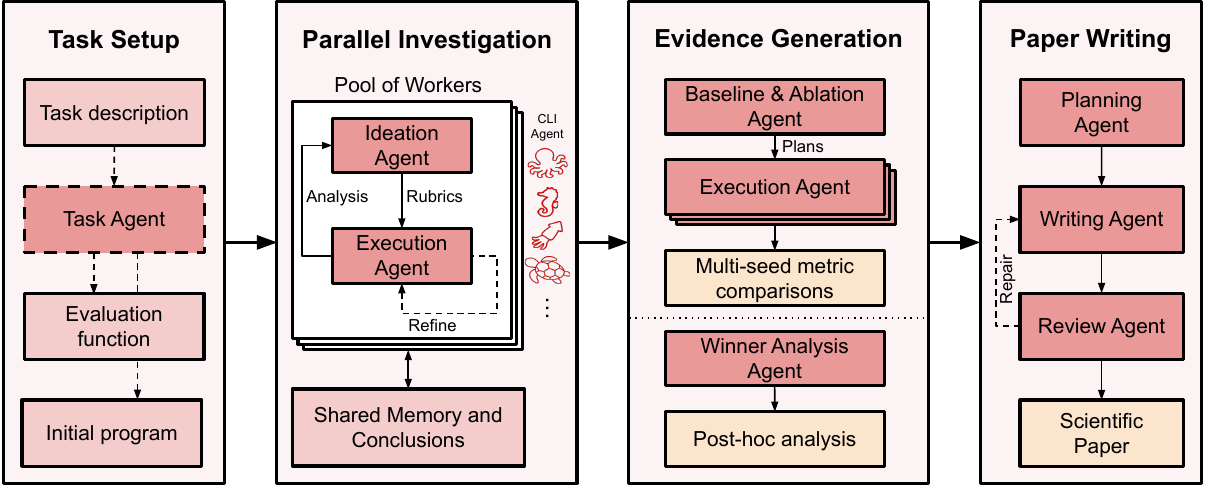}
    \caption{\agentscientist Four Stage Workflow.}
    \label{fig:conceptual}
\end{figure}

\newpage
\section{Introduction}
Large Language Model (LLM)-based tool-calling agents can now complete increasingly long coding tasks with minimal human intervention~\citep{kwa2025horizon}.
Computational research, however, is arguably more than a single long coding task: a practical system supporting human scientists must generate hypotheses, execute controlled experiments, validate claims, iterate on ideas, and communicate the results.
Existing AI scientist systems demonstrate parts of this loop~\citep{lu2024ai,yamada2025ai,schmidgall-etal-2025-agent,tang2025airesearcher,jansen2025codescientist,karpathy2026autoresearch}, but myopic hill-climbing can make agents behave merely as flexible hyperparameter optimization engines \citep{ferreira2026can, van2024loop}.

Here, we tackle the systems-level gap between an agent that writes code and a system that can conduct reliable research.
We introduce \agentscientist (\Cref{fig:conceptual}), a human-steerable, agentic scaffold for end-to-end computational research.
The workflow drafts a task specification from a text description, coordinates parallel sandboxed workers, accumulates retrospective evidence, and writes a manuscript grounded in the results.
During the investigation phase, parallel workers revise scientific hypotheses in response to implementation failures, measured outcomes, and shared performance notes.
The resulting artifacts seed subsequent experiments and remain available for retrospective analysis, ablation design, and manuscript writing.
Predecessors of our system (\Cref{tab:version_diff}) relied on initial implementation directions and loose assessment of the idea implementation. This could result in weak experiment validation and overfitting. \agentscientist, on the other hand, uses self-drafted rubrics-based verification and CLI agents for broader code edits and information protocols.

\begin{table}[htbp]
    \centering
    \caption{\textbf{Comparison of Sakana AI's AI Scientist Frameworks.} 
\agentscientist enhances previous versions with rubric-based verification and the option for co-scientist enhancement.}
    \begin{adjustbox}{width=\textwidth}
    \begin{tabular}{lcccccc}
        \toprule
\textit{Feature} & \textbf{Codebase} & \textbf{Execution} & \textbf{Parallel} & \textbf{Rubric-based} & \textbf{Human Evaluation} \\
        & \textbf{Drafting} & \textbf{Workflow} & \textbf{Experiments} & \textbf{Self-Verification}  & \textbf{Result} \\
        \midrule
        \aiscientistfirst & Topic-Specific & Linear (Aider) & \xmark & \xmark & Not Submitted\\
        \aiscientistsecond & Domain-General & Tree-Based (AIDE) & \cmark & \xmark & Conference Workshop Acceptance\\
        \rowcolor{gray!20}
        \textbf{\textsc{Agentic Scientist}} & Domain-General & Beam Search (CLI Agents) & \cmark & \cmark & Co-Scientist Conference Acceptance \\
        \bottomrule
    \end{tabular}
    \end{adjustbox}
    \label{tab:version_diff}
\end{table}

A development version of our system was used to automate parts of the research process on Error Diffusion~\citep{fujita2026diagnostic,kaneko2000error_diffusion}. Afterwards, we submitted a single paper which was accepted as an ALIFE 2026 talk \citep{yamada2026diffusing}. 
We further demonstrate our framework across three Machine Learning research directions: our system discovers composition and optimization biases for hypernetwork-based LLM adaptation~\citep{charakorn2026doctolora}.
\agentscientist also extends research on neural network-parametrized meta-black-box optimization~\citep{ma2025toward}, yielding an optimizer now integrated in \texttt{evosax}~\citep{lange2023evosax}.
Finally, our system manages to iteratively improve an RL-based LLM fine-tuning program by combining recent algorithmic advances.
All of these demonstrate the broad application of the system to enhance scientific knowledge beyond mere hill-climbing applications.
Our main contributions are:
\begin{enumerate}
    \setlength{\itemsep}{0pt}
    \item \textbf{\agentscientist}, a (co-)scientist orchestration scaffold for human-steerable ideation, parallel execution, shared memory, rubric-based verification, and paper manuscript writing.
    \item \textbf{Co-scientist paper conference acceptance.} A paper created with assistance by \agentscientist on backpropagation-free error diffusion learning algorithms was accepted at ALIFE 2026 \citep{yamada2026diffusing} after limited human edits and verification.
    \item \textbf{Cross-domain \& harness validation} spanning long-context LLM adaptation, evolutionary optimization, RL fine-tuning, and 5 additional domains. Across various tasks, model harnesses, and base models, our system managed to generate high-quality grounded research manuscripts.
    \item \textbf{\autoillustrate}, an agentic pipeline for editable scientific figure generation that outperforms PaperBanana on PaperBananaBench ~\citep[$37.5$ vs.\ $25.0$ overall score,][]{zhu2026paperbanana}.
\end{enumerate}

\newpage
\section{Background \& Related Work}
\paragraph{End-to-end automated scientific discovery.}
Research on LLM-driven science increasingly targets complete workflows rather than isolated steps~\citep{si2025can}.
The \aiscientistfirst introduced a closed loop for proposing hypotheses, running experiments, analyzing results, and drafting manuscripts~\citep{lu2024ai}, while The \aiscientistsecond removed human-provided code templates and introduced progressive agentic tree search~\citep{yamada2025ai}. Ultimately, these developments combined with base LLM capabilities have led to end-to-end autonomous systems capable of producing workshop-level research content \citep{lu2026towards, intology2025zochi, autoscience2025carl}.
Agent Laboratory and AI-Researcher similarly connect literature analysis, experimentation, and report writing, with different opportunities for human feedback~\citep{schmidgall-etal-2025-agent,tang2025airesearcher}.
CodeScientist combines literature and code search with code review, replication, and external assessment~\citep{jansen2025codescientist}, while CycleResearcher inserts automated review into the loop~\citep{weng2025cycleresearcher} and the AI co-scientist emphasizes scientist-guided hypothesis development~\citep{gottweis2025towards}.
These systems vary in autonomy, search, and validation; here our focus is the reusable plus scalable orchestration and evidence layer that supports heterogeneous coding agents.

\vspace{-0.25em}
\paragraph{Hill-Climbing via Program Evolution.}
A complementary line casts discovery as search over executable artifacts.
FunSearch showed that language-model-guided program search can yield mathematical discoveries~\citep{romera2024mathematical}; AlphaEvolve broadened evolutionary code search to algorithms and computational infrastructure~\citep{novikov2025alphaevolvecodingagentscientific}, and ShinkaEvolve targets open-ended, sample-efficient program evolution~\citep{lange2026shinkaevolve}.
At a lighter weight, \texttt{autoresearch} uses coding agents to repeatedly edit and evaluate a training program under a fixed budget~\citep{karpathy2026autoresearch}, motivating direct comparison with classical hyperparameter optimization~\citep{ferreira2026can}. ALE-Agent \citep{imajuku2026ale} demonstrated that simple beam search at sufficient scale provides a competitive harness for program optimization.
Across these approaches, the search unit is a program under a pre-specified objective.
Our scaffold uses such executable search within a broader hypothesis-to-report process: it broadens the scope of application by automating task drafting, validation rubric generation, and supporting objective creation.

\vspace{-0.25em}
\paragraph{Benchmarking Research Agent Capabilities.}
Research-agent benchmarks separate several capabilities that a single final score can obscure.
SWE-agent~\citep{yang2024sweagent} established that the agent--computer interface strongly affects real-world coding performance.
MLAgentBench evaluates iterative machine-learning experimentation~\citep{huang2024mlagentbench}, whereas MLE-Bench covers longer-horizon engineering workflows involving data preparation, debugging, training, and optimization~\citep{chan2024mle}; subsequent work studies how search strategy affects exploration and generalization in this setting~\citep{toledo2025neurips}.
At the research level, PaperBench measures replication of published AI work~\citep{starace25a}, MLR-Bench targets open-ended machine-learning research~\citep{chen2025mlrbench}, and EXP-Bench isolates whether agents can conduct valid AI experiments~\citep{kon2026expbench}.
\citet{lu2024ai} introduced an LLM-driven pipeline for paper reviewing, which was later validated on research papers published after the training data cutoff date \citep{lu2026towards}.
Here, we make use of such automated proxy AI reviews and additionally report external reviews of the co-produced ALIFE submission as complementary evidence.

\vspace{-0.25em}
\paragraph{Verification and Grounding.}
Verification and communication are becoming explicit parts of agent design.
Rubric rewards can train AI co-scientists and provide feedback in domains without verifiable objectives~\citep{goel2026training,gunjal2026rubrics}.
Here, rubrics do not train the coding-agent backbone; instead, they serve as runtime contracts that ground each hypothesis to implementation checks and empirical evidence. The \agentscientist manuscript drafting runs additional checks for facts and figures outlined in the paper. We discuss further related work in \Cref{app:additional_related_work}.

\section{\agentscientist: An Orchestration Layer for Research Agents}
\label{sec:method}
\agentscientist is a versatile agentic orchestrator scaffold for end-to-end computational \textit{research campaigns} (\Cref{fig:conceptual}).
A campaign begins with a task specification, which researchers can provide directly or generate from a research brief with a drafting agent.
Then, the campaign proceeds through three phases: parallel investigation, retrospective evidence generation, and grounded paper writing.
Each phase runs as a separate process and feeds the next: the selected program and experimental artifacts produced by the first phase determine which baselines, ablations, and post-hoc analyses to run; the resulting evidence then grounds the manuscript. Finally, we leverage the previously introduced AI reviewer \citep{lu2026towards} to provide proxy evaluation scores for the manuscripts.

\subsection{Research Task Setup}
\label{sec:task_spec}

Before idea execution, \agentscientist requires an executable task specification that defines the research goal and its evaluation. This can either be specified manually or drafted agentically:

\begin{wrapfigure}{l}{0.4\linewidth}
    \vspace{-0.8em}
    \centering
    \includegraphics[width=0.95\linewidth]{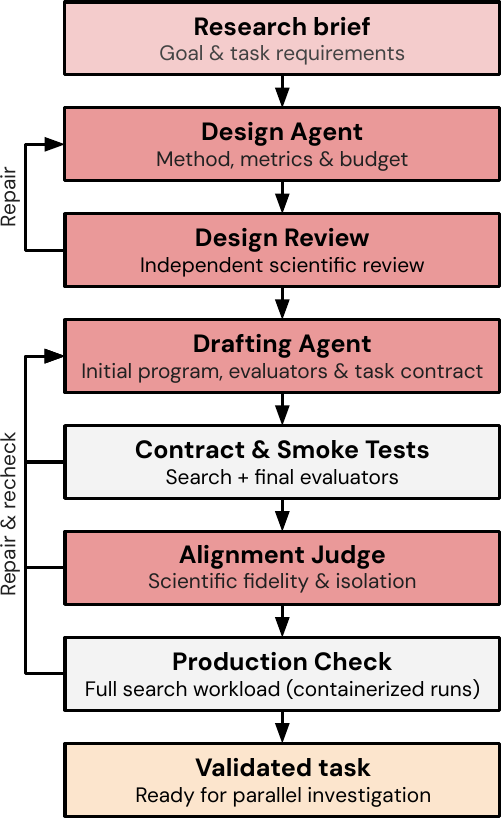}
    \caption{Task drafting workflow.}
    \label{fig:method_task_drafting}
    \vspace{-1em}
\end{wrapfigure}

\paragraph{Task specification.}
A task specification comprises an initial program, a search evaluation program, an optional final evaluation program, and a machine-readable contract:
the initial program provides a deterministic baseline, while the search evaluation program determines whether a candidate program is correct, returns the scalar score used to rank and select candidates during the investigation phase, and records metrics and artifacts for later analysis.
When provided, the final evaluation program runs the selected program after exploration and reports post-search evaluation metrics.
The contract defines the candidate interface and evaluation procedures.
Tasks may additionally stage papers or code as research context, as in our hypernetwork adaptation and error-diffusion studies (\Cref{sec:doc_to_lora,sec:ai-co-scientist-alife}).
\vspace{-0.75em}
\paragraph{Task drafting.}
Researchers may provide the task specification directly or use the task drafting pipeline to create it from a human-written research brief (\Cref{fig:method_task_drafting}).
The pipeline has two main stages: a design agent converts the research brief into an experimental design, and a drafting agent converts this design into the task specification.

The experimental design defines the method, candidate interface, compute budget, and protocols for search and final evaluations.
Before drafting begins, an independent review agent assesses the design and repairs any issues.
The drafting agent then implements the initial program, the search and final evaluation programs, and the task contract, while pinning dependencies and staging the required data.
The harness checks the contract and runs smoke tests of the evaluation programs before the alignment judge assesses scientific fidelity and verifies that the search and final evaluators are independent and use disjoint data.
In containerized runs, the production check additionally requires the initial program to complete the full search workload to ensure operational feasibility.
Failed scaffolds are repaired, and only versions that pass all checks proceed to parallel investigation.

\subsection{Parallel Investigation}
\label{sec:parallel_workers}

The idea investigation phase spends a fixed budget of containerized research cycles across parallel workers, each exploring different hypotheses in a separate workspace while sharing distilled knowledge (\Cref{fig:method_parallel_loop}).
Multiple agent-based workers explore ideas in parallel, each advancing an independent sequence of research cycles.
These workers lease cycles from a synchronized queue, maintain separate run directories and worker-local best search score programs, which resemble a per-worker beam search.
A cycle starts with the worker importing the shared memory and the other workers' distilled conclusions, and ends with the worker publishing new memory entries and updated conclusions to all other workers.
Communication occurs at cycle boundaries, while logs and program artifacts remain in worker directories.
At regular intervals, optional context pruning archives the histories and memory before resetting them, so that accumulated context does not bias subsequent cycles; the best program is carried into the next round to preserve progress.
After parallel execution, the aggregator merges worker histories and selects the best correct program found across workers. This selected program becomes the input to the retrospective evidence-generation phase.
\subsection{Contracted Research Cycles}
\label{sec:individual_workers}
\begin{wrapfigure}{r}{0.38\linewidth}
    \vspace{-3.5em}
    \centering
    \includegraphics[width=0.95\linewidth]{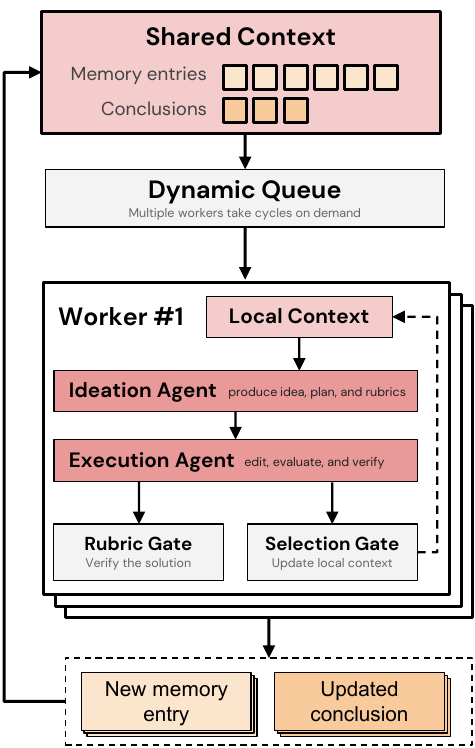}
    \vspace{-0.5em}
    \caption{Investigation workflow.}
    \label{fig:method_parallel_loop}
    \vspace{-3.0em}
\end{wrapfigure}
Each worker advances a sequence of research cycles, and each cycle runs two agents in turn: an ideation agent and an execution agent.
The ideation agent turns one hypothesis into a typed contract, and the execution agent then tests that contract against the evaluator, so that intent and evidence remain separate(\Cref{fig:method_parallel_loop}).

\paragraph{Ideation agent.}
At the start of a cycle, the ideation agent receives the task, recent idea and attempt histories, the manifest for the worker's best program, and distilled conclusions.
The protocol requires inspecting this context without experimenting or modifying candidates (`read-only').
The ideation agent must instead emit a structured idea bundle containing a title, a falsifiable hypothesis, an ordered implementation plan, and the target rubrics.
Those records are normalized, and any missing fields or mismatched rubric identifiers are rejected.
This boundary records the idea as a structured artifact, so the execution agent does not have to reconstruct its intent from conversational context.

\paragraph{Execution agent.}
The accepted contract, evaluator, task files, and the worker's best program are staged into an experiment workspace.
The execution agent creates append-only candidate variants and follows an edit--evaluate--verify loop within a fixed evaluation budget, producing metrics, correctness checks, and rubric evidence.
Two distinct gates are then applied: an all-rubric-pass attempt marks the contract complete, whereas updating the worker's best program requires a correct candidate with a strictly higher score; rubric completion and updates to the worker's best program are therefore independent decisions.
All logged idea provenance and attempt outcomes are retained and later condensed to guide interpretation during paper writing.

\subsection{Retrospective Evidence Generation}
\label{sec:retrospective_evidence}

This phase studies the selected program from the investigation phase in depth: (1) it runs controlled baselines and ablations that isolate the idea effects, and (2) it derives post-hoc analyses that characterize the behavior of the method on the considered task.

\paragraph{Controlled baselines and ablations.}

The selected program is first instrumented with additional metrics, such as extra scores and time series data, so that every generated baseline and ablation can be compared in detail on common ground.
Given the code diff between the selected and initial programs, an agent then proposes baselines and ablations with explicit hypotheses and implementation plans, where each ablation must isolate an innovation introduced by the selected program.
A separate agent filters these proposals for relevance, feasibility, and single-change scope.
Each validated plan receives an isolated workspace and follows an agent-guided edit--evaluate--verify loop that modifies the selected program to implement the intended change and score it against the evaluator, alongside the unchanged selected and initial programs.
The evaluator can be run for multiple seeds to make comparisons statistically meaningful, and their summaries and metrics are aggregated across seeds, ablations, and baselines by task.

\paragraph{Post-hoc analysis.}

The post-hoc pipeline proposes, filters, executes, and aggregates diagnostics that characterize the behavior of the frozen selected program on the considered task.
Diagnostics may run new inference or analyze existing artifacts, such as checkpoints and logs.
Programmatic checks reject proposals that retrain or otherwise alter the selected program, or require unavailable artifacts.
When no usable checkpoint exists, an optional bootstrap stage enables checkpoint saving and reruns the evaluator once; all subsequent analyses keep the resulting version frozen.
The resulting post-hoc analyses, together with the baseline and ablation results, form the evidence that the paper writing phase turns into empirical claims.

\subsection{Grounded Paper Writing}
\label{sec:paper_writing}

\begin{wrapfigure}{l}{0.38\linewidth}
    \vspace{-1.5em}
    \centering
     \includegraphics[width=\linewidth]{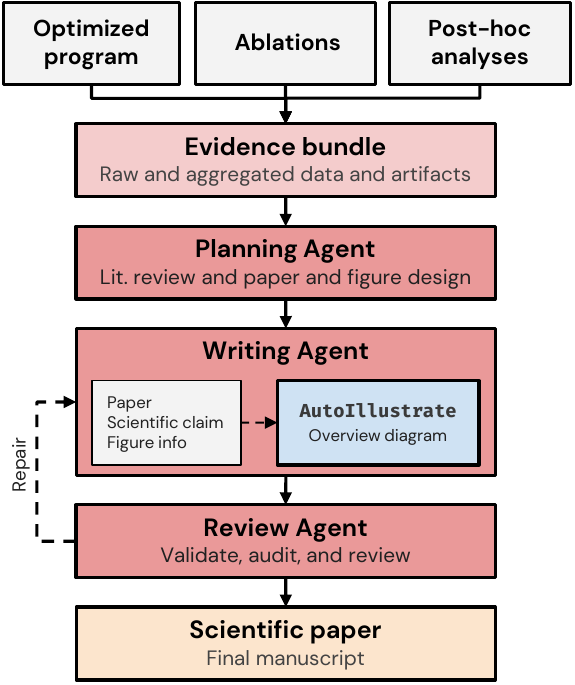}
    \caption{Evidence-to-paper pipeline.}
    \label{fig:method_grounded_reporting}
    \vspace{-3.5em}
\end{wrapfigure}

The paper writing phase converts the outputs of the preceding phases into a manuscript through an artifact-grounded workflow that reviews and audits claims made in the paper (\Cref{fig:method_grounded_reporting}).

\paragraph{Evidence staging and planning.}
We merge worker results into aggregate outputs and stage them in a dedicated workspace, with a memory summary and raw artifacts.
Quantitative claims resolve to those outputs and raw artifacts, while the memory summary only guides interpretation.
Before and after drafting, reference planning retrieves bibliography entries (\Cref{app:bibliography-retrieval}). An optional stage fixes the central thesis, result emphasis, and figure plan.

\paragraph{Manuscript generation.}
The writer agent produces a compilable LaTeX manuscript together with a claim-to-source map and figure manifest, whose plots are derived from staged workspace results.
The strategy may also request \autoillustrate, our agentic figure generation pipeline (\Cref{sec:autoillustrate}), to produce a conceptual overview diagram.

\paragraph{Validation and review.}
At the end of our pipeline, a single repair loop alternates deterministic checks with agent judgment and feeds any failures back to the writer agent for revision.
The validation checks cover structural completeness, provenance and claim-to-source links, compilation, and venue and page budget constraints.
An independent audit agent runs every round on a read-only copy of the workspace and flags where the paper's numbers, comparisons, or conclusions overclaim or contradict the canonical data and evidence trail.
An optional reviewer-style agent assesses framing, caveats, figures, and clarity, and may ask the writer agent to narrow or qualify claims.

\newpage
\subsection{\autoillustrate: Conceptual Diagram Generation}
\label{sec:autoillustrate}
\begin{figure}[htbp]
    \centering
    \vspace{-0.8em}
    \includegraphics[width=0.99\linewidth]{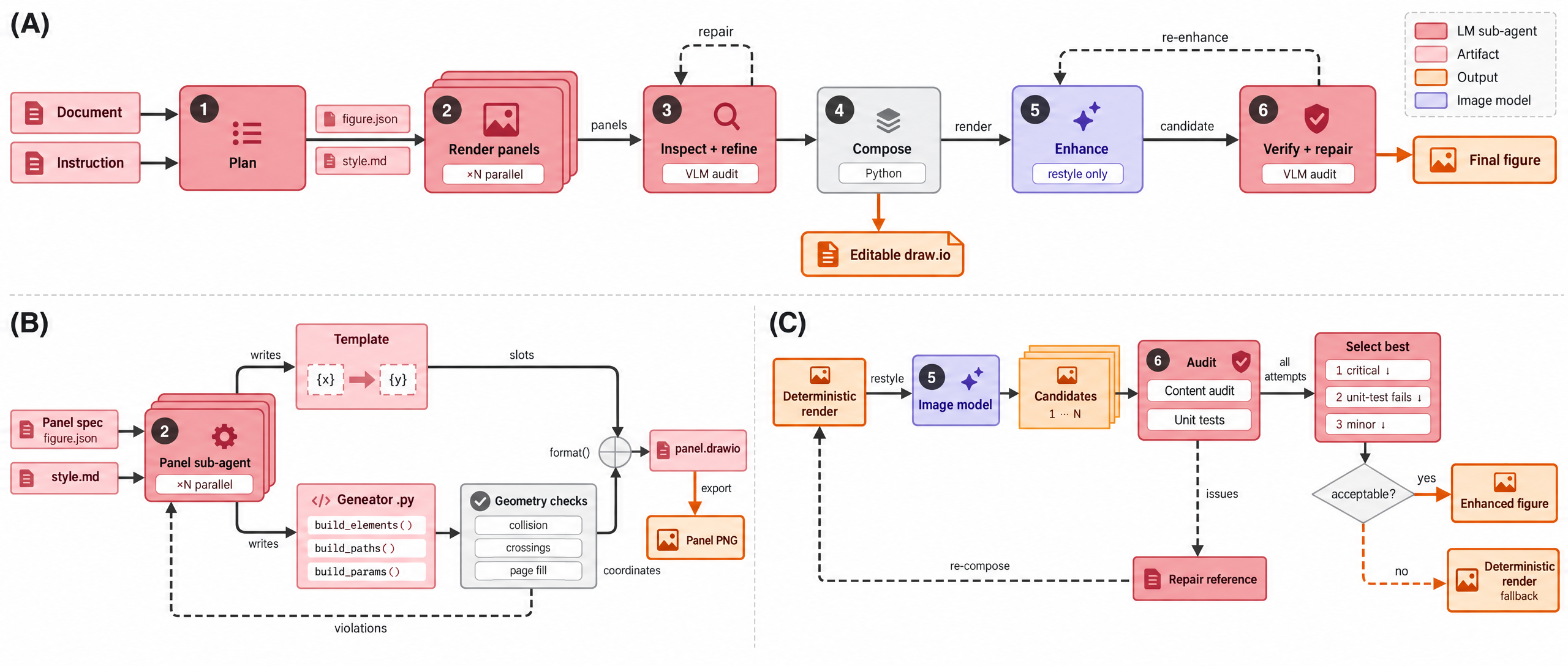}
    \vspace{-1.4em}
    \caption{\textbf{\autoillustrate pipeline}, generated by \autoillustrate itself. \textbf{(A)} Six phases turn a document and an instruction into an editable \texttt{draw.io} figure: LLM sub-agents decide content, while deterministic Python owns geometry and the generative image model only restyles. \textbf{(B)} In Phase 2, panels are rendered from a combination of automatically-generated templates whose coordinates are determined by LLM-generated Python code. \textbf{(C)} In Phases 5--6, the enhanced candidate with the best audit results is selected; if none is acceptable, the deterministic render is used as a fallback.}
    \label{fig:writing_pipeline_gpt_image_2}
\end{figure}

\autoillustrate is designed around two requirements: scientists should be able to modify conceptual figures, and they should represent the methodology faithfully.
To meet these requirements, the pipeline first produces an editable \texttt{draw.io} diagram to ground the figure's structure, which is later exported and augmented using a separate image generation model.
As illustrated in \Cref{fig:writing_pipeline_gpt_image_2}, the concept generator runs a six-phase pipeline that turns a method brief into a multi-panel figure whose layout source remains explicitly editable.
Unless stated otherwise, all papers generated by our system in this work have their main conceptual figure generated by \autoillustrate.

\vspace{-0.5em}

\paragraph{Editable panel generation.}
The pipeline begins with a planning agent that reads the staged source and instructions and describes the panels within one top-level configurable layout.
A styling agent then defines a shared color palette and typography specification to keep independently rendered panels consistent.
Rendering agents create the panels independently and in parallel as editable \texttt{draw.io} diagrams.
Each agent relies on a template to define the panel topology and a Python generator script to compute coordinates and check bounding boxes for overlap.
Visuals that \texttt{draw.io} cannot produce remain as placeholder boxes.
When possible, a generated \texttt{matplotlib} script fills these boxes; the enhancement stage handles any remaining placeholders.
A vision-model audit then evaluates each panel for rendering fidelity, factual accuracy, and sufficient detail.
When the inspection detects a defect, the pipeline edits the source, rebuilds the panel, and repeats the audit.

\vspace{-0.5em}

\paragraph{Composition, enhancement, and verification.}
After all panels pass, the pipeline composes them into a single editable \texttt{draw.io} source and exports a raster version.
An image generation model restyles the raster export, after which a final verification step compares the enhanced candidate with the editable reference and repeats the checks defined during panel inspection.
Any discrepancy triggers another source-editing, export, enhancement, and verification pass.
After several repair passes, the pipeline returns the enhanced candidate with the strongest verification results.
If no candidate passes verification, the pipeline returns the raster export from the editable source.
Because the pipeline preserves this source throughout the process, scientists can revise the content or layout directly without regenerating the figure.
We provide additional details in \Cref{app:autoillustrate}.

\newpage
\section{AI Co-Scientist and Conference Paper Submission}
\label{sec:ai-co-scientist-alife}

\begin{figure}[h]
    \centering
    \includegraphics[width=0.99\linewidth]{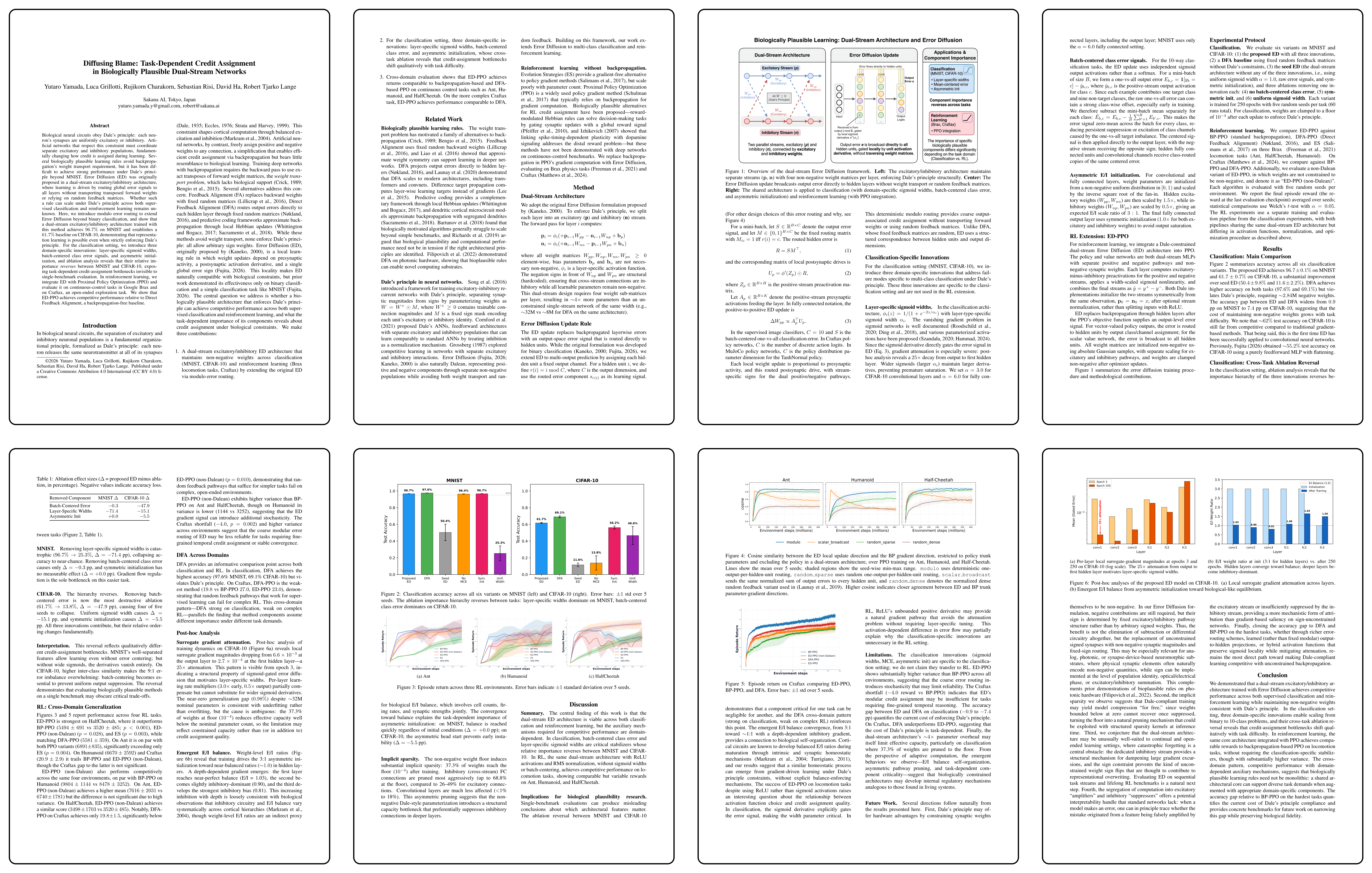}
    \caption{Camera-ready ALIFE 2026 paper after the single submitted manuscript was accepted. The full manuscript and additional details can be found in \Cref{app:error_diffusion_details}.}
    \label{fig:alife_camera_ready}
\end{figure}

We validated a development version of our system within a co-scientist setting: \agentscientist was used to automate parts of the research process targeting the extension of a backpropagation-free neural network training method. 
After thorough human verification and manuscript edits, we submitted a single paper to ALIFE 2026, which was accepted after external peer review as a main conference talk.
We emphasize that this single submitted ALIFE paper was produced by human researchers using \agentscientist as a tool, and not submitted as an unaltered system output. The human authors remain responsible for its methods, claims, authorship, and disclosure.
In this section, our focus is to describe the process by which experimental outputs became a submitted manuscript and were subsequently revised for the camera-ready version (\Cref{fig:alife_camera_ready}). 

\subsection{A Human-Supervised Research Workflow}
\label{sec:ed-workflow}

The project began with a human-selected research direction: investigating how Error Diffusion~(ED) could be extended beyond its earlier demonstrations \citep{kaneko2000error_diffusion,fujita2026diagnostic}. ED leverages a three-factor learning rule (presynaptic activity, postsynaptic activity, and a globally broadcast error signal) originally studied in a dual-stream setting with paired excitatory (E) and inhibitory (I) neurons. We chose to illuminate this research direction together with \agentscientist since it investigates the following question: how can classical biologically-inspired learning algorithms be revived with modern deep learning techniques? 

\begin{figure}[tbp]
    \centering
    \includegraphics[width=0.99\linewidth]{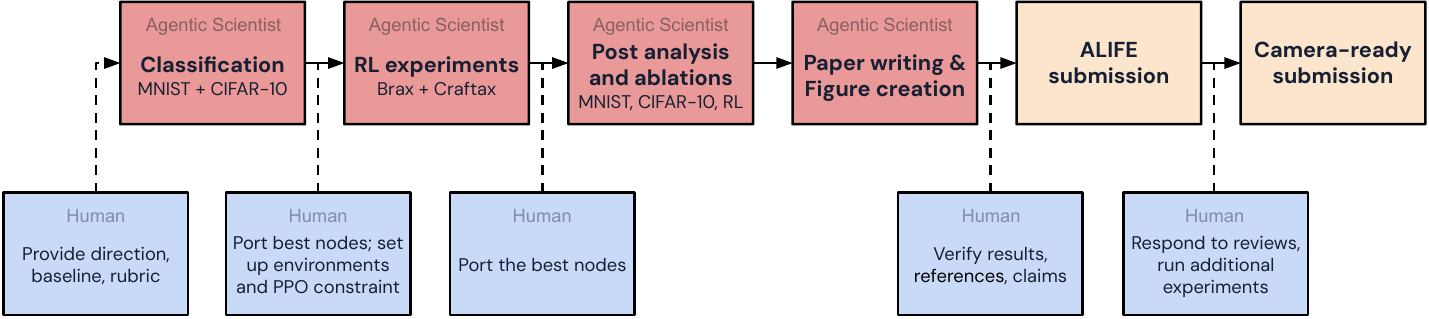}
    \caption{Creation process of the ALIFE paper: human researchers defined the initial direction, baseline, and rubrics; supported the experiment domain transition; verified results, references, and claims before submission; and incorporated reviewer feedback into the camera-ready paper.}
    \label{fig:human_intervention_flow}
\end{figure}

Human researchers specified the scientific objective and the constraints for a valid implementation. 
Within that contract, \agentscientist proposed candidates, implemented training procedures, ran experiments, inspected failures, and generated plots. 
The human researchers selected benchmarks (see creation process overview in \Cref{fig:human_intervention_flow}) and specified initial rubrics: 
A numerically strong result was not treated as evidence of progress unless its implementation also satisfied the rubric constraints. 
For example, one rubric prohibited hidden credit assignment through automatic differentiation or transposed forward weights. This required the positive and negative streams to remain distinct dual-stream, and specified the intended local information available to an update.

\subsection{Summary of Discovery and Dynamics}

\agentscientist first extended ED from binary to multiclass classification: during the parallel investigation phase, \agentscientist developed \emph{modulo error routing}, a structured mechanism for assigning multi-class output errors to hidden units. 
This enabled multiclass learning while preserving the dual-stream, Dale-constrained architecture, in which weight signs remain fixed throughout training. 
Subsequent experiments focused on robustifying learning dynamics: \agentscientist identified layer-specific sigmoid widths, batch-centered error signals, and asymmetric E/I initialization as important stabilizers, raising accuracy to $96.7\%$ on MNIST \citep{lecun1998mnist} and $61.7\%$ on CIFAR-10 \citep{krizhevsky2009cifar10}. 
Ablations further showed that the individual contributions were task-dependent: widened sigmoids were critical on MNIST, whereas batch-centering became substantially more important on CIFAR-10.

At this stage, human researchers transferred the best-performing ED implementation from the MNIST and CIFAR-10 experiments to the RL setting, and prepared the Brax \citep{brax2021github} and Craftax \citep{matthews2024craftax} environments.
For RL experiments, an additional methodological constraint was added to the rubrics: the method was required to retain the PPO objective \citep{schulman2017proximal}, while using the gradient of the PPO loss with respect to the logits as the output-level error signal for Error Diffusion. 
Within this constraint, \agentscientist adapted the dual-stream ED mechanism to PPO and explored the resulting ED-PPO variants. 
ED-PPO achieved performance comparable to standard baselines on several Brax tasks, although its performance was less consistent on Humanoid and Craftax (see figures in ALIFE paper manuscript; \Cref{paper:alife2026_camera_ready}).

Post-hoc analyses linked these results to the dynamics of credit assignment: local update signals weakened substantially in earlier layers, while training shifted E/I ratios toward balance. 
In summary, the results obtained demonstrate that ED generalizes beyond its original binary-classification setting. The mechanisms required for effective learning depend on task difficulty and domain.

\subsection{Human Result Verification \& Manuscript Edits}

Before submission to ALIFE, researchers verified the generated research artifacts. 
This included checking that the generated implementations matched the scientific hypotheses being tested, confirming that the reported tables and figures accurately reflected results produced by the corresponding code, and revising the manuscript where the generated text overstated or inaccurately characterized the evidence.
For example, human researchers added appropriate statistical analyses and, when the results showed comparable performance, revised claims such as ``ED-PPO significantly exceeds BP-PPO'' to ``ED-PPO performs on par with BP-PPO''.
Additionally, all references in the BibTeX bibliography were manually verified to ensure that the cited works existed and supported the statements for which they were cited.
During this process, human researchers removed several citations that did not support the corresponding claims, and added important references omitted from the generated manuscript.
These included the reference to the original ED work, which is available only through an archived webpage.
The generated manuscript was treated as an intermediate research artifact, which was modified by human researchers, improving the writing of several sections (e.g., future work, etc.). We then submitted this single paper to the ALIFE 2026 conference.

\begin{table}[t]
    \small
    \centering
    \caption{Revision statistics across manuscript stages.
    Counts are word-equivalents computed directly from the
    \texttt{latexdiff} LaTeX sources.
    Percentages are relative to the pre-revision manuscript
    length.}
    \vspace{-0.5em}
    \begin{tabular}{lrrrr}
        \toprule
        Revision stage & Inserted & Deleted & Net change & \\
        \midrule
        Agentic Scientist $\rightarrow$ submitted
            & 664 (17.5\%) & 405 (10.7\%)
            & $+259$ ($+6.8$\%)  \\
        Submitted $\rightarrow$ camera-ready
            & 1,329 (32.8\%) & 630 (15.6\%)
            & $+699$ ($+17.3$\%) \\
        \bottomrule
    \end{tabular}
    \label{tab:revision_statistics}
\end{table}

\subsection{Peer Review, Camera-Ready Paper Revisions}

This submission process illustrates the AI co-scientist approach enabled by our system: outputs were not just optimized code or isolated experimental reports, but a coherent working paper making a scientific claim, positioning it relative to prior work, and presenting evidence across multiple domains. We emphasize that only a single paper was submitted after our careful manual verification.

The subsequent ALIFE peer review process provides external evidence for the quality of the co-created paper submission.
Overall, the paper was accepted as a talk, with individual recommendations of \textit{Strong ACCEPT}, \textit{Weak REJECT}, and \textit{Weak ACCEPT}:

\vspace{-0.2em}
\texttt{Reviewer 1} framed the work as a relevant artificial-life contribution, as it evaluates the consequences of a biological assumption, Dale's principle, and found the methods and results technically sound.\\
\texttt{Reviewer 2} highlighted that the neuroscience connection of the work was interesting but questioned whether the paper would better fit a neural network venue. Additionally, they asked for clearer explanations of notation, matrix conventions, modulo indexing, and technical details. \\
\texttt{Reviewer 3} was positive and requested more intuition about ED updates, non-negative activations, per-neuron error signals, baseline comparisons, ED-PPO architecture, and the meaning of gradient attenuation without backpropagation. 

We afterward used this feedback to enhance the camera-ready version of the manuscript: the post-acceptance changes are best characterized as targeted scientific clarifications rather than a change in the underlying study. 
In particular, the camera-ready version tightens several claims about prior ED results, adds Dale-compliant reinforcement-learning results, corrects the Error Diffusion citation context, clarifies that Direct Feedback Alignment \citep{nokland2016direct} routes errors through fixed random matrices, and cleans up the dual-stream architecture description and ED update equation. 
It also replaces potentially misleading gradient language with ``proxy'' or ``local surrogate'' gradient terminology, refines the biological interpretation of E/I balance, repositions the discussion of Craftax and the dual-stream parameter overhead, and fixes typographic details such as quotation marks. 
These edits make the paper more precise about biological plausibility and local learning while leaving the empirical evidence and overall narrative intact.
Table~\ref{tab:revision_statistics} summarizes human intervention throughout the manuscript-development process.
In summary, this study provides evidence of \agentscientist being capable of supporting researchers to explore unorthodox AI topics across the entire research process.

\newpage
\section{Evaluating \agentscientist Across Tasks, Harnesses, and LLMs}
\label{sec:broad_results}
We evaluate \agentscientist across five manually prepared research tasks that span optimization, representation learning, and predictive modelling. Each campaign starts from a task specification with different research objectives, starting points, and search iteration resources (see \Cref{tab:aggregate_tasks} for the task suite overview).

\paragraph{Task, harness, and model sweep.}
We run the task sweep across four agent harnesses: Codex, Claude Code, Pi, and OpenCode~\citep{openai_codex_github_2026,anthropic_claude_code_github_2026,earendilworks_pi_github_2026,anomaly_opencode_github_2026}. Codex and Claude Code each use four assigned models, while Pi and OpenCode each use three, all at high reasoning effort (\Cref{tab:aggregate_harness_models}). Each run uses 20 idea investigation iterations and concludes with a written manuscript. Here, we focus on aggregated insights and provide more details on task-specific settings and discoveries can be found in \Cref{app:additional_sweep}.

\begin{table}[htbp]
    \centering
    \scriptsize
    \setlength{\tabcolsep}{5pt}
    \renewcommand{\arraystretch}{1.1}
    \caption{Research tasks, starting implementations, and objectives used in the aggregate comparison.}
    \label{tab:aggregate_tasks}
    \begin{tabularx}{\linewidth}{@{}>{\raggedright\arraybackslash}p{0.19\linewidth}>{\raggedright\arraybackslash}X>{\raggedright\arraybackslash}X@{}}
        \toprule
        Task & Starting Point & Research Objective \\
        \midrule
        Learned optimization & \underline{Baseline}: ELO's expert-supervised MLP optimizer, meta-trained on (Fashion-)MNIST \citep{huang2026efficient}; \underline{Evaluation}: CIFAR-10 and SVHN. & Improve transfer to unseen datasets and network architectures under a fixed meta-training budget. \underline{Search iteration budget}: 2~H100s, 150~min. \\
        World models & \underline{Baseline}: Action-free Pong model, with a trained discrete tokenizer and masked-token dynamics; \underline{Evaluation}: Error reduction against TinyWorlds & Reduce held-out free-running rollout error across prediction horizons relative to \citet{majmudar2026tinyworlds}. \underline{Search iteration budget}: 4~H100s, 180~min. \\
        Sparse autoencoders (SAE) & \underline{Baseline}: TopK SAE activation-trained on Pythia- 160M-deduped \& Gemma-2-2B-it; \underline{Evaluation}: Task-specific SAEBench \citep{karvonen2025saebench}. & Improve reconstruction at fixed sparsity while preserving model behavior and feature utilization. \underline{Search iteration budget}: 2~H100s, 60~min. \\
        Tabular distribution shift & \underline{Baseline}: TabICLv2's \citep{qu2026tabiclv2} stage-one synthetic-prior pretraining from scratch recipe; \underline{Evaluation}: 4 held-out real datasets vs. reference. & Improve in-context balanced accuracy on held-out real table data \& across context sizes. \underline{Search iteration budget}: 4~H100s, 150~min. \\
        Hyperparameter optimization & \underline{Baseline}: Adaptive incumbent-centred local search; \underline{Evaluation}: BBOB \citep{hansen2021coco}, HPO-B \citep{arango2021hpo}, YAHPO \citep{pfisterer2022yahpo}. & Improve sample-efficient optimization over continuous, mixed, and hierarchical search spaces. \underline{Search iteration budget}: 8~CPU cores, 30~min. \\
        \bottomrule
    \end{tabularx}
\end{table}

\paragraph{Paper-level evaluation protocol.}
All generated research runs share the same scoring function within a task instance and can thereby be directly compared on their numerical optimization outcome. Additionally, we use the AI Reviewer's \citep{lu2026towards} 1--10 \emph{Overall} score as a paper-level proxy metric:
For each generated manuscript, five separate calls to \texttt{o4-mini-2025-04-16} review the extracted PDF text using NeurIPS-style guidelines and a fixed in-context example of a paper--review pair. A sixth call acts as an area chair to synthesize the reviews.
The original study in \citet{lu2026towards} validated the reviewer against acceptance decisions for ICLR submissions from OpenReview, including a post-training-data-cutoff 2025 cohort.
We note that the AI reviewer score solely serves as a proxy for reviewer-perceived paper quality, and not as ground-truth evidence of full scientific correctness.

\vspace{-0.5em}

\begin{figure}[h]
    \centering
    \includegraphics[width=1.0\linewidth]{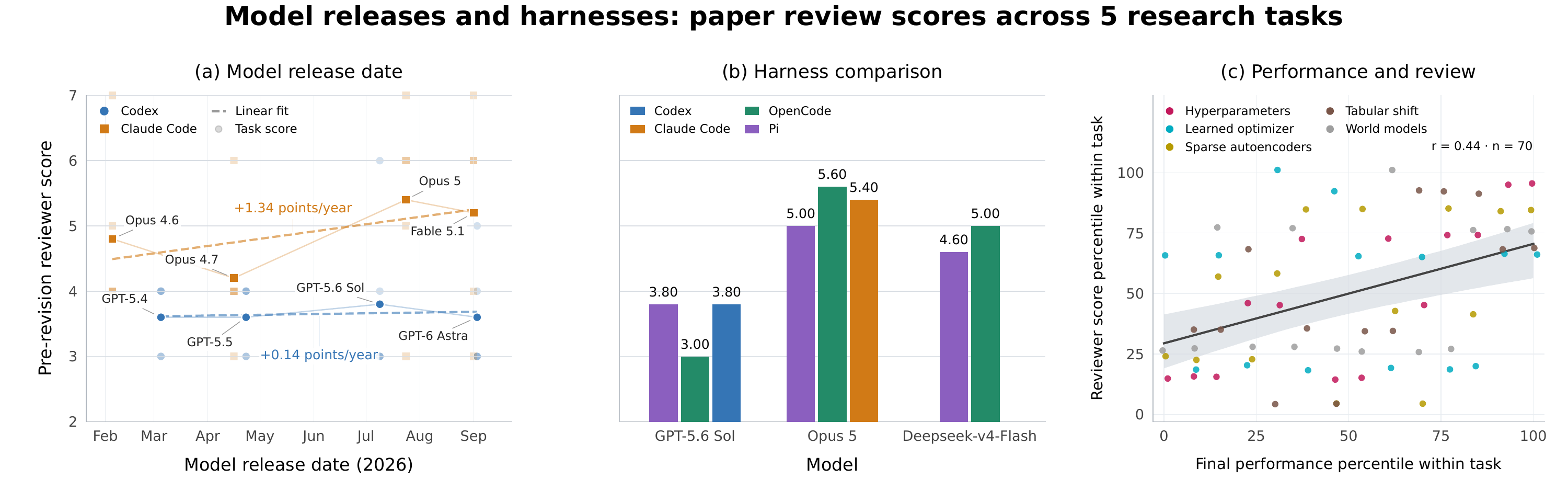}
    \caption{Pre-revision reviewer scores by (a) model release date and (b) harness, averaged across five tasks; (c) within-task percentile ranks of final performance and reviewer scores, colored by task. The least-squares fit uses a pointwise 95\% band from 10,000 bootstrap resamples of the 14 complete harness--model profiles across all five tasks, conditional on the observed tasks and fixed ranks.}
    \label{fig:aggregate_reviewer_scaling}
\end{figure}

\paragraph{Model capability, harness impact, and paper quality.}
Generally speaking, newer model releases tend to be associated with higher reviewer scores (\Cref{fig:aggregate_reviewer_scaling}a). When evaluating the models within their respective native coding harnesses, this observation appears to be more strongly pronounced for Claude- than GPT-based models.
Matched-model comparisons show minor performance differences across harness choice, while the underlying model matters more (\Cref{fig:aggregate_reviewer_scaling}b). No harness is uniformly best. OpenCode ranks lowest with GPT-5.6-Sol but highest with Opus~5 and DeepSeek-v4-Flash. The model developers' own harnesses offer no consistent advantage: Codex ties Pi with GPT-5.6-Sol, and Claude Code falls between OpenCode and Pi with Opus~5.

\vspace{-0.5em}

\begin{figure}[H]
    \centering
    \includegraphics[width=1.0\linewidth]{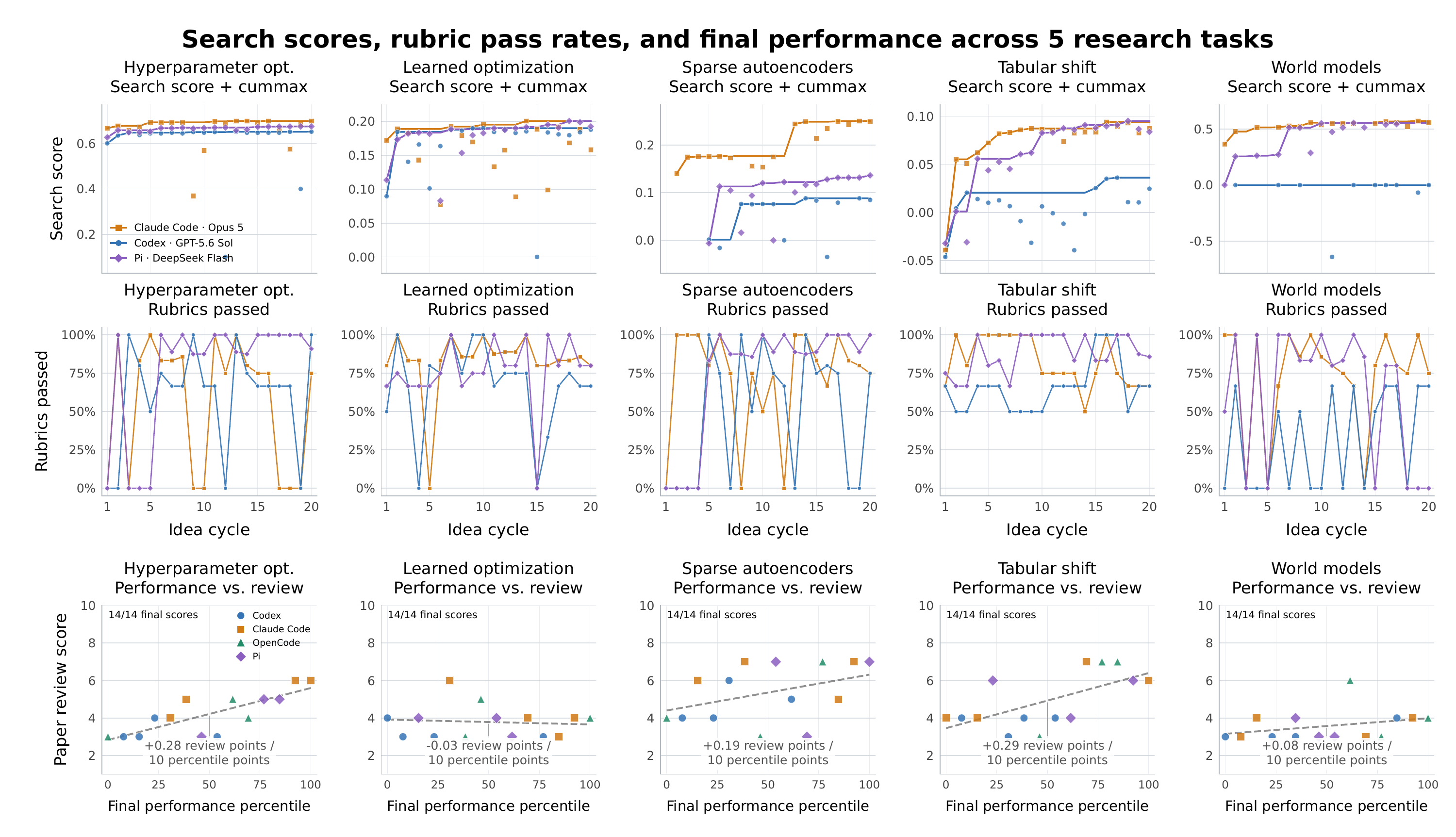}
    \caption{Search scores (top), rubric pass rates (middle), and final performance versus reviewer score (bottom) across five tasks. Bottom panels show within-task performance percentiles against original 1--10 reviewer scores; annotated least-squares slopes give reviewer-score points per 10 percentile points. Final performance is positively associated with reviewer scores on four of the five tasks.}
    \label{fig:aggregate_search_cycles}
\end{figure}

\paragraph{Search progress and rubric dynamics.}
The search traces show gains in best observed scores alongside fluctuating rubric pass rates. Final performance is positively associated with reviewer scores on four of the five tasks, most strongly for hyperparameter optimization, while learned optimization shows little association (\Cref{fig:aggregate_search_cycles}).
Pooling all 70 campaigns after converting both final performance and reviewer scores to within-task percentile ranks gives a positive aggregated association ($r=0.44$; \Cref{fig:aggregate_reviewer_scaling}c). 
This observation is consistent with manuscript reviewer scores tracking measured numerical performance.

\vspace{-0.5em}

\newpage
\section{Additional Case Studies: Putting \agentscientist to work}
\label{sec:case_studies}

We further evaluate \agentscientist on three case studies: Language model adaptation (\Cref{sec:doc_to_lora}) through hypernetworks \citep{charakorn2026doctolora}, evolutionary optimization (\Cref{sec:lcma}) through meta-learned neural network parametrizations \citep{lange2023discovering_es}, and language model RL fine-tuning (\Cref{sec:rlvr}) on mathematical reasoning tasks.

\subsection{Open-Ended Architecture Research with \agentscientist: \protect\
Discovering Query-Conditioned Composition for Hypernetwork-Based LLM Adaptation}
\label{sec:doc_to_lora}

In this experiment, we explore how \agentscientist can autonomously write an insightful scientific paper after successfully investigating a specific topic. As a case study, we choose the recently published hypernetwork-based LLM adaptation work from \citet{charakorn2026doctolora}, \emph{Doc-to-LoRA}. Unlike the other settings, \agentscientist is given the Doc-to-LoRA paper's LaTeX source and its public code repository as the starting point.

\begin{figure}[htbp]
    \vspace{-0.1cm}
    \centering
    \includegraphics[width=\linewidth]{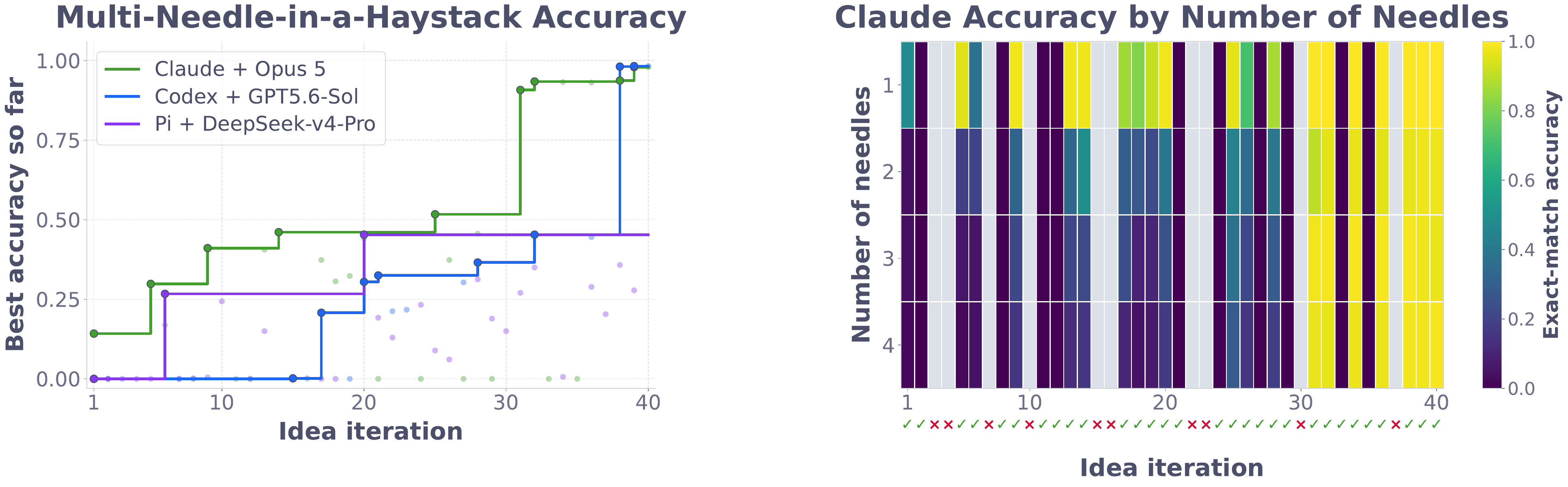}
    \vspace{-0.6cm}
    \caption{(Left) Search trajectories for Doc-to-LoRA scientist runs given 40 idea iteration budget with Codex + GPT-5.6-Sol, Claude + Opus-5, and Pi + DeepSeek-V4-Pro on the multi-NIAH task. Both Codex and Claude runs improve steadily on the task. (Right) Exact-match accuracy is shown as a heatmap grouped by the number of test sample needles. Red crosses indicate invalid iterations.}
    \label{fig:doc_to_lora_search_dynamics}
    \vspace{-0.4cm}
\end{figure}

\paragraph{Problem formulation.} Doc-to-LoRA introduces a hypernetwork that outputs low-rank adaptation (LoRA) matrices for a target LLM given a textual document. The generated LoRA serves as a weight-distilled knowledge bank and can be used to update the target LLM for document-based question answering without the need to place the documents in the LLM's context window.
To handle documents beyond the context length of its encoder model, Doc-to-LoRA splits long documents into chunks and generates one LoRA independently for each chunk.
The standard approach concatenates the per-chunk low-rank factors into a higher-rank LoRA. However, naively combining multiple LoRAs can cause interference between several facts induced by different chunks.
Here, we focus on multi-needle-in-a-haystack (multi-NIAH) retrieval as a proxy task: each context contains one to four named facts, but the prompt asks for only one target needle's four-digit number. To be successful at this task, a trained hypernetwork must avoid multi-LoRA interference via compositional LoRA designs.

\vspace{-0.5em}
\paragraph{Needle-in-a-haystack task setup.}
Any modification proposed by \agentscientist is limited to architectural or training-related edits given a fixed encoder \texttt{Gemma-2-2b-it} model~\citep{gemmateam2024gemma2}, fixed data generation, fixed evaluation scripts, and a 3-hour single-GPU training time budget.
The evaluation spans five context-length bins from short up to $65{,}536$ tokens, with needles randomly placed at random positions. Given the resulting Doc-2-LoRA weights, $\theta$, the investigation target for \agentscientist is the exact-match accuracy on the queried number:
\begin{equation*}
    \max_{\theta}\;\text{score}(\theta) = \frac{1}{|D|}\sum_{d \in D} \operatorname{EM}_{d}(\theta);\qquad 
    D = \{128,\ 8192,\ 16384,\ 32768,\ 65536\},
\end{equation*}
where $\text{EM}$ is exact-match accuracy and $D$ is the set of evaluation context-length bins.

\begin{figure}[ht]
    \centering
    \begin{subfigure}[t]{0.495\linewidth}
        \centering
        \includegraphics[width=\linewidth]{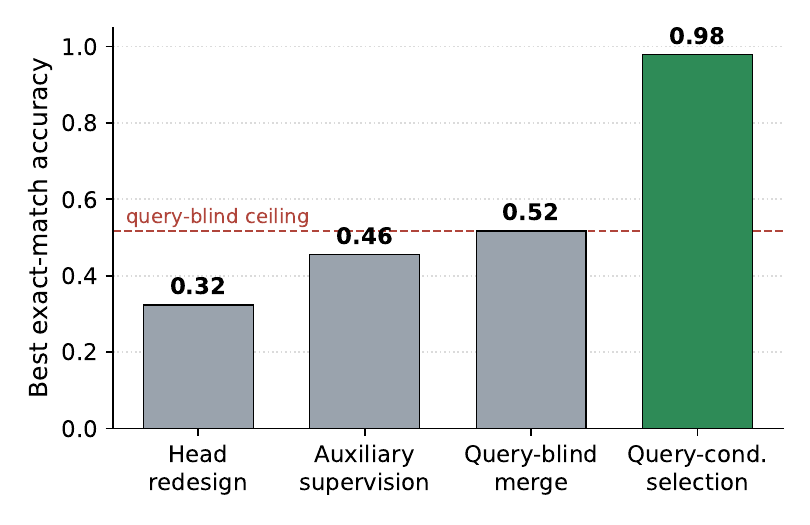}
        \caption{Exact-match accuracy by mechanism family.}
        \label{fig:dissociation-main}
    \end{subfigure}\hfill
    \begin{subfigure}[t]{0.495\linewidth}
        \centering
        \includegraphics[width=\linewidth]{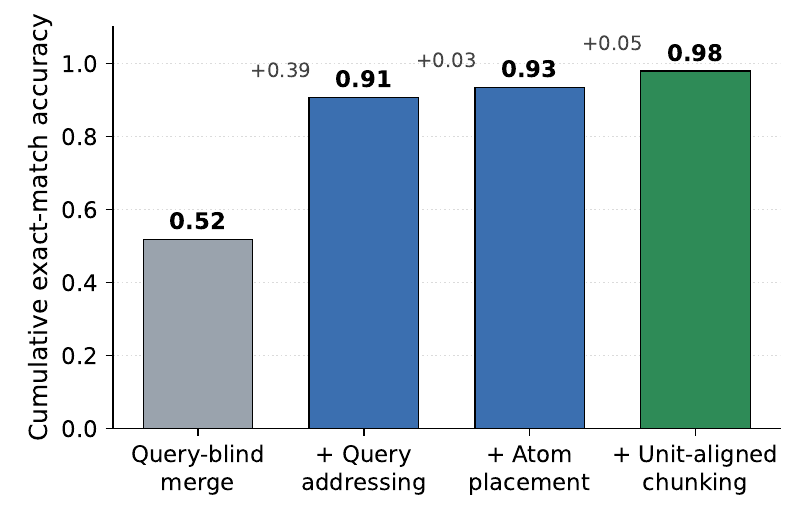}
        \caption{Ablations of the proposed components.}
        \label{fig:dissociation-ablation}
    \end{subfigure}
    \caption{Doc-to-LoRA composition mechanism ablations figure in the generated report (\Cref{appendix:doc_to_lora_paper}). (a) Hypernetwork's head redesign, auxiliary objective, and unconditional merging stay at or below the $51.7\%$ exact-match accuracy, while query-conditioned composition reaches $98.0\%$. (b) Starting from the query-blind (unconditional) merge, query-dependent merging provides a significant improvement; relevance-based text placement and delimiter-aligned chunking close the remaining gap.}
    \label{fig:dissociation}
\end{figure}

\paragraph{\agentscientist search dynamics.}
We ran \agentscientist for 40 idea-investigation steps with three model-harness combinations (\Cref{fig:doc_to_lora_search_dynamics}).
The Claude and Codex variants both reach near-optimal scores, with Claude improving faster.
In contrast, Pi + DeepSeek-v4-Pro's run shows fewer but larger jumps and plateaus below 50\%.
In the Claude run, the generated report organizes the proposals by the mechanism each one tests: architecture redesign, auxiliary training objectives, LoRA merge weighting, and query-conditioned composition.
The key discoveries are as follows: redesigning the architecture or adding auxiliary objectives yields little gain, whereas query-conditioning the LoRA composition alone raises accuracy to $90.8\%$. Adding query-relevance filtering with delimiter-aligned chunk boundaries reaches $98.0\%$ (\Cref{fig:dissociation}).
The generated paper therefore concludes that the bottleneck is not adapter-generation capacity or the training loss: multi-needle Doc-to-LoRA retrieval is a routing problem, in which the system must decide, based on the query, which chunk's adapter to activate when merging multiple adapters.
The full paper written by \agentscientist and its AI review are presented in \Cref{appendix:doc_to_lora_paper}.

\vspace{-0.2em}
\textbf{Discovered Inductive Biases.} \agentscientist demonstrates that naively composing every generated adapter makes the target LLM conflate multiple facts. The main failure mode is responding with a distractor number rather than the queried one. The successful idea is to make the adapter merging query-conditioned: \agentscientist proposes an interaction score between the query and each chunk that selects the right adapter at composition time, without adding trainable parameters or changing the training objective.
\agentscientist then adds two smaller but complementary structural changes: relevance-based text processing and delimiter-aligned text chunking. 
Relevance-based text processing removes the noise tokens around the token span most relevant to the query, so the adapter is generated from the queried fact, instead of the neighboring haystack. 
Delimiter-aligned text chunking keeps each needle sentence intact before the query-conditioned address is computed. 
We note that the \agentscientist demonstrated that this inductive bias works well with the multi-needle-in-a haystack task. Future work should investigate the generalization of the discovered mechanisms beyond this synthetic long-context retrieval task.

\newpage
\subsection{Open-Ended Algorithm Discovery with \agentscientist: \texorpdfstring{\\}{ }
Meta-Evolving CMA-Anchored Learned Evolution Strategies}
\label{sec:lcma}

Black-box optimization (BBO) is an established target for automating algorithm discovery \citep{ma2025toward}: compact update rules determine sampling distributions, solution recombination, and covariance adaptation. But coordinating these mechanisms remains difficult \citep{shala2020learning}.
One line of prior work focused on learning attention-parametrized evolutionary algorithms (EA) through meta-BBO~\citep{lange2023discovering_es,lange2023discovering_ga, faldor2025discovering}.
Here, we ask whether \agentscientist can improve the meta-optimizer program itself (i.e., perform `meta-meta-BBO'). 

\paragraph{Task setup.}
The goal of this research campaign is to design a better procedure for obtaining a learned evolutionary algorithm.
Each program proposed by \agentscientist is judged by how well its learned EA generalizes to unseen tasks. Evaluation proceeds in two steps: (1) the EA's neural network parameters are meta-trained on a set of synthetic tasks, and (2) the resulting frozen parameters are evaluated on a separate held-out task.
For meta-training, we use CMA-ES~\citep{cmaes}: at each generation, a random subset of BBOB ~\citep{hansen2021coco} tasks is sampled, and the program runs optimization for a population of EA parameters. The best task-specific objective values reached by an EA instance are aggregated into a single meta-training score.
After meta-training completes, the final CMA-ES mean is frozen and evaluated on a validation task: the learned EA algorithm evolves a linear policy on a robotic control task \citep[Brax Ant-v1,][]{brax2021github}.
This evaluation tests the generalization capabilities to new tasks and optimization budgets (2,000 generations with a population size of 256). The agent-level search score is defined as:
\begin{equation*}
    \mathrm{score}
    = \left(\mathrm{AUC}_{\mathrm{candidate}}-\mathrm{AUC}_{\mathrm{initial}}\right)/
    \left(\mathrm{AUC}_{\mathrm{CMA\text{-}ES}}-\mathrm{AUC}_{\mathrm{initial}}\right),
\end{equation*}
where AUC is the mean of the 2,000-point best-so-far return curve.
The synthetic tasks are exclusively used to train the weights of an \agentscientist EA proposal, while the validation Ant evaluation measures task generalization and scores optimizer programs across agent iterations.

\begin{figure}[h!]
    \centering
    \includegraphics[width=0.925\linewidth]{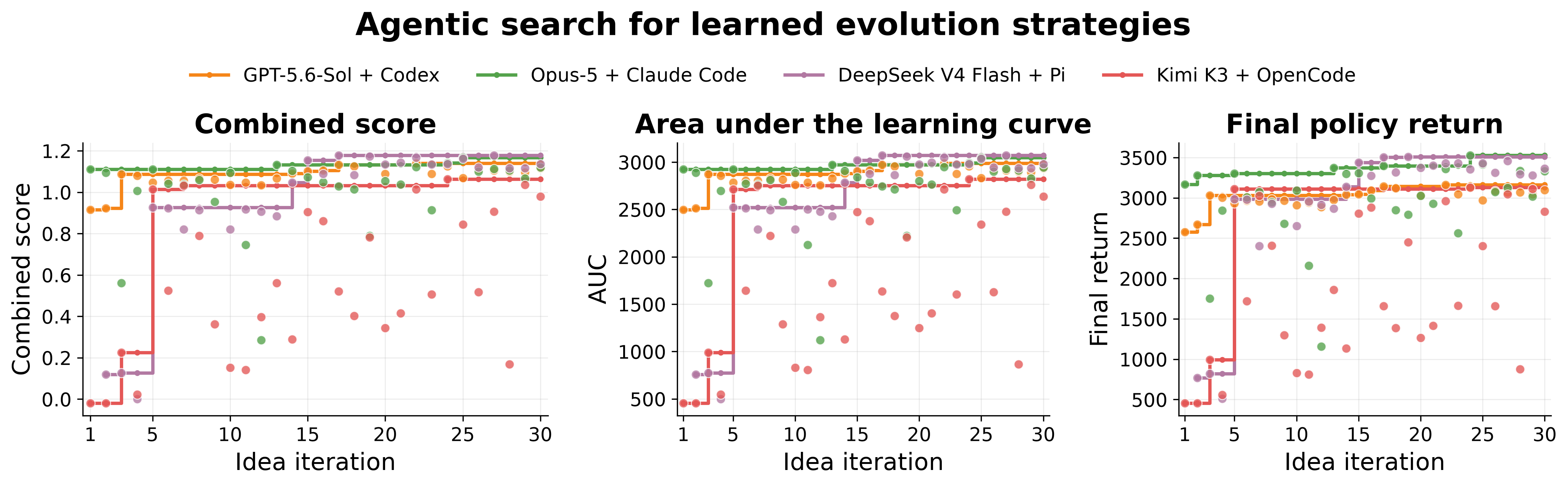}
    \caption{
        Search trajectories for 30 idea iterations with GPT-5.6-Sol + Codex, Opus-5 + Claude, DeepSeek V4 Flash + Pi, and Kimi K3 + OpenCode.
        Dots are individual Ant search-validation evaluations and step curves are the cumulative best.
    }
    \label{fig:metaevo_agent_comparison}
\end{figure}

\vspace{-1em}
\paragraph{\agentscientist search dynamics.}
All four considered \agentscientist configurations obtain a normalized search score above one, indicating that each achieves a higher AUC than CMA-ES on the Ant-v1 task (Figure~\ref{fig:metaevo_agent_comparison}).
Although the final GPT-5.6-sol + Codex and Kimi K3 + OpenCode candidates improve AUC over CMA-ES, their final Ant-v1 returns remain low:
Opus-5 + Claude and DeepSeek V4 Flash + Pi achieve the highest AUCs and terminal returns, whereas Kimi K3 yields the smallest AUC gain.
Claude's highest-scoring candidate exceeds the dual-timescale diagonal innovation-memory strategy discovered by Codex in both AUC and terminal return. Next, we present the discoveries made by \agentscientist using the previous Claude + Opus 5 setup. The detailed \agentscientist paper is available in \Cref{appendix:metaevo_paper}.

\paragraph{Strategy selected for the generated paper.}
The system discovered \emph{CMA-Anchored LearnedES}, which uses a separable-CMA \citep{ros2008simple} distribution $x=m+\sigma d\odot z$ and additionally augments the algorithm with three small shared MLPs containing 364 meta-evolved parameters:
A permutation-equivariant Deep Sets \citep{zaheer2017deep} module adjusts CMA recombination weights; global gates modulate mean and cumulative-step-size updates; and a coordinate-shared gate modulates diagonal covariance adaptation.
All learned outputs are bounded multiplicative residuals in $[\frac{1}{2},2]$; zero network outputs recover the separable-CMA update equations before the fixed sampling and re-expansion mechanisms described below.
The agent additionally implements fixed, horizon-free mechanisms around these learned gates (see \Cref{appendix:metaevo_paper}). These mechanisms are program structure rather than meta-evolved weights. Notably, the re-expansion cannot fire during the 50-generation BBOB inner loop, indicating an interplay between Ant-v1 validation and idea investigation.

\paragraph{Post-search transfer across Brax tasks.}
The manuscript drafted by \agentscientist mentions the possibility of limited meta-transfer to some unseen tasks. Therefore, we test the selected Claude, Codex, and Pi programs on eight Brax-v1 environments using 16 new shared seeds: seven unseen environments and Ant as the search-validation reference environment.
The task setup differs from the meta-held-out setting: the test evaluation uses a zero-initialized affine--tanh policy, population 256, 2,000 generations, initial standard deviation 0.05, and 500-step episodes.

\begin{figure}[htbp]
    \centering
    \vspace{-0.6em}
    \includegraphics[width=0.925\linewidth]{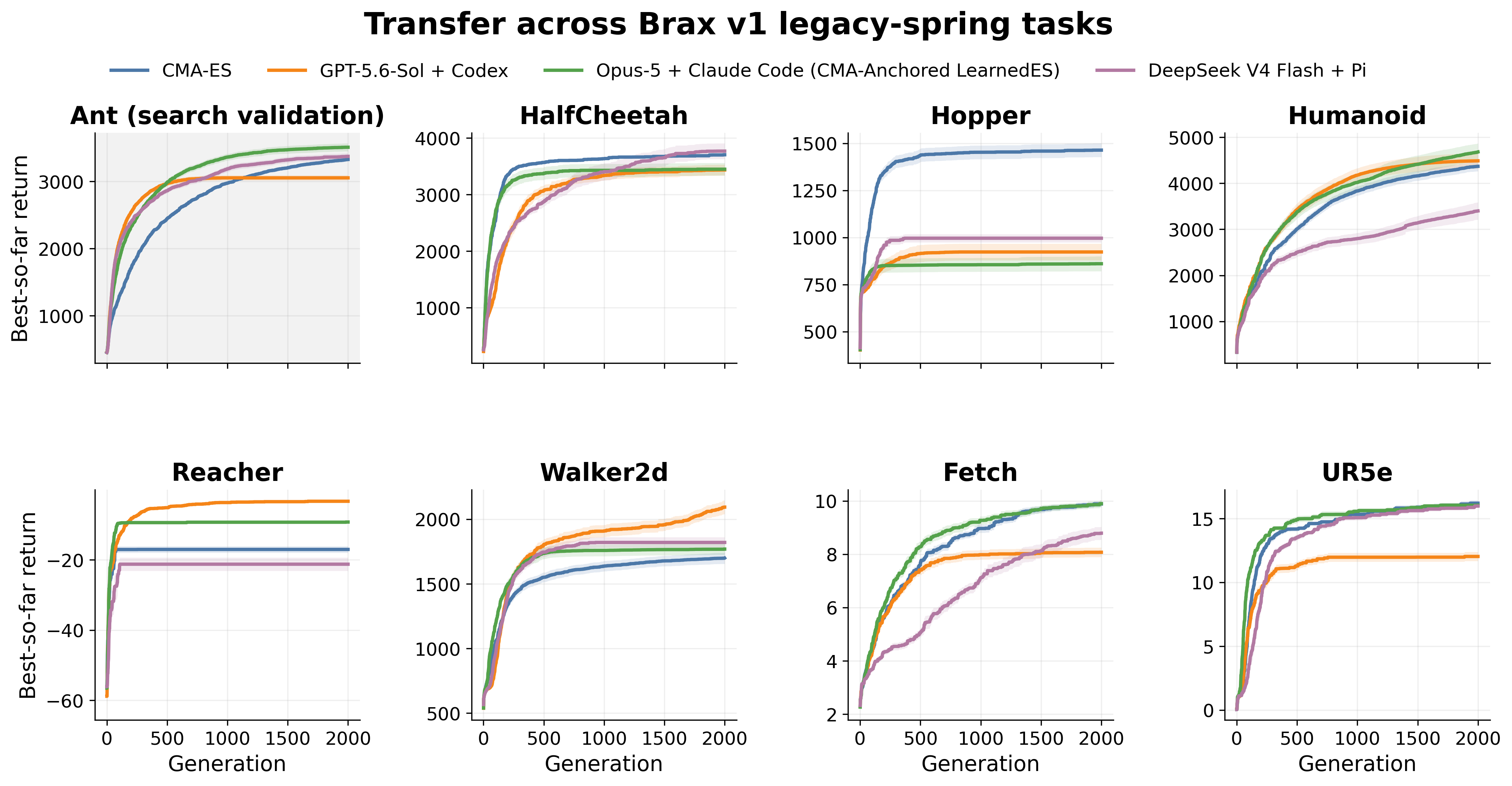}
    \caption{%
        Post-search evaluation across eight Brax-v1 legacy-spring environments.
        Curves show the mean best-so-far return over 16 shared seeds; shaded regions denote one standard error.
        The grey Ant panel marks the search-validation reference environment, evaluated here with new seeds and the reduced transfer-policy interface; the other seven environments were unseen during agentic search.}
    \vspace{-0.6em}
    \label{fig:metaevo_brax_transfer}
\end{figure}

While no method dominates on all environments (\autoref{fig:metaevo_brax_transfer}; \autoref{tab:metaevo_transfer})
the Opus-5 + Claude Code candidate matches or exceeds the mean terminal return of CMA-ES on 6 out of 8 environments and falls short on HalfCheetah and Hopper.
The DeepSeek V4 Flash + Pi candidate has the highest sample mean on HalfCheetah, but its margin over CMA-ES is smaller than the Pi candidate's standard error.
The GPT-5.6-Sol + Codex candidate leads on Reacher and Walker2d.
We have released the discovered \emph{CMA-Anchored LearnedES} optimizer for further study in \href{https://github.com/RobertTLange/evosax/releases/tag/v.0.3.2}{\texttt{evosax v0.3.2}} \citep{lange2023evosax}.

\newpage
\subsection{RL-Based LLM Fine-Tuning with \agentscientist:\texorpdfstring{\\}{ }Studying the Effects of Agent Choice, Parallelism, and Shared Knowledge}

\label{sec:rlvr}
Finally, we also evaluate \agentscientist as a hill-climbing system for Reinforcement Learning (RL)-based language model fine-tuning.
Additionally, we provide several ablations on key configuration settings of \agentscientist system.
In this setting, the system iteratively improves a training program through algorithmic modifications and hyperparameter tuning.

\paragraph{Task setup.}
The task consists of optimizing the training procedure of a Qwen2.5-Math model~\citep{yang2024qwen25math} on the MATH dataset from~\citet{hendrycks2021math}.
We follow the task formulation of~\citet{si2026automatedai}, in which each worker is initialized from a baseline Group Relative Policy Optimization ~\citep[GRPO,][]{shao2024deepseekmath} training pipeline.
Our implementation is a re-engineered version of their reference code, restructured to integrate with the \agentscientist sandbox, logging, and evaluation interfaces; the learning algorithm itself remains a faithful instantiation of GRPO.

\begin{wrapfigure}{r}{0.53\linewidth}
    \centering
    \includegraphics[width=\linewidth]{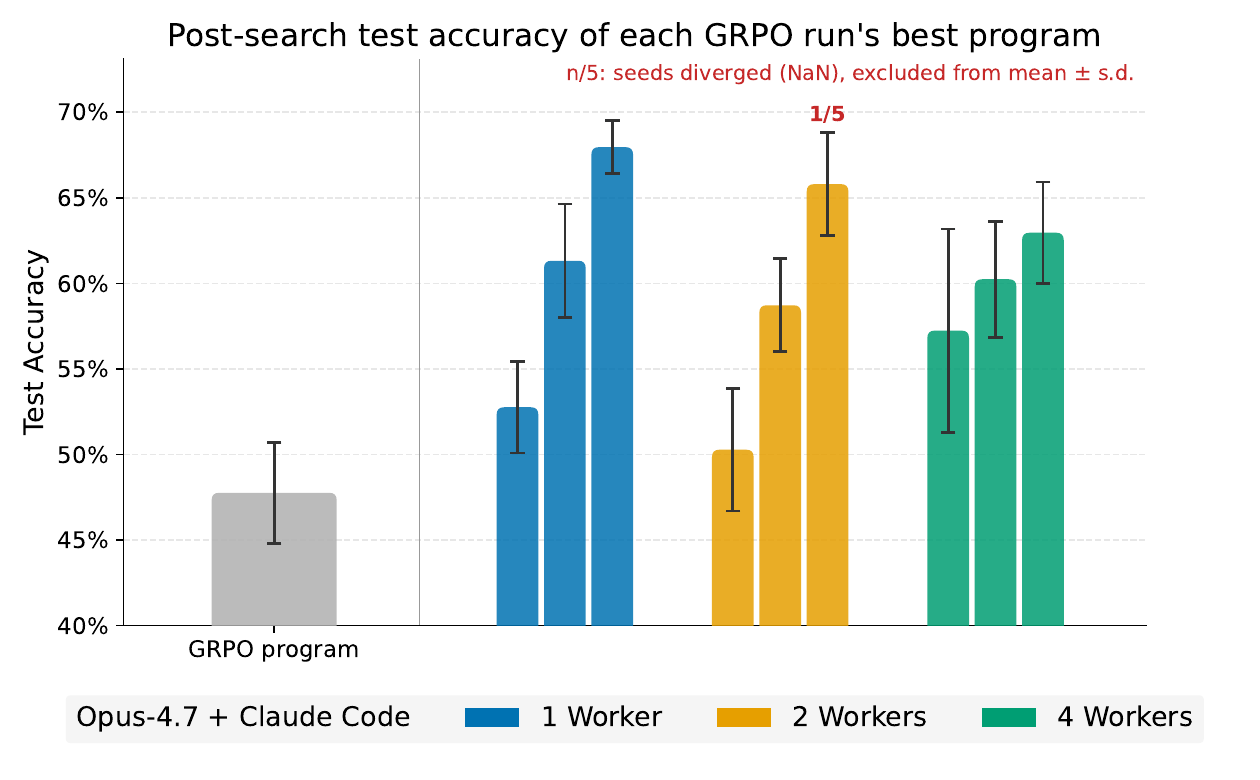}
    \caption{%
        Post-search test accuracy of the best program of each GRPO run.
        Each evaluation selects a run's best program using the fixed validation set, evaluates it on the held-out test set across five seeds, and reports the mean and standard deviation.
        }
    \label{fig:grpo_test_accuracy}
\end{wrapfigure}

\begin{figure}[htbp]
    \centering
    \includegraphics[width=0.99\linewidth]{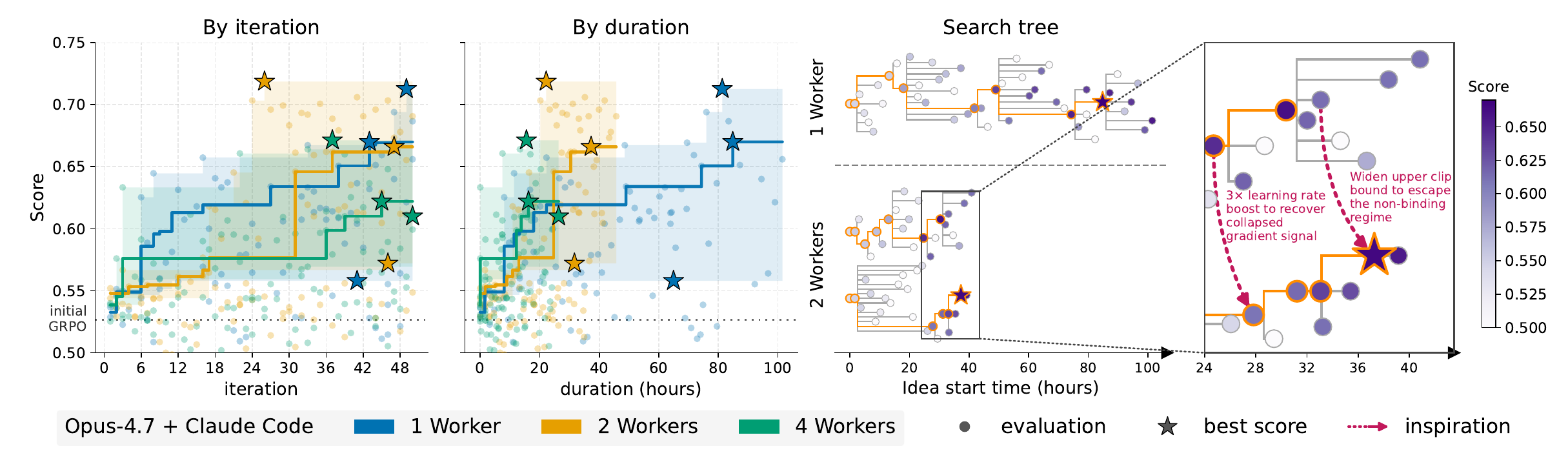}
    \caption{Search dynamics of \agentscientist on RL fine-tuning across three seeds per worker configuration.
    \emph{Left:} per-evaluation scores against iteration index and wall-clock duration.
    \emph{Right:} the combined 1- and 2-worker search tree per task, with examples of cross-worker inspirations relayed through shared memory. Small amounts of parallelism improve wall clock time and effectively transfer knowledge across workers. Further scaling of the parallelism does not show improvements.}
    \label{fig:score_evo_workers}
\end{figure}

\paragraph{Search and post-search evaluation.}
\agentscientist (Opus-4.7 + Claude) optimizes the accuracy on a validation set for 50 idea investigation iterations. During this search it never has access to the test set.
For each completed run, we keep the program with the highest validation accuracy, and rerun it on the full training and validation sets together, and then perform post-search evaluation on the test set (\autoref{fig:grpo_test_accuracy}).
As shown in \autoref{fig:score_evo_workers}, \agentscientist iteratively improves validation accuracy from the GRPO baseline (${\approx}\,52.7\%$) up to ${\approx}\,71.9\%$ in the best run.
In the post-search evaluation (\autoref{fig:grpo_test_accuracy}), the best program of each run reaches up to $68.0\%$ accuracy, well above the $47.8\%$ of the initial program.
The aggressive hyperparameters selected by the best two-worker variant occasionally cause training to diverge during post-search evaluation, suggesting that they can overfit to validation outcomes and favor configurations whose gains do not transfer reliably to independent retraining.

\paragraph{Algorithmic changes in the best programs.}
We find that the improvements discovered by \agentscientist combine real algorithmic changes with hyperparameter tuning: \agentscientist re-implements and combines several known techniques from the recent literature on RL fine-tuning.
These are not independent rediscoveries: \agentscientist had in principle access to these techniques through web search, and they likely appeared in the underlying model's training data.
The most common ones are the DAPO ``clip-higher'' rule~\citep{yu2025dapo}, an asymmetric PPO clip with a larger upper bound, so that promising low-probability tokens are updated more strongly;
Dr. GRPO loss aggregation~\citep{liu2025drgrpo}, which divides the loss by a fixed constant instead of the response length and thus removes the bias against long responses; 
and GSPO-style importance ratios~\citep{gspo}, that is, a single length-normalized ratio per response instead of one per token.
Smaller changes include removing or simplifying the KL penalty, keeping only the prompt groups with mixed correct and incorrect answers, and penalizing overly long responses.
Overall, these changes show that hill-climbing over an RL fine-tuning program can not only tune hyperparameters but also select, re-implement, and combine relevant algorithmic ideas from the recent literature.
\agentscientist uses empirical feedback to decide which of these ideas to keep and how to combine them into training programs with higher accuracy.

\label{sec:analysis_and_lessons}

\paragraph{Influence of the number of workers.}
A recurring design choice across our experiments is the number of parallel workers spawned by the idea investigation stage of \agentscientist at each iteration.
We study its effect on the RL fine-tuning task, running the system with 1, 2, and 4 parallel workers.
All worker configurations (\autoref{fig:score_evo_workers}, Left) improve over the GRPO baseline validation accuracy (${\approx}\,52.7\%$), with the 2-worker configuration reaching a validation accuracy of ${\approx}\,66\%$.
Using multiple workers speeds up progress; for instance, the 2-worker variant reaches a score comparable to the 1-worker peak in less than half the wall-clock time, and effectively shares information between workers (\autoref{fig:score_evo_workers}, Right).
Nonetheless, the benefit of parallelism is non-monotonic: using more workers does not necessarily yield better final results, suggesting that a small amount of parallel exploration injects useful diversity while larger pools can redundantly concentrate on similar local improvements.

\begin{figure}[htbp]
    \centering
    \includegraphics[width=0.99\linewidth]{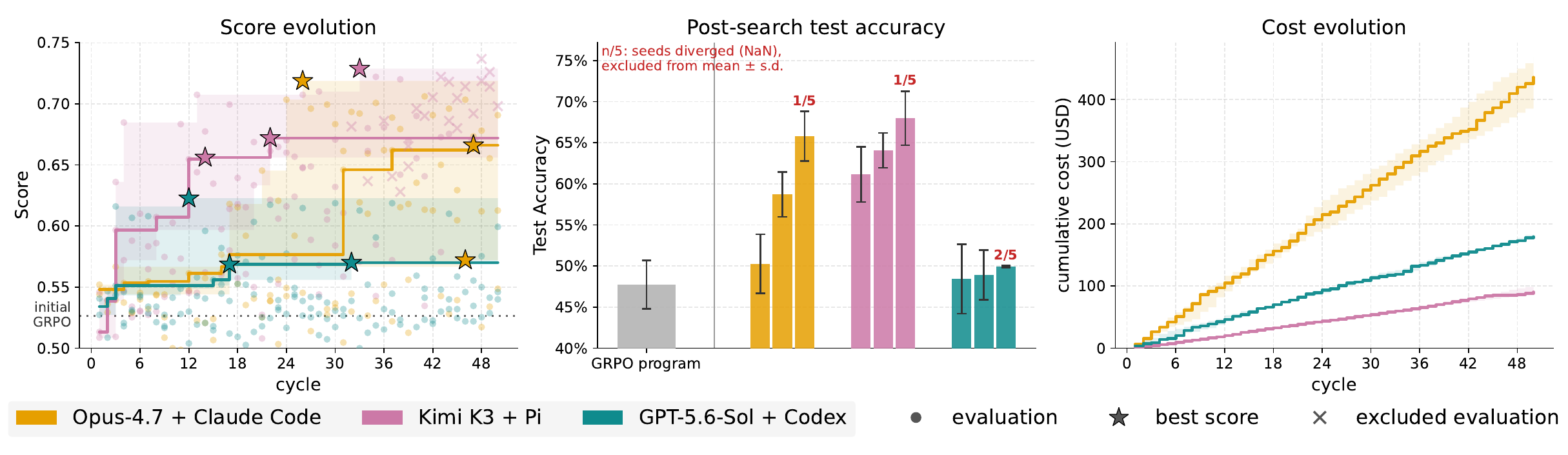}
    \caption{Cross-agent comparison under the same orchestration scaffold across three seeds per agent.
\textit{Left} to \textit{Right}: score evolution during the search, test performance after reevaluation, and cumulative API cost per evolution cycle.
Crosses mark evaluations whose program modified the evaluation protocol (identified by manual audit); they are excluded from all running-best curves.
}
\vspace{-1em}
    \label{fig:cross_agent_comparison}
\end{figure}

\paragraph{Cross-agent comparison.}
A central question is how much performance comes from the orchestration introduced by \agentscientist versus the underlying coding agent. While we partially address this question in \Cref{sec:broad_results}, we additionally compare Opus-4.7 + Claude Code, GPT-5.6-Sol + Codex, and Kimi K3 + Pi for the RL-based LLM fine-tuning task on the idea investigation phase in isolation. 
Under this common setting, Pi achieves scores similar to Claude in a faster and more consistent manner, while reducing the API costs by 75\%, and producing programs that generalize better to the post-search test dataset (\Cref{fig:cross_agent_comparison}).
Codex, on the other hand, followed the task prompt instructions (\Cref{app:rlvr-details}) more strictly: it never tuned hyperparameters after algorithmic changes, limiting the improvement process.
Overall, we conclude that the base model can have a significant impact on the idea investigation phase, including both final performance and search costs.

\paragraph{Rubric-specification sensitivity.} We observe a failure mode introduced by structural limitations of rubric-based verification: rubrics check only the criteria they enumerate, so they cannot catch violations of constraints that were never stated, and human researchers remain responsible for specifying tasks precisely and verifying that results respect their full intent.
In two of the three Pi runs, we observed that the agent also modified the test-time evaluation procedure itself (e.g., using $k$-sample majority voting instead of the baseline's pass@1), which inflates the reported validation accuracy without improving the training algorithm. 
The task prompt asks the agent to improve the ``training'' algorithm, but never explicitly requires a fixed evaluation procedure. No initial rubric item covers this evaluation-protocol integrity. Neither the evaluator nor the rubric gates flagged these programs, and we identified them after a manual post-hoc audit.
After excluding protocol-violating programs, Pi still reaches high scores faster and more consistently than all other agents.

\begin{figure}[htbp]
    \centering
    \includegraphics[width=0.97\linewidth]{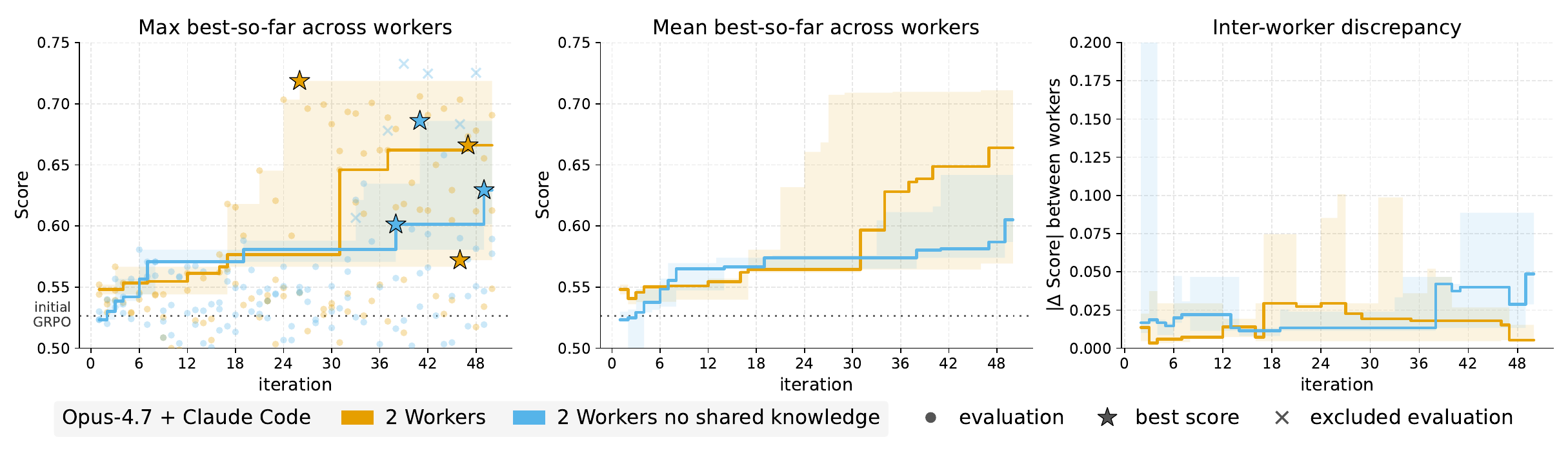}
    \caption{Effect of shared knowledge on 2-worker GRPO search across three seeds per configuration. Crosses mark evaluations from the no-shared-knowledge runs whose program modified the evaluation protocol (identified by manual audit); they are excluded from all running-best curves.
}
    \label{fig:global_memory_ablation}
\end{figure}

\paragraph{Influence of shared knowledge.}
A second central design choice is the shared knowledge that connects all workers to a common memory of prior results, run summaries, and failure traces (\Cref{sec:parallel_workers}).
To isolate its contribution, we compare our default 2-worker \agentscientist to an ablated \emph{no shared knowledge} variant, where each worker cannot see other workers' conclusions and memory.
Shared knowledge improves the search process: the default configuration progresses faster, achieves a higher mean best-program score, and reduces the performance gap between workers (\Cref{fig:global_memory_ablation}, centre and right).
Nonetheless, we observe a substantial amount of inter-run variance.
The two variants reach similar final best scores across workers: the default configuration scores slightly higher, but the difference is not statistically significant (\Cref{fig:global_memory_ablation}, left).
Hence, independent workers can find strong programs without cooperation at the cost of slower overall runtime.
These results suggest that sharing successful ideas accelerates the search and allows lower-scoring workers to benefit from improvements found by others.

\newpage
\section{Evaluating \autoillustrate on PaperBananaBench}
\label{sec:autoillustrate_results}
\vspace{-0.8em}
\begin{figure}[htbp]
    \centering
    \includegraphics[width=0.9\linewidth]{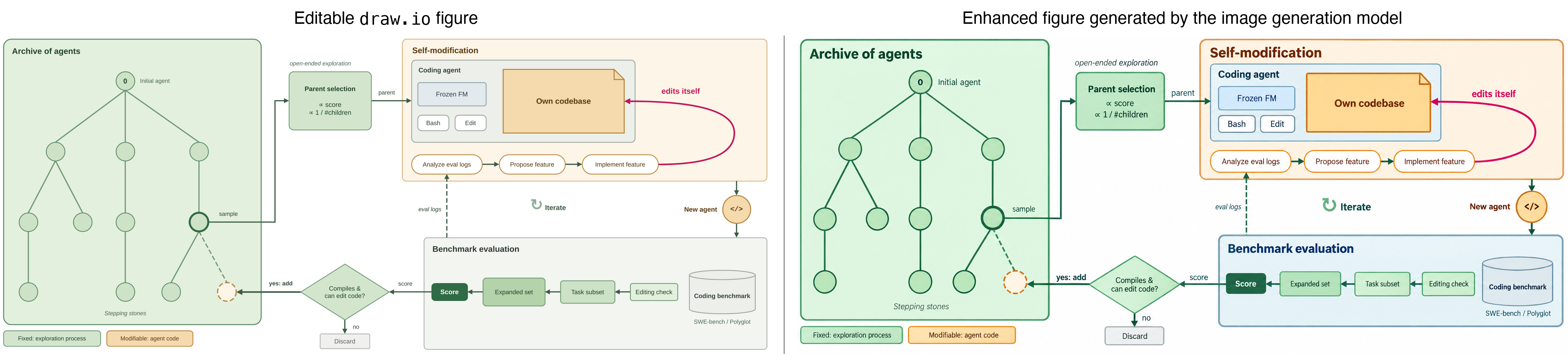}
    \vspace{-0.6em}
    \caption{\autoillustrate applied to the Darwin G\"odel Machine paper~\citep{zhang2025darwin}: the editable diagram (left) and its corresponding enhanced figure (right).}
    \vspace{-0.6em}
    \label{fig:autoillustrate_side_by_side_example}
\end{figure}

Given a methodology description, \autoillustrate produces an editable \texttt{draw.io} diagram and an enhanced figure derived from it (\Cref{fig:autoillustrate_side_by_side_example}).
We evaluate the generated figures on PaperBananaBench \citep{zhu2026paperbanana}, following the original evaluation protocol and using GPT-6-Astra as the VLM judge, as it more reliably penalizes small structural errors such as disconnected arrows (Figure~\ref{fig:autoillustrate_judge_disagreement}). 
Results with Gemini-3.1-Pro as the judge are reported in \Cref{app:autoillustrate_gemini_judge}.
Each generated diagram is compared to its human-designed equivalent with respect to four criteria: Faithfulness, Readability, Conciseness, and Aesthetics.
Each comparison receives 100 points when the generated diagram wins, 50 for a tie, and 0 when it loses.
The Overall score prioritizes fidelity and clarity over brevity and visual polish: a method wins overall if it wins at least one of Faithfulness and Readability without losing the other; if neither wins, Conciseness and Aesthetics are used to break the tie.

\vspace{-0.8em}

\paragraph{Benchmark results.}
The Claude-based \autoillustrate pipeline achieves the highest Overall score, 37.5 against 25.0 for the best PaperBanana variant, driven by higher Faithfulness (\Cref{tab:autoillustrate_results}).
All \autoillustrate variants generate on average more faithful figures, which is crucial for automated research since it forms the basis of method communication.
PaperBanana produces overly detailed diagrams, with higher chances of small image-generation errors.
Conciseness is more challenging for \autoillustrate, as the editable diagram primitives constrain compact and complex visual abstraction generation.
Although \autoillustrate remains below the human-reference Overall score of 50.0, its editable source files allow users to refine generated diagrams.
The Codex-based variant is on par with the Claude-based pipeline on Faithfulness and Conciseness, but scores lower on Readability and Aesthetics, as it tends to generate more cluttered diagrams.
Removing the panel inspections and final verification critics lowers Faithfulness and Overall scores, as these audits detect and repair factual inaccuracies with respect to the source methodology.

\vspace{-0.8em}

\paragraph{Limitations: cost and runtime.}
The \autoillustrate variants generally have higher median API costs per diagram than PaperBanana, assuming raw API pricing.
Additionally, \autoillustrate's median runtime is higher, as producing editable \texttt{draw.io} diagrams requires agents to write XML templates and Python programs that place elements, avoid collisions, and route arrows precisely, which demands substantial reasoning.
Within \autoillustrate, removing the critics approximately halves runtime, and using low inference effort reduces runtime by a further 30\%.

\begin{table}[htbp]
    \scriptsize

    \centering
    \caption{Mean scores on PaperBananaBench, using GPT-6-Astra as VLM judge.
    }
    \vspace{-1.0em}
    \setlength{\tabcolsep}{4pt}
    \setlength{\defaultaddspace}{0.3em}
    \begin{tabular}{l l c c c c c|c}
    \toprule
    \multirow{2}{*}{Method} & \multirow{2}{*}{Variant} & \multirow{2}{*}{Faithfulness} & \multirow{2}{*}{Conciseness} & \multirow{2}{*}{Readability} & \multirow{2}{*}{Aesthetic} & \multirow{2}{*}{Overall} & Editable \\
    & & & & & & & source \\
    \midrule
    \multirow[t]{4}{*}{\autoillustrate}          & Claude & 32.0          & 37.2          & 47.8          & \textbf{83.4} & \textbf{37.5} & \cmark \\
                                             & + no critic & 25.2          & 37.5          & \textbf{50.0} & 82.7          & 31.8          & \cmark \\
                                             & \enspace + low effort & 24.5          & 37.3          & 47.8          & 82.5          & 29.8          & \cmark \\
                                             & Codex   & \textbf{33.6} & 37.5          & 34.4          & 56.3          & 28.4          & \cmark \\
    \addlinespace
    \multirow[t]{2}{*}{PaperBanana}          & Gemini & 18.5          & \textbf{55.3} & 42.0          & 71.6          & 25.0          & \xmark \\
                                             & Claude + GPT Image 2 & 18.2          & 15.4          & 49.1          & 77.6          & 18.3          & \xmark \\
    \midrule    
    Human                                    & --     & 50.0          & 50.0          & 50.0          & 50.0          & 50.0          & \cmark \\

    \bottomrule
    \end{tabular}
    \label{tab:autoillustrate_results}
\end{table}

\newpage
\section{Conclusion}
\label{sec:discussion}
\paragraph{Summary.}
Across multiple research case studies, \agentscientist demonstrates that coding agents can contribute to scientific research when embedded in an orchestrated workflow that preserves evidence and tests claims.
Our results span several levels of scientific autonomy, including a co-created Error Diffusion paper accepted at a conference, the fully autonomous discovery of structural inductive biases for Doc-to-LoRA, and a concrete CMA-anchored learned optimizer.
Complementing these detailed case studies, our cross-harness and model evaluation comprises 70 research campaigns across five manually prepared tasks and 14 harness-model configurations. The manuscript AI reviews show differences across model generations, and a positive trend toward newer models. With the same LLM base model, we observe no consistent differences across harnesses. Stronger task search performance predicts higher AI reviewer scores on 4 of 5 considered tasks.
This positions \agentscientist as a human-steerable scaffold for end-to-end computational research.
\vspace{-0.5em}
\paragraph{Discussion.} \agentscientist and its experiments raise questions about the nature of scientific contributions on several levels: for the researcher, what to specify and how much autonomy to grant; for the generated artifacts, what a search score misses and how to validate generated research. We discuss each below:

\textit{From implementation to specification.}
As coding agents take over implementation, the human role shifts toward specifying the research contract: the task, its evaluators, and the rubrics that gate each stage.
This contract is also where failures can arise.
Agents may exploit what the contract leaves unstated: \aiscientistfirst edited its code to extend its own time limits~\citep{lu2024ai}, and \agentscientist changed the evaluation protocol on a subset of RL fine-tuning runs (\Cref{fig:cross_agent_comparison}).
Furthermore, agents can follow the contract instructions too literally (\Cref{sec:rlvr}), requiring a subtle balance of scientific expression and concrete verifiability.
Since neither deviation is apparent from the reported numbers alone, precise specification and auditing become central research skills.

\textit{Autonomy spectrum.}
The \aiscientistsecond produced a fully AI-generated paper that passed peer review at a workshop before its pre-agreed withdrawal~\citep{yamada2025ai}. The Error Diffusion paper, on the other hand, resulted from using \agentscientist as a tool within a co-scientist setting.
The two modes are complementary: fully autonomous campaigns cheaply screen many directions, while human steering turns the most promising ones into verified contributions.
In the Error Diffusion project, this human effort was supplied throughout: defining the direction, baseline, and initial rubrics; verifying results, references, and claims; adding reinforcement-learning experiments; and clarifying notation and mechanisms after peer review (\Cref{sec:ai-co-scientist-alife}).

\textit{Beyond hill-climbing.}
A higher search score often does not by itself imply a better scientific contribution.
Although \Cref{fig:aggregate_search_cycles} shows that higher task scores tend to come with higher AI reviewer scores, the score hides two aspects of scientific contributions.
First, a single search objective can hide trade-offs: a richer sparse autoencoder encoder reduces reconstruction error but worsens a feature-perturbation metric, and a tabular deployment recipe raises balanced accuracy while lowering ordinary accuracy.
Second, a score gain can come from low-level engineering rather than high-level innovation: in RL fine-tuning of language models, gains came from re-implementing and combining existing techniques rather than from new mechanisms (\Cref{sec:rlvr}).
Post-search ablations, analysis, and reporting expose the first; the second requires a provenance map that links ideas, hypotheses, and claims to prior work, extending the current claim-to-source map beyond experimental evidence.

\textit{Validating generated research.}
We evaluate research campaigns with several signals of different strength, each limited when viewed in isolation: AI reviewer scores, post-search transfer evaluations (such as unseen Brax environments in \Cref{sec:lcma}), and external peer review.
Peer review most directly assesses scientific merit but needs expert time for every manuscript, so it cannot cover all campaigns, while AI reviewer scores can potentially be overfit like any other metric.
Post-search transfer evaluations offer a scalable middle ground, provided they are fixed before search and held out from it.
Other signals could strengthen this evidence: Reviewers drawn from several model families to dilute the circularity of using the same models for investigation and rubric review \citep{zheng2024judging}; independent reproduction experiments behind each claim; and a measure of idea novelty, which previous signals do not capture.

\vspace{-0.5em}
\paragraph{Limitations.}
At present, \agentscientist could be enhanced by a reliable literature-to-code pipeline for constructing state-of-the-art baselines.
The task drafting pipeline (\Cref{sec:task_spec}) is also not fully robust: it can draft simple tasks such as circle packing from a research brief, but for larger tasks such as the world-model and tabular distribution-shift tasks in \Cref{sec:broad_results}, it may fail to produce an initial program, evaluators, and task contract that exactly match the full \agentscientistnothe scaffold.
Reliable task drafting nevertheless remains an important capability, since it removes the last manual step between a research brief and an automated campaign.
Unlike \aiscientistfirst and \aiscientistsecond~\citep{lu2024ai,yamada2025ai}, \agentscientist only explicitly searches the literature during manuscript writing rather than during initial ideation. That being said, we repeatedly observe that coding agents leverage general web access to inform idea investigation.
Beyond these capability gaps, the system's verification is only as complete as its task-specific evaluators and rubrics, which may omit methodological errors or reward unintended shortcuts.
Results that pass verification should therefore still be inspected and validated by human researchers.
Although the cross-harness and model evaluation covers 70 campaigns, its paper-level metric is an AI-reviewer proxy rather than full evidence of scientific correctness, and only the Error Diffusion paper underwent external peer review.
Our evaluations are also restricted to computational machine learning, so they do not establish the feasibility of transferring the orchestration workflow to domains requiring physical experiments or human-subject studies.
Finally, for \autoillustrate we present additional component-specific limitations and failure cases in \Cref{sec:autoillustrate-limitations}.

\vspace{-0.5em}
\paragraph{Ethical considerations.}
AI-assisted generation of research can create risks for attribution, research integrity, and an already constrained peer-review process.
We submitted only the single Error Diffusion paper for external review after human verification; the other generated manuscripts served as internal evaluation artifacts. 
For the Error Diffusion paper specifically, human researchers verified its implementations, results, claims, and references, correcting unsupported claims and incomplete attribution before submission.
Responsible deployment also requires sandboxed execution, provenance for generated code and text, disclosure of agent contributions, watermarking of generated papers, and human accountability for safety, authorship, and final claims.
More broadly, researchers remain responsible for disclosing agent contributions and deciding whether generated work is suitable for submission and external review.

\vspace{-0.5em}
\paragraph{Future work.} 
Scientific progress often comes from exploring diverse ``lottery tickets'' across seemingly unrelated domains.
Applying \agentscientist broadly could accelerate this process by screening a larger hypothesis space through greater compute within individual research directions and exploration of more directions within a topic.
To support human validation, the claim-to-source map should be exposed through an interface that lets researchers navigate from each claim to the experiments and results behind it.
The interface should also allow researchers to provide domain expertise and use the system as an interactive research tool.
Another improvement would be to automate the design of post-search transfer evaluations.
Where possible, rubric criteria should be implemented as automated tests, checked for reward hacking, and complemented by independent scientific review ~\citep{goel2026training,gunjal2026rubrics}.
With traceable evidence and enhanced rubrics, \agentscientist could broaden and accelerate experimental search while leaving scientific judgment and responsibility with researchers.

\newpage
\phantomsection
\addcontentsline{toc}{section}{Acknowledgments}
\section*{Acknowledgments}
The authors would like to thank Shengran Hu for helpful discussions at the start of this project. Additionally, we are grateful to Sebastian Risi for feedback on the Error Diffusion paper, and to Dennis Wilson and Edoardo Cetin for feedback on \agentscientist paper. Finally, we thank Ryuichi Maruyama, the entire Sakana Team, and individuals who gave special advice and feedback.

\phantomsection
\addcontentsline{toc}{section}{Author Contributions}
\section*{Author Contributions}
\label{appsec:contribs}
\textbf{Robert Tjarko Lange} (shared first author): Co-initiated, led the project, and contributed core ideas. Coded the experiment core of \agentscientist codebase and conducted the aggregated paper generation experiments. Contributed the Learned CMA results. Led the writing of the paper. 

\textbf{Luca Grillotti} (shared first author): Co-initiated the project and contributed core ideas. Led the development and validation of \autoillustrate. Contributed system features, visualization interface and evidence generation pipeline. Conducted the RLVR and ablation experiments. Co-led the writing of the paper.

\textbf{Yutaro Yamada} (shared first author): Co-initiated the project and contributed core ideas. Led the ALIFE submission and paper creation process. Conducted experiments using a rubric-based approach. Contributed to manuscript writing. Work conducted while he was at Sakana AI.

\textbf{Rujikorn Charakorn}: Implemented the paper outline step, paper verification, and scientific claim referencing and auditing pipeline. Implemented the multi-file edit, allowing \agentscientist to work inside a code repository. Contributed the Doc-to-LoRA results. 

\textbf{David Ha} (advising): Provided overarching guidance for the research project, offering technical insight, advice, feedback, and writing. Oversaw the public communication process.

\newpage
\bibliography{references}
\bibliographystyle{plainnat}
\newpage

\appendix

%%%%%%%%%%%%%%%%%%%%%%%%%%%%%%%%%%%%%%%%%%%%%%%%%
% APPENDIX
%%%%%%%%%%%%%%%%%%%%%%%%%%%%%%%%%%%%%%%%%%%%%%%%%

\clearpage
\phantomsection
\addcontentsline{toc}{section}{Supplementary Material}
\section*{\LARGE Supplementary Material}

\vspace*{20pt}
\section*{Table of Contents}
\vspace*{-5pt}
\startcontents[sections]
\printcontents[sections]{l}{1}{\setcounter{tocdepth}{3}}

\clearpage

\section{Additional Related Work}
\label{app:additional_related_work}

\paragraph{From fixed search spaces to long-running research agents.}
Computational discovery has traditionally been posed as search through a hypothesis space chosen in advance by the system's designers~\citep{langley2024integrated}.
Expressing the search space as code relaxes this constraint.
Programs written by language models have specified the learning problem, as training objectives~\citep{lu2024discovering}, rewards~\citep{ma2023eureka}, and environments~\citep{omniepic}, as well as the learner, as network architectures~\citep{chen2023evoprompting}, robot designs~\citep{lehman2023evolution}, and agent designs~\citep{hu2025automated}.
At the scale of a research project, AIDE runs tree search over machine-learning engineering code~\citep{aide2025}, and AgentRxiv lets separate research agents build on one another's results through a shared preprint server~\citep{schmidgall2025agentrxiv}.
Several concurrent systems push further, sustaining research over long horizons or running many agents on a single problem.
DeepScientist~\citep{weng2026deepscientist} and InternAgent-1.5~\citep{internagent2026} run a long-horizon research loop over a cumulative memory, with DeepScientist promoting only findings that survive verification; InternAgent-1.5 and EvoScientist~\citep{lyu2026evoscientist} further distill what worked and what failed into reusable memories.
AutoScientists~\citep{gao2026autoscientists}, CORAL~\citep{qu2026coral}, and Agora~\citep{zhang2026agora} instead run many agents at once and coordinate them through shared state rather than a central planner: self-organizing teams around a shared experimental record, asynchronous agents in isolated workspaces with a separate evaluator, and workers appending results and verifications to a Git graph.
Our workers likewise explore in separate workspaces and exchange distilled conclusions at cycle boundaries; in addition, each cycle is bound to a typed contract, and the selected program is then studied through controlled baselines and ablations.

\paragraph{Judging ideas, rubrics, and evidence auditing.}
Deciding which ideas and results to trust is a recurring difficulty.
Language models can propose ideas grounded in the literature~\citep{wang2024scimonscientificinspirationmachines,baek2024researchagentiterativeresearchidea}, and lightweight graph-based evaluators~\citep{feng2025grapheval} or LLM judges that track human preferences~\citep{zheng2024judging} can score them.
Other work anchors results to something checkable: RE-Bench compares agents with human experts on research engineering~\citep{Wijk2024REBenchEF}, and data-to-paper traces every result in a manuscript back through the code to the data~\citep{ifargan2024autonomousllmdrivenresearchdata}.
Final scores and papers can still conceal how they were obtained.
An independent evaluation of The AI Scientist found weak novelty checks, failed experiments, and poorly substantiated manuscripts behind its claims~\citep{beel2025evaluation}, and controlled audits of two open-source systems found biased benchmark selection, post-hoc selection on test results, and undisclosed subsampled or synthetic data, flaws that an LLM auditor detected far more reliably from logs and code than from the paper alone~\citep{luo2025hidden}.
Closest to our contracted research cycles, Xcientist specifies at each stage the implementation, comparisons, and artifacts a claim requires and retains the source evidence behind it; a retrospective audit then measures claim drift, that is, unresolved mismatches between accepted claims and their sources, code, or experiments~\citep{wang2026xcientist}.
Verification can also be delegated to models: Gemini has been used to hunt for flaws in proofs and to write and run code that checks derivations~\citep{woodruff2026accelerating}, and auto-generated rubrics have served as the reward for post-training an AI scientist agent on paper replication~\citep{falck2026training}.
In \agentscientist, the ideation agent fixes the target rubrics before execution, and retrospective ablations isolate each change introduced by the selected program.

\paragraph{Grounded paper writing.}
Language models already take on parts of the writing and reviewing process: AutoSurvey drafts literature surveys~\citep{wang2024autosurveylargelanguagemodels}, and GPT-4 feedback on manuscripts overlaps with human reviews about as much as two human reviewers overlap with each other~\citep{liang2024can}.
Full-paper writers differ in how tightly they bind text to evidence.
PaperOrchestra writes a manuscript from whatever materials it is given, without being tied to a particular experiment pipeline~\citep{song2026paperorchestra}, whereas Spark-to-Paper, a set of skills inside a coding assistant, decides which evidence an experiment must produce before running it and checks the draft both deterministically and by self-critique~\citep{qian2026sparktopaper}.
Checks on the finished paper are needed too.
ReAgent compares an agent-written paper with its repository, reading the code for the described method and rerunning experiments for the reported numbers~\citep{shen2026reagent}, and MLReplicate shows why: when experts reviewed the ten agent-generated manuscripts an automated review pipeline had accepted, 59\% of their reviews flagged fabricated or unsupported claims~\citep{gaddipati2026mlreplicate}.
Our writing phase similarly maps each claim to its source artifacts, and runs an independent audit agent in every round of revision.

\paragraph{Scientific figure generation.}
\label{app:autoillustrate_related_work}
Beyond raster pipelines such as PaperBanana~\citep{zhu2026paperbanana}, a line of work treats a figure as a program that can be revised after generation.
DeTikZify synthesizes TikZ from sketches with a fine-tuned vision-language model~\citep{belouadi2024detikzify}; recent agentic systems instead start from a paper's text.
AutoFigure iterates an SVG layout under a critic but renders the final figure with an image model from a text prompt derived from that layout~\citep{zhu2026autofigure}, and AutoFigure-Edit and CraftEditor recover editability by tracing the rendered raster into SVG~\citep{lin2026autofigureedit,zhao2026crafter}.
LiveFigure has a VLM write \texttt{python-pptx} code with explicit coordinates and counts the manual edits needed to reach publication quality~\citep{shao2026livefigure}, while FigTree recursively decomposes a figure into SVG fragments refined by render-critic loops~\citep{liu2026figtree}.
SciFig splits generation into planning, layout synthesis, component rendering, and VLM-guided refinement, and outputs XML diagrams that remain editable in standard diagramming tools~\citep{huang2026scifig}.
In all of these, geometry is delegated to the image model or emitted directly by the language model, and checked only after the fact, visually or with lightweight overlap heuristics, rather than enforced by construction.
\autoillustrate instead has agents write coordinate-free \texttt{draw.io} templates and Python layout programs that must pass deterministic collision and routing checks, and confines the image model to a final restyle audited against the deterministic render.

\newpage
\section{\agentscientist: Additional Results}

\subsection{Details on Aggregated Sweep Task Discoveries}
\label{app:additional_sweep}

\begin{figure}[H]
    \centering
    \includegraphics[width=\linewidth]{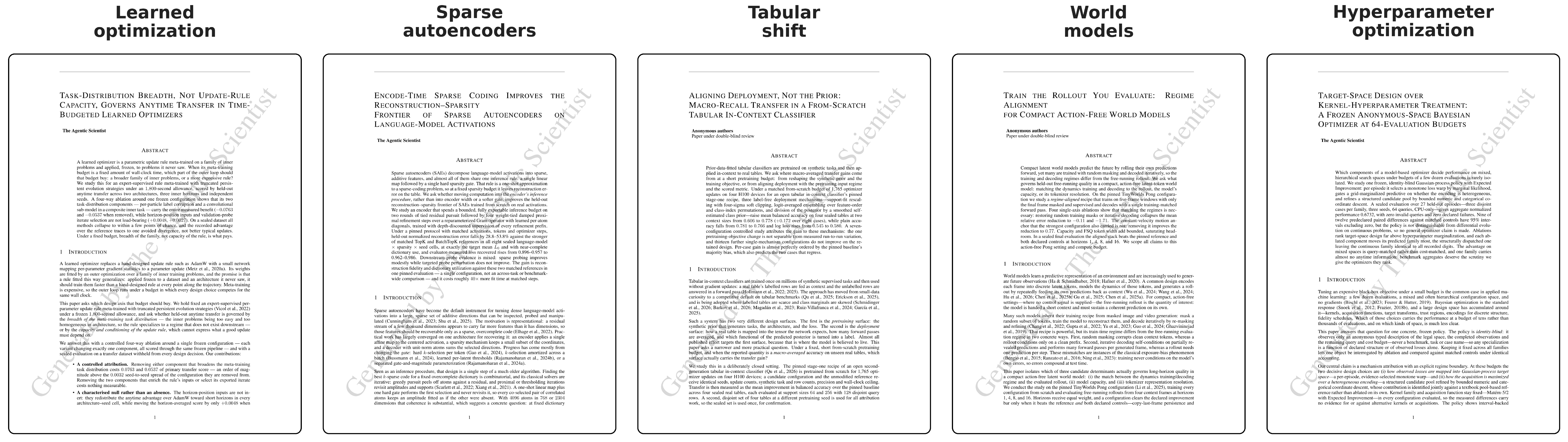}
    \caption{First pages of the five highlighted task manuscripts, selected by highest AI reviewer score and final task performance. The generated papers are watermarked by \agentscientist.}
    \label{fig:highlighted_task_papers}
\end{figure}

To illustrate the discoveries behind these aggregate scores, we highlight the manuscript with the highest AI reviewer score for each task, breaking ties by final task performance (\Cref{fig:highlighted_task_papers}).

\paragraph{\textit{Learned optimization.}}
Claude Code with Opus~5 broadens meta-training from clean dense-network tasks to a composite task that averages dense and convolutional network losses, with per-particle label corruption. It also adds horizon conditioning and probe-based checkpoint selection. Ablations attribute the development gains primarily to task diversity and label corruption, with little benefit from the latter additions. On sealed SVHN, performance remains near chance, and the small advantage over the reference mainly reflects one avoided divergence rather than better typical updates.

\paragraph{\textit{Sparse autoencoders.}}
Pi with Opus~5 replaces a one-shot TopK encoder with two residual-pursuit rounds followed by four learned sparse-code refinement steps, while preserving the sparsity budget. Intermediate reconstruction losses and changes to dictionary optimization accompany this richer encoder. Against matched TopK and BatchTopK references, held-out reconstruction error falls by 28.8--53.8\%, but fitting takes approximately $40\times$ longer. A targeted feature-perturbation metric worsens, so the reconstruction gains do not establish improved interpretability.

\paragraph{\textit{Tabular distribution shift.}}
OpenCode with Opus~5 augments a TabICLv2 stage-one baseline trained from scratch with support-fitted rescaling and clipping, feature-permutation and class-rotation ensembling, and posterior correction using a class prior estimated from query predictions. Class-balanced training loss is also implemented, but ablations find no measurable benefit from that change. The deployment recipe raises sealed mean balanced accuracy from 0.606 to 0.778, while ordinary accuracy falls from 0.781 to 0.766, exposing a trade-off between macro-recall and aggregate accuracy.

\paragraph{\textit{World models.}}
OpenCode with DeepSeek Flash aligns TinyWorlds training and decoding with next-frame prediction: four clean context frames condition a fully masked fifth frame, and one prediction pass replaces iterative MaskGIT decoding. The selected model also enlarges the dynamics transformer and quantizer codebook. On sealed Pong, its mean relative rollout-error reduction against the reference is 49\%. Ablations support the training and decoding changes, while removing the selected model's constant-velocity latent initializer improves development results. Evidence remains limited to action-free Pong and one seed per evaluation setting.

\paragraph{\textit{Hyperparameter optimization.}}
Claude Code with Opus~5 replaces the incumbent-centred local-search baseline with a Gaussian-process optimizer using Expected Improvement. It selects a loss transformation per problem, averages over kernel hyperparameters for mixed search spaces, and refines acquisition candidates through trust-region proposals and coordinate search. Ablations identify target transformation as the largest measured contributor. Gains are strongest on mixed and hierarchical spaces; the comparisons match query counts rather than evaluation cost and do not establish superiority over differential evolution on continuous tasks.

\subsection{Learned Covariance Matrix Adaptation Additional Results}

\begin{table}[H]
    \centering
    \small
    \caption{Mean terminal best-so-far return $\pm$ one standard error over 16 shared seeds after 2,000 generations. Ant is the search-validation reference environment and uses new transfer seeds and the reduced transfer-policy interface; the other seven environments were unseen during agentic search. Bold denotes the highest mean within an environment.}
    \label{tab:metaevo_transfer}
    \begin{tabular}{@{}lrrrr@{}}
\toprule
Environment & CMA-ES & \shortstack{GPT-5.6-Sol \\+ Codex} & \shortstack{Opus-5 + Claude Code \\(CMA-Anchored LearnedES)} & \shortstack{DeepSeek V4 Flash \\+ Pi} \\
\midrule
Ant & $3328.12 \pm 28.41$ & $3053.53 \pm 39.02$ & $\mathbf{3509.26 \pm 59.09}$ & $3373.30 \pm 48.44$ \\
HalfCheetah & $3704.73 \pm 19.53$ & $3435.84 \pm 95.04$ & $3448.89 \pm 111.84$ & $\mathbf{3768.87 \pm 138.32}$ \\
Hopper & $\mathbf{1464.10 \pm 36.34}$ & $923.09 \pm 44.29$ & $860.66 \pm 39.68$ & $996.09 \pm 21.72$ \\
Humanoid & $4368.72 \pm 105.59$ & $4488.64 \pm 165.19$ & $\mathbf{4682.69 \pm 187.04}$ & $3400.08 \pm 187.88$ \\
Reacher & $-17.03 \pm 0.94$ & $\mathbf{-3.32 \pm 0.30}$ & $-9.23 \pm 0.42$ & $-21.23 \pm 1.89$ \\
Walker2d & $1699.83 \pm 43.04$ & $\mathbf{2094.39 \pm 53.41}$ & $1769.62 \pm 43.21$ & $1821.07 \pm 41.05$ \\
Fetch & $9.888 \pm 0.109$ & $8.074 \pm 0.179$ & $\mathbf{9.899 \pm 0.129}$ & $8.792 \pm 0.239$ \\
UR5e & $\mathbf{16.196 \pm 0.233}$ & $12.025 \pm 0.358$ & $16.042 \pm 0.179$ & $15.972 \pm 0.134$ \\
\bottomrule
\end{tabular}

\end{table}

\subsection{Error-Diffusion Additional Results}

\subsubsection*{Bias offsets and Dale-style sign preservation}
Because biases are unconstrained and may become negative, we ask whether bias offsets are large enough to functionally dominate the synaptic preactivation terms, which would weaken the interpretation of the learned activity as primarily shaped by Dale-style synaptic interactions.

We decomposed each preactivation as
\[
z = z_{\mathrm{syn}} + b,
\]
and measured \(\mathbb{E}|b| / \mathbb{E}|z_{\mathrm{syn}}|\), the fraction of
elements for which \(|b| > |z_{\mathrm{syn}}|\), and the fraction for which
adding \(b\) flips the preactivation sign. Across all 5000 validation examples,
bias effects were concentrated almost entirely in the first convolutional layer: 
\texttt{conv1\_p} 
had a bias/synaptic ratio of 0.390, with biases dominating $30.5\%$ of preactivation elements and flipping $14.9\%$ of signs; \texttt{conv1\_n} showed a similar ratio of 0.379, with $31.2\%$ bias-dominant elements and a $15.2\%$ sign-flip rate. 
In deeper layers, bias effects were negligible: conv2
and conv3 had bias/synaptic ratios at most 0.0105 and sign-flip rates below
0.004, while all fully connected layers had ratios at most 0.0032 and sign-flip
rates below 0.001.

These results suggest that negative biases primarily act as threshold shifts, especially in the first convolutional layer, rather than as a mechanism for violating Dale-style synaptic signs. 
The formal pathway-level excitatory and inhibitory structure remains intact, while bias offsets provide a limited stream-specific adjustment to activation thresholds.

\newpage
\subsubsection*{Rubric ablations}

\begin{table}[h]
\centering
\resizebox{\textwidth}{!}{%
\begin{tabularx}{\textwidth}{l l X}
\hline
Rubric strictness & Key ED constraints & Observed effect \\
\hline

Loose
&
\begin{tabular}[t]{@{}l@{}}
No explicit wording to prohibit backpropagation
\end{tabular}
&
Best validation accuracy $0.9694$ but failed to implement ED \\

Intermediate
&
\begin{tabular}[t]{@{}l@{}}
Explicit wording to prohibit backprop \\
\end{tabular}
&
Best validation accuracy $0.976$. Implemented ED correctly \\

Strict
&
\begin{tabular}[t]{@{}l@{}}
No backprop \\
No $W^\top$ backward transport \\
Explicit local update with one global error sign \\
Detailed polarity invariance
\end{tabular}
&
Constrained search toward global/mean error broadcast ED; best validation accuracy $0.9708$ \\

\hline
\end{tabularx}%
}
\caption{Rubric ablation design for MNIST Error Diffusion runs. The experiment varied the strictness and wording of the method-validity rubric while keeping the MNIST task, fixed split, search budget, and high-level exclusions constant.}
\end{table}

\newpage
\section{Highlighted Papers (Co-) Generated by \agentscientist}

\subsection{Doc-to-LoRA Query-Conditioned Selection Paper \& AI Review}\label{appendix:doc_to_lora_paper}
\embedpdfpaperfour{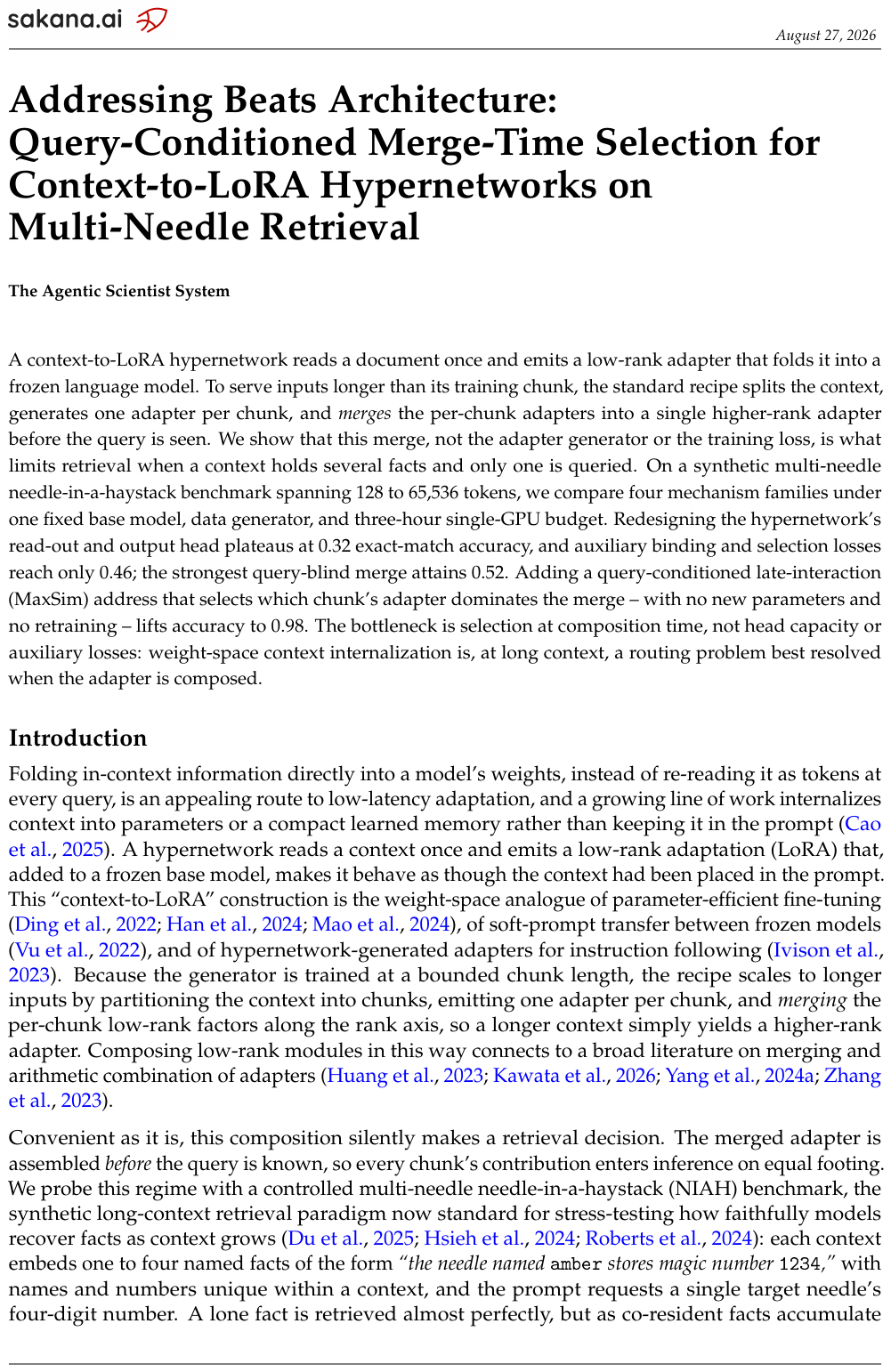}{Doc-to-LoRA Query-Conditioned Selection Paper}

\begin{reviewbox}
\appendixfileinput{assets/reviews/doc_to_lora_review.txt}
\end{reviewbox}

\newpage
\subsection{Learned Covariance Matrix Adaptation Paper \& AI Review}\label{appendix:metaevo_paper}
\embedpdfpaperfour{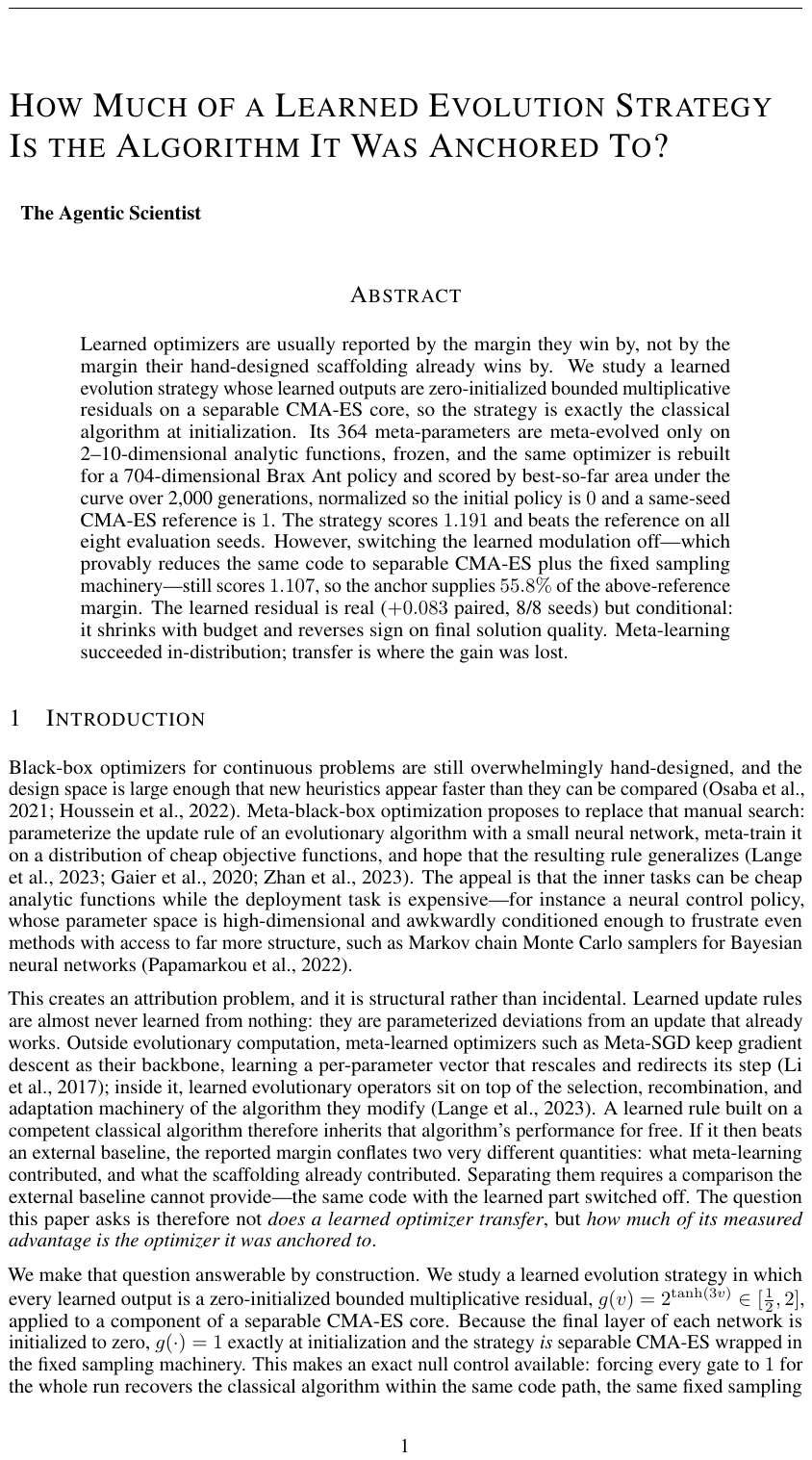}{Learned CMA-ES Paper}

\begin{reviewbox}
\appendixfileinput{assets/reviews/metaevo_review.txt}
\end{reviewbox}

\newpage
\subsection{Error Diffusion ALIFE Paper}
\label{paper:alife2026_camera_ready}

\embedpdfpaperfour{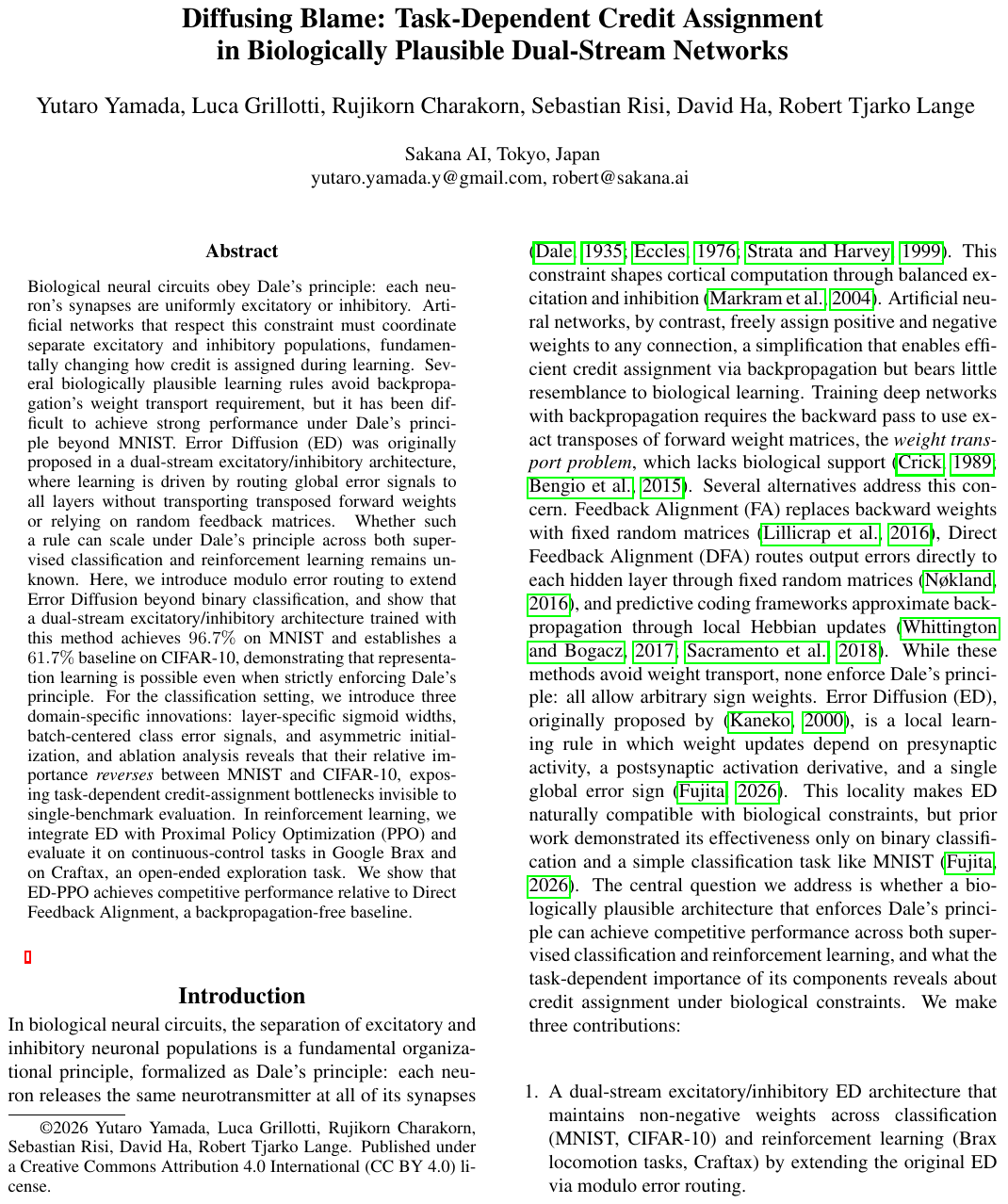}{Error Diffusion ALIFE Paper}

\newpage

\section{Error Diffusion ALIFE Paper Details}
\label{app:error_diffusion_details}
\subsection{Error Diffusion ALIFE Paper: Agent Draft to Submission Diff}
\label{paper:alife2026_diff_agentic_submitted}
\begin{figure}[htbp]
    \centering
    \begin{subfigure}[t]{\linewidth}
        \centering
        \includegraphics[width=\linewidth]{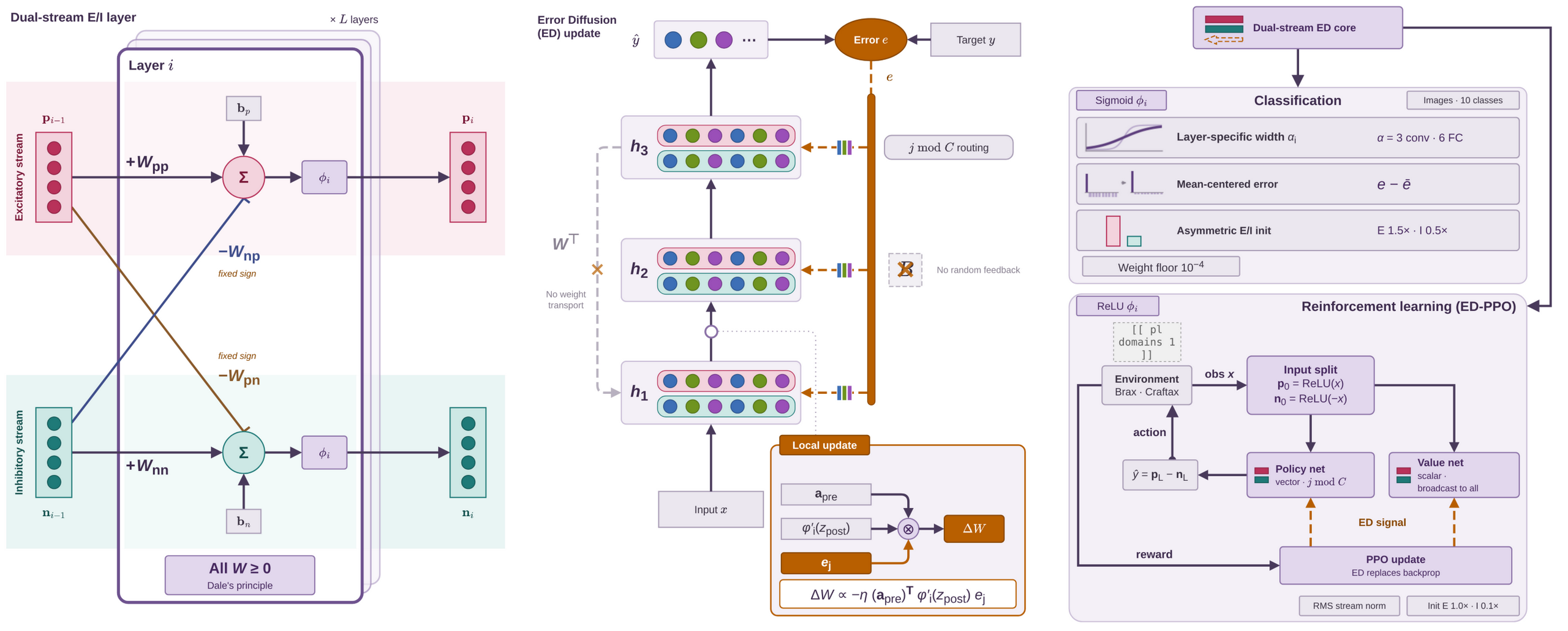}
        \caption{\centering Editable \texttt{draw.io} diagram.}
        \vspace{1.0em}
        \label{fig:autoillustrate_alife_editable}
    \end{subfigure}
    \begin{subfigure}[t]{\linewidth}
        \centering
        \includegraphics[width=\linewidth]{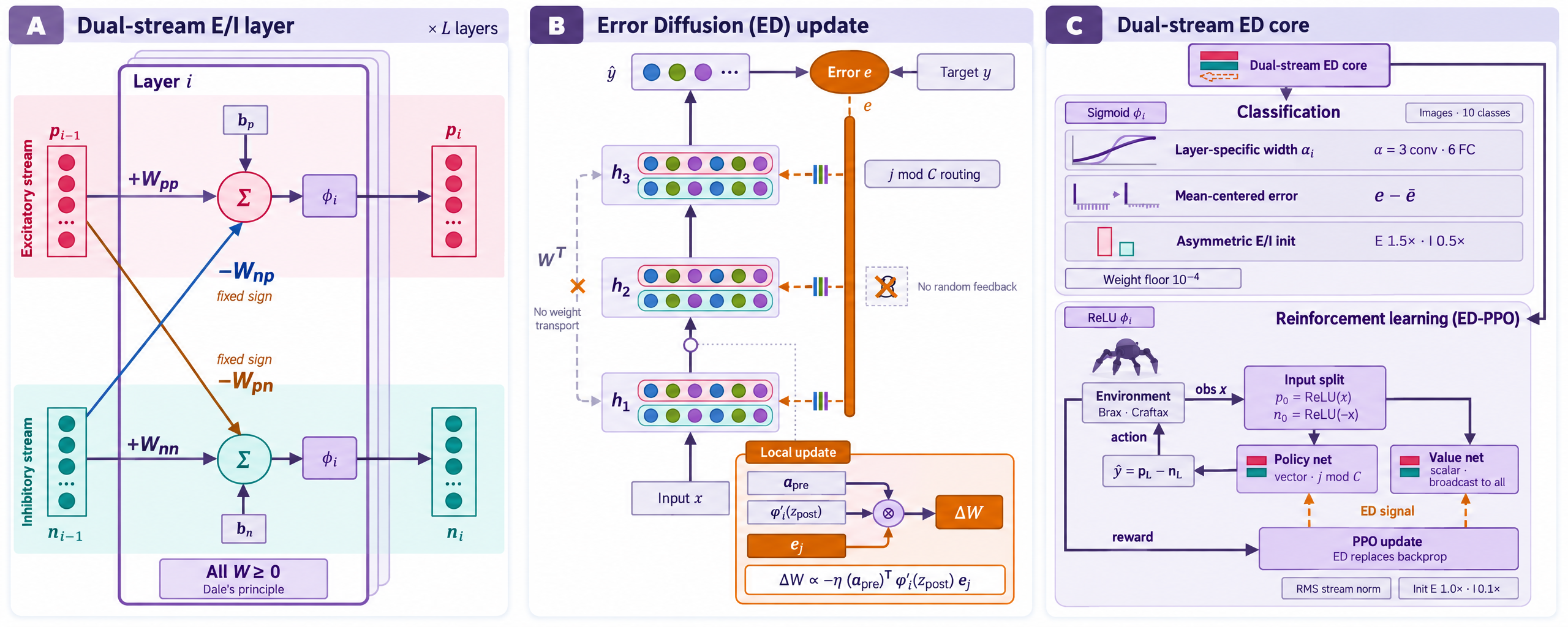}
        \caption{\centering Enhanced figure derived from the editable diagram.}
        \label{fig:autoillustrate_alife_enhanced}
    \end{subfigure}
    \caption{\autoillustrate applied to the agent draft of the Error Diffusion ALIFE paper, before human edits: the editable diagram \textbf{(a)} and its corresponding enhanced figure \textbf{(b)}. The figure in the submitted and camera-ready versions of the ALIFE paper was instead generated with Nano Banana Pro as a non-editable raster image; the human authors kept its correct right panel and manually redrew the left and middle panels to correct their inaccuracies.
    The prompt given to \autoillustrate was: ``Overview of the dual-stream Error Diffusion framework. \textbf{Left:} The excitatory/inhibitory architecture maintains separate streams ($\mathbf{p}$, $\mathbf{n}$) with four non-negative weight matrices per layer, enforcing Dale's principle structurally. \textbf{Center:} The Error Diffusion update broadcasts output error directly to hidden layers without weight transport or random feedback matrices. \textbf{Right:} The shared architecture is applied to classification (with domain-specific sigmoid widths, mean-centered error, and asymmetric initialization) and reinforcement learning (with ReLU activations and PPO integration).''}
    \label{fig:autoillustrate_alife_example}
\end{figure}

As this paper was produced with a development version of \agentscientist, its conceptual figure was originally generated with Nano Banana Pro~\citep{google2025nanobananapro}, rather than with \autoillustrate.
Since Nano Banana Pro outputs a raster image without an editable source, the human authors could not correct its factual inaccuracies in place: before submission, they kept the right panel, which was correct, and manually redrew the left and middle panels.
The current version of \agentscientist uses \autoillustrate by default for conceptual figures (\Cref{sec:paper_writing,sec:autoillustrate}), which produces more faithful figures than the Nano Banana Pro-based PaperBanana pipeline~\citep{zhu2026paperbanana} (\Cref{sec:autoillustrate_results}).
For reference, we applied the current version of \autoillustrate to the agent draft of the ALIFE paper, before any human edits and adjustments.
\Cref{fig:autoillustrate_alife_example} shows the resulting editable \texttt{draw.io} diagram and the enhanced figure derived from it.

\newpage

\embedpdfpaperfour{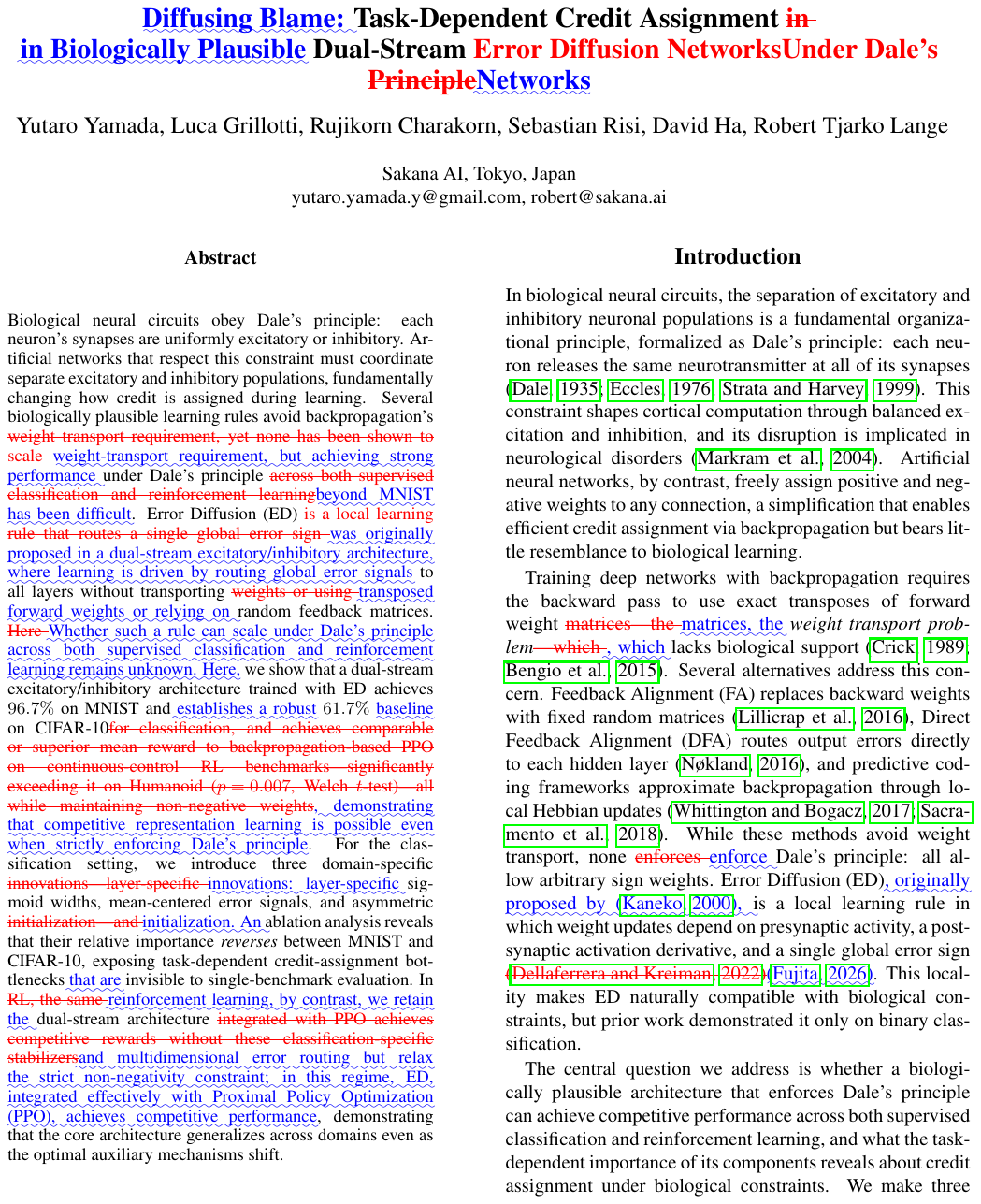}{Error Diffusion ALIFE Paper: agent draft to submission diff}

\newpage
\subsection{Error Diffusion ALIFE Paper: Submission to Camera-Ready Diff}
\label{paper:alife2026_diff_submitted_camera_ready}
\embedpdfpaperfour{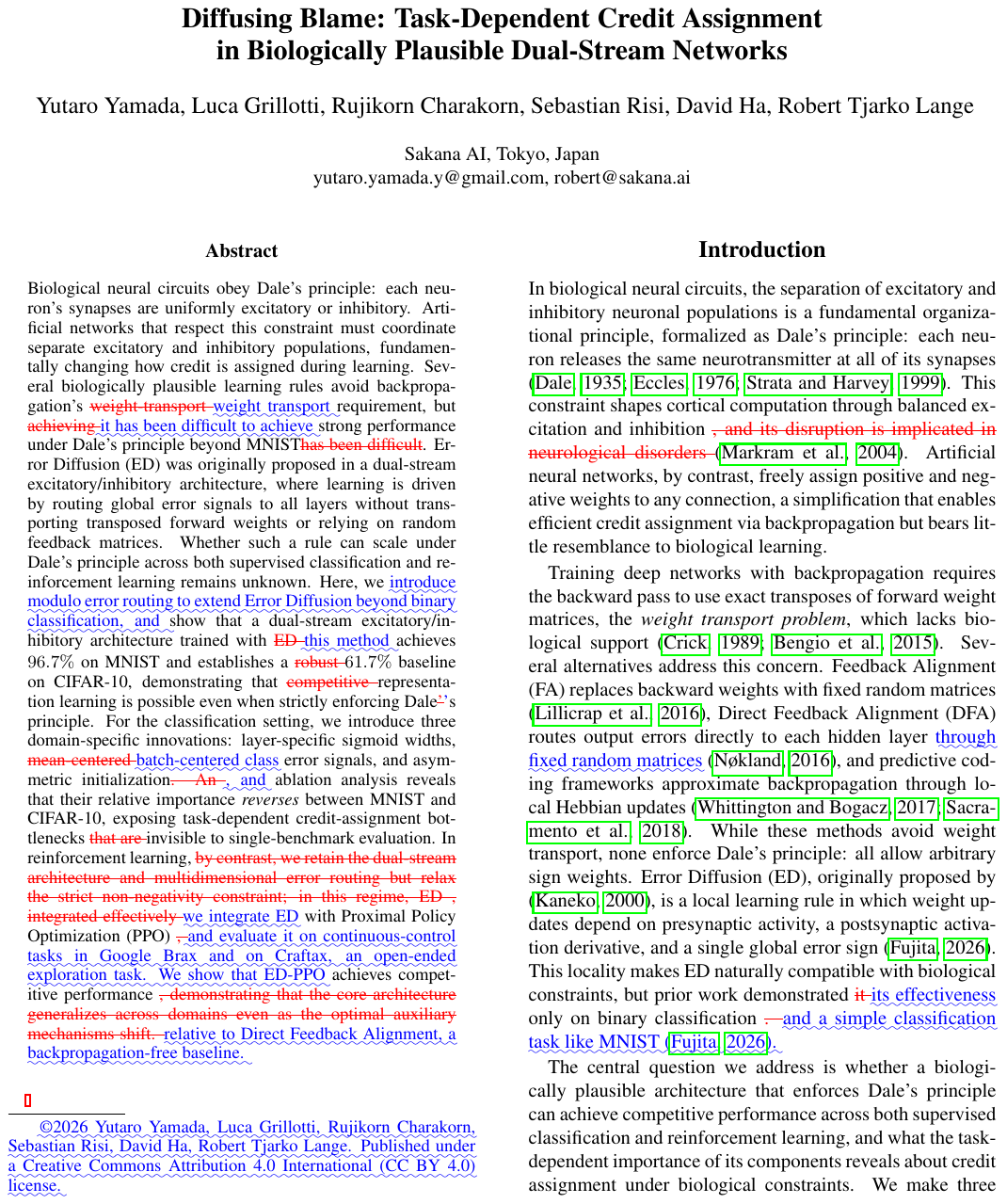}{Error Diffusion ALIFE Paper: submission to camera-ready diff}

\newpage
\section{\autoillustrate: Supplementary Material}
\label{app:autoillustrate}

\begin{figure}[H]
    \centering
    \includegraphics[width=\linewidth]{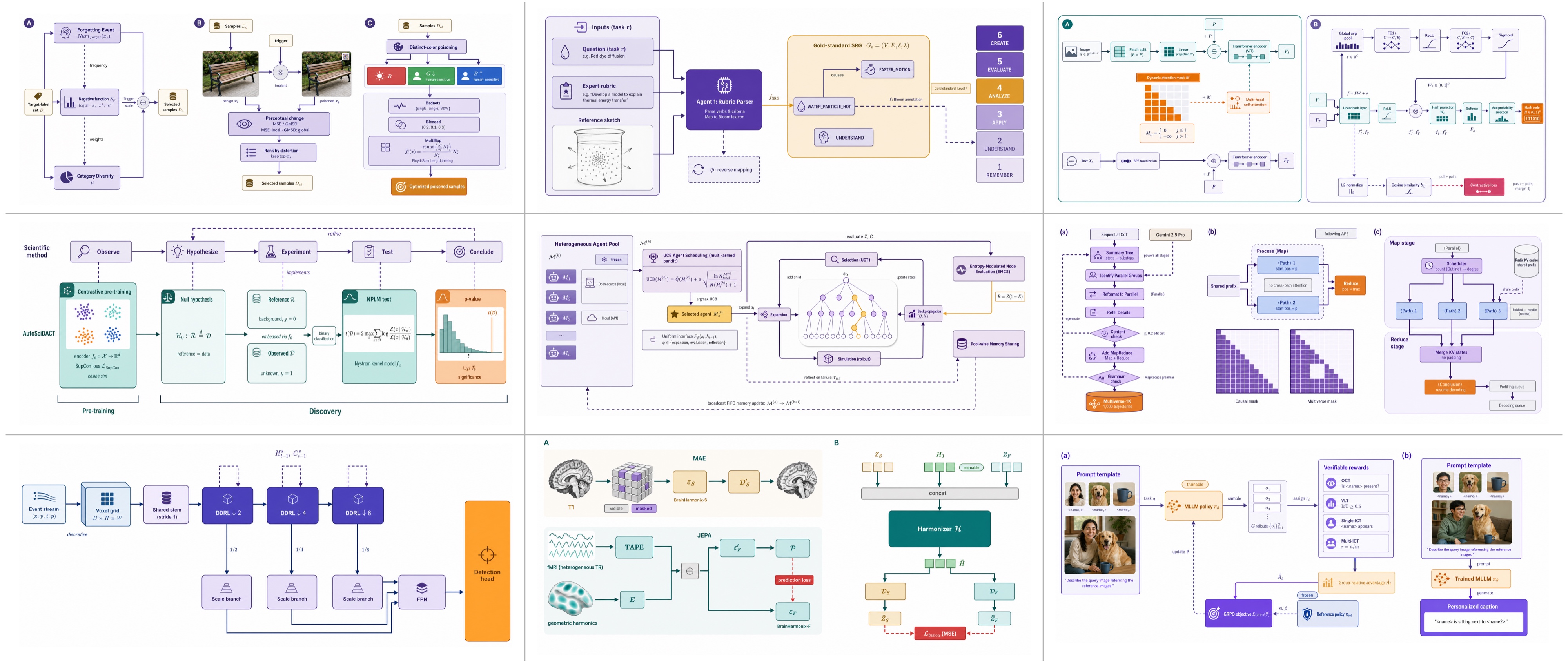}
    \caption{Examples of figures generated by \autoillustrate on PaperBananaBench~\citep{zhu2026paperbanana}.}
    \label{fig:autoillustrate_mosaic}
\end{figure}

\newpage
\subsection{Hyperparameters}

We use a fixed benchmark configuration across the \autoillustrate variants, changing only the coding-agent model, critic availability, or effort level indicated by each variant.
\Cref{tab:bananabench-hyperparams} summarizes the shared settings.
The original PaperBanana baseline relies on \texttt{gemini-3.1-pro-preview} and \texttt{gemini-3-pro-image-preview} (Nano Banana Pro).
For the PaperBanana ``Claude + GPT~Image~2'' variant, similarly to our pipeline, we use Claude Opus 4.8 for the Stylist, GPT-5.5 for the Critic, and GPT Image 2 for the Visualizer.
We retain Gemini for the Retriever because its long inputs exceed Claude's one-million-token context limit, and retain PaperBanana's original Planner because \autoillustrate has no corresponding component that accepts images as input.

\begin{table}[htbp]
    \centering
    \small
    \caption{Hyperparameters for the PaperBananaBench evaluation.}
    \label{tab:bananabench-hyperparams}
    \begin{tabular}{@{}p{0.475\linewidth} p{0.475\linewidth}@{}}
    \toprule
    \textbf{Hyperparameter} & \textbf{Value} \\
    \midrule
    Codebase version & \texttt{autoillustrate} v0.2.0 \\
    \addlinespace
  Coding agent & \texttt{Claude Code}; \texttt{Codex} for the Codex variant \\
  Agent model & \texttt{claude-opus-4-8}; \texttt{gpt-5.6-sol} for the Codex variant \\
    Effort level & high \\
    \addlinespace
    Panel inspects (Phase 3) & 3: one inspect $+$ up to two repair passes per panel \\
    Enhancement attempts & 4: Phase 5 enhance $+$ 3 verify $+$ 2 repair passes \\
    \addlinespace
    Audit VLM (Phase 3 inspect, Phase 6 verify Stage A) & \texttt{gpt-5.5} \\
    Unit-test backend (Phase 6 Stage B) & \texttt{gpt-5.5} \\
    Enhancement model (Phase 5) & \texttt{gpt-image-2}, quality \texttt{high} \\
    \addlinespace
    Judge model & \texttt{gpt-6-astra}, high reasoning effort (\Cref{tab:autoillustrate_results}); \texttt{gemini-3.1-pro-preview} (\Cref{tab:autoillustrate_results_gemini_judge}) \\
    \bottomrule
    \end{tabular}
\end{table}

\subsection{Results with Gemini-3.1-Pro as VLM judge}
\label{app:autoillustrate_gemini_judge}

\begin{figure}[htbp]
    \centering
    \includegraphics[width=0.85\linewidth]{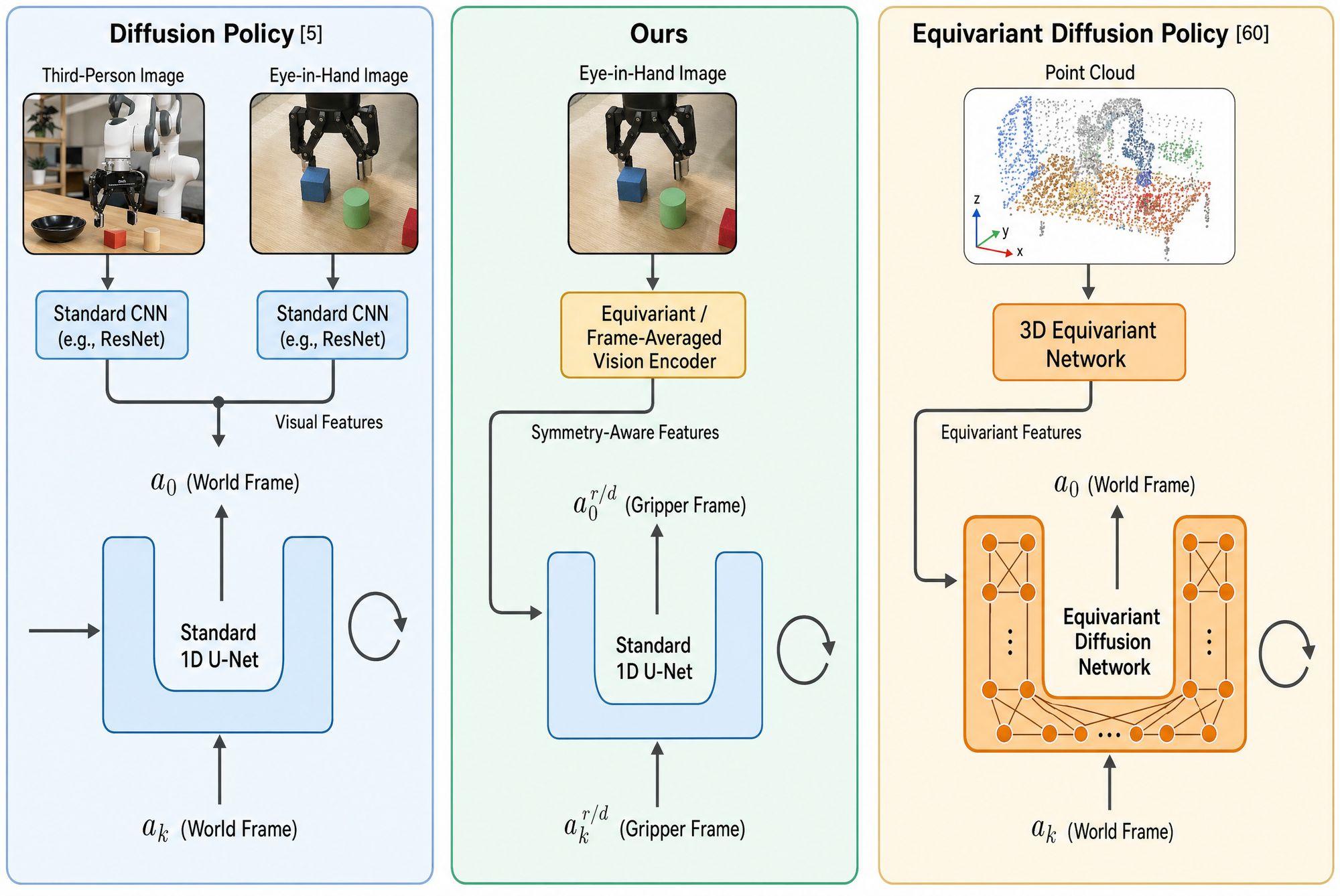}
    \caption{Example PaperBanana (Claude Opus 4.8 + GPT Image 2) diagram on which the two VLM judges disagree on Faithfulness. \\
    \textbf{Gemini-3.1-Pro rationale:} ``Accurately captures the structural differences and integrates specific text details like delta/relative trajectories ($a^{r/d}$) and frame-averaged encoders. It strictly aligns with the caption's visual simplification. No veto errors.'' \\
    \textbf{GPT-6-Astra rationale:} ``Correctly depicts the proposed method's encoder/backbone distinction and action frames. However, in the original Diffusion Policy panel, the visual-feature connection terminates above the clean-action label rather than entering the U-Net. The separate, unlabeled arrow entering the U-Net is disconnected from those features. This omits the essential visual-conditioning connection and constitutes a logical-flow veto, not merely simplification.''}
    \label{fig:autoillustrate_judge_disagreement}
\end{figure}

\Cref{tab:autoillustrate_results_gemini_judge} reports the PaperBananaBench scores of the same generated diagrams when judged by Gemini-3.1-Pro instead of GPT-6-Astra (\Cref{tab:autoillustrate_results}).
Gemini-3.1-Pro awards higher scores than GPT-6-Astra across all methods.
The ablation trends within \autoillustrate are preserved, and \autoillustrate still outperforms the Gemini-based PaperBanana on Overall score (50.0 vs.\ 46.7).
However, PaperBanana with Claude Opus 4.8 and GPT Image 2 obtains the highest Overall score (56.5) under this judge, mainly due to higher Faithfulness (53.1 vs.\ 43.7), whereas GPT-6-Astra ranks it last (18.3).
This discrepancy stems from PaperBanana's raster diagrams, which frequently contain small image-generation errors, such as disconnected or dangling arrows, that GPT-6-Astra treats as faithfulness vetoes while Gemini-3.1-Pro overlooks them; \Cref{fig:autoillustrate_judge_disagreement} shows an example diagram together with the Faithfulness rationales of both judges.
A likely cause of these errors is that PaperBanana renders complex, highly detailed diagrams with a prompt-only image generation model: the model receives text alone; on the other hand, the image generation model used in \autoillustrate takes the rendered editable \texttt{draw.io} diagram as an input image.

\begin{table}[htbp]
    \scriptsize

    \centering
    \caption{Mean scores on PaperBananaBench, using Gemini-3.1-Pro as VLM judge.}
    \setlength{\tabcolsep}{4pt}
    \setlength{\defaultaddspace}{0.2em}
    \begin{tabular}{l l c c c c c|c}
    \toprule
    \multirow{2}{*}{Method} & \multirow{2}{*}{Variant} & \multirow{2}{*}{Faithfulness} & \multirow{2}{*}{Conciseness} & \multirow{2}{*}{Readability} & \multirow{2}{*}{Aesthetic} & \multirow{2}{*}{Overall} & Editable \\
    & & & & & & & source \\
    \midrule
    \addlinespace
    \multirow[t]{4}{*}{\autoillustrate}          & Claude & 43.7          & 45.7          & 45.0          & 75.2          & 50.0          & \cmark \\
                                             & + no critic & 42.8          & 41.6          & 45.2          & 75.3          & 46.6          & \cmark \\
                                             & \enspace + low effort & 37.0          & 43.2          & 46.2          & 74.8          & 42.1          & \cmark \\
                                             & Codex   & 42.3          & 39.9          & 31.0          & 49.8          & 34.9          & \cmark \\
    \addlinespace
    \multirow[t]{2}{*}{PaperBanana}          & Gemini & 36.0          & \textbf{57.0} & 48.1          & 74.7          & 46.7          & \xmark \\
                                             & Claude + GPT Image 2 & \textbf{53.1} & 30.0          & \textbf{51.5} & \textbf{79.5} & \textbf{56.5} & \xmark \\
    \midrule
    Human                                    & --     & 50.0          & 50.0          & 50.0          & 50.0          & 50.0          & \cmark \\

    \bottomrule
    \end{tabular}
    \label{tab:autoillustrate_results_gemini_judge}
\end{table}

\subsection{Limitations}
\label{sec:autoillustrate-limitations}

\Cref{tab:autoillustrate_failures} shows representative failure cases of \autoillustrate, together with their causes.
Another recurring issue is that the enhancement model tends to turn unidirectional arrows into bidirectional ones: for instance, in \Cref{fig:autoillustrate_alife_enhanced}, the ``No weight transport'' arrow of panel B acquires a second arrowhead that is absent from the editable diagram (\Cref{fig:autoillustrate_alife_editable}).
The enhancement model may also struggle to render uniform background colors for boxes, as can be seen in the same figure.

\begin{table}[H]
\centering
\captionsetup{justification=centering}
\caption{Example failure cases of AutoIllustrate.}
\label{tab:autoillustrate_failures}
\begin{tabular}{|>{\centering\arraybackslash}m{0.72\textwidth}|>{\raggedright\arraybackslash}m{0.23\textwidth}|}
\hline
\textbf{AutoIllustrate Diagrams} & \multicolumn{1}{c|}{\textbf{Cause of Failure}} \\
\hline
\adjustbox{max width=\linewidth, max height=0.18\textheight, center, margin=2pt}{%
  \includegraphics{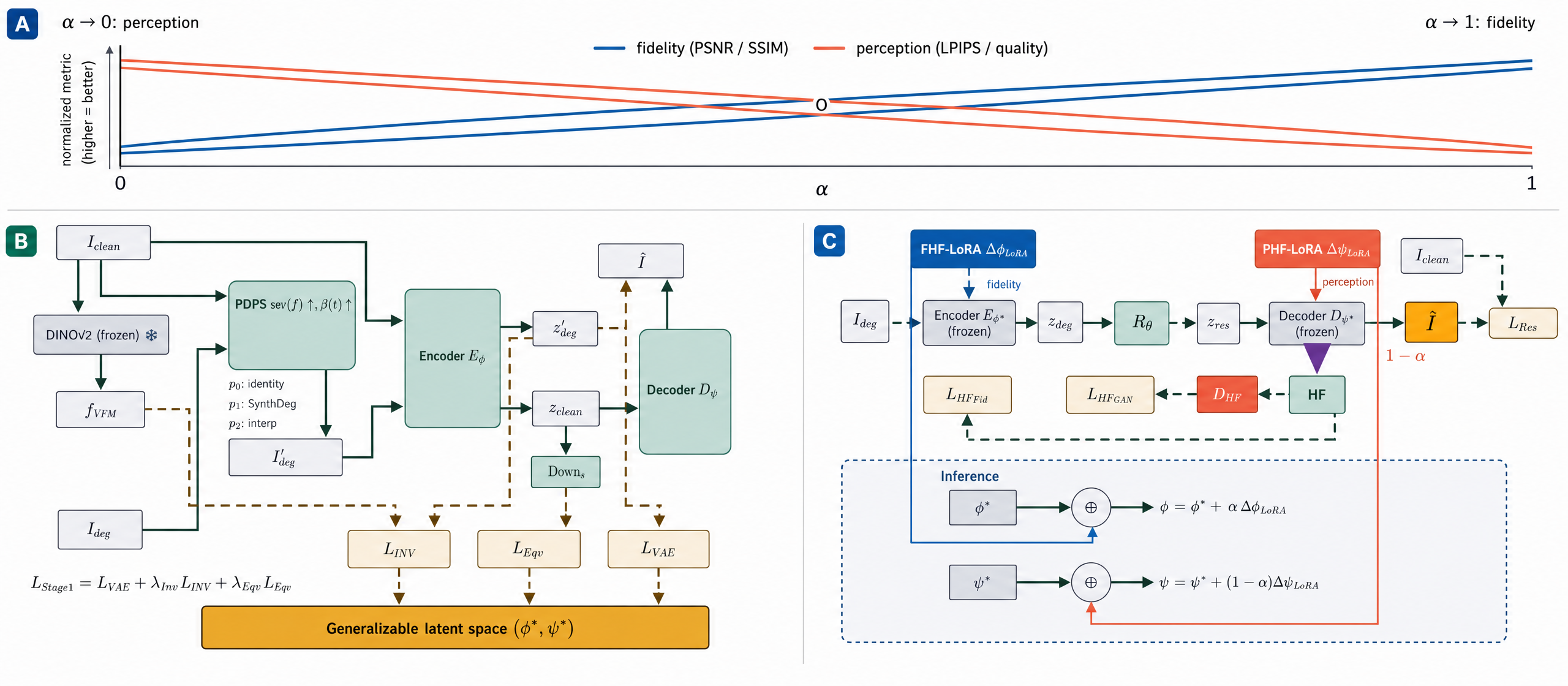}%
}
&
\textbf{Edge attached to the wrong source.} The equivariance loss $\mathcal{L}_{Eqv}$ is fed the downsampled latent directly from $Down_s$, bypassing the decoder $D_\psi$, whereas the method defines the loss on the decoded output
$D_\psi(z_{down})$. \\
\hline
\adjustbox{max width=\linewidth, max height=0.18\textheight, center, margin=2pt}{%
  \includegraphics{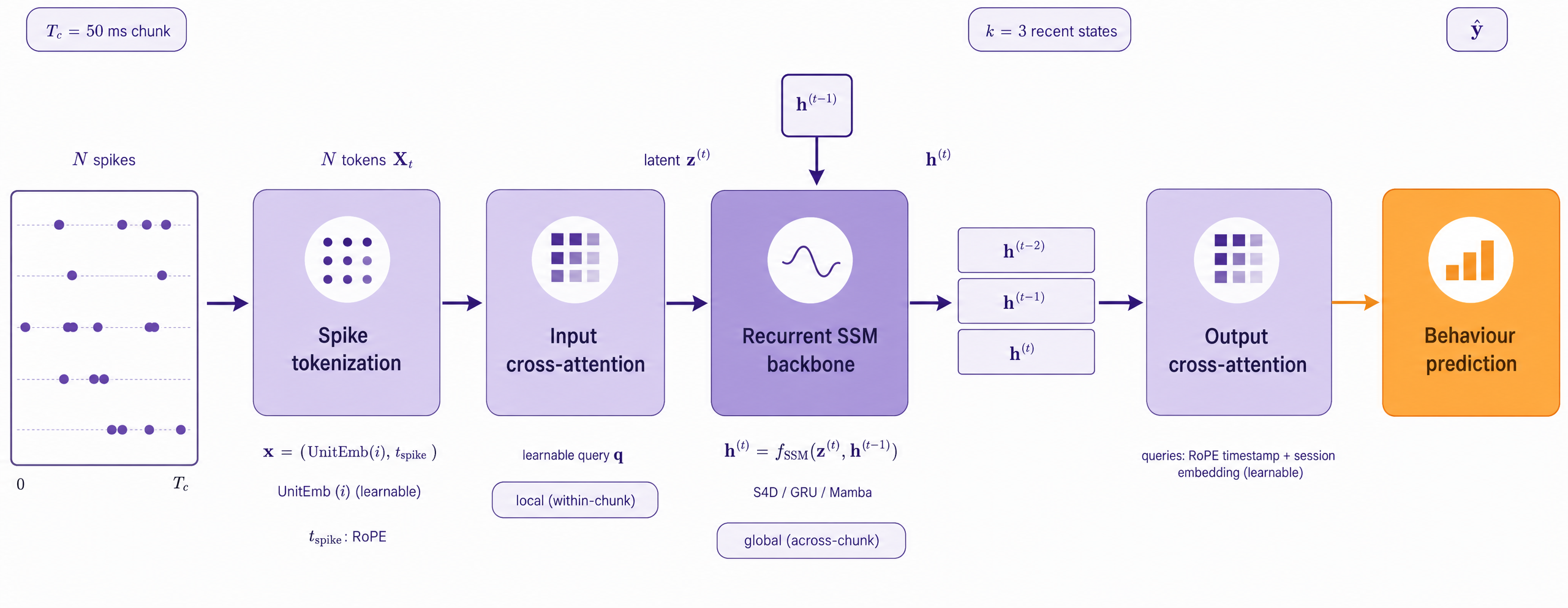}%
}
&
\textbf{Space-inefficient layout.} Disconnected annotations float above the main pipeline, yielding a non-rectangular bounding box with large dead zones that waste vertical space under standard publication formatting. \\
\hline
\adjustbox{max width=\linewidth, max height=0.18\textheight, center, margin=2pt}{%
  \includegraphics{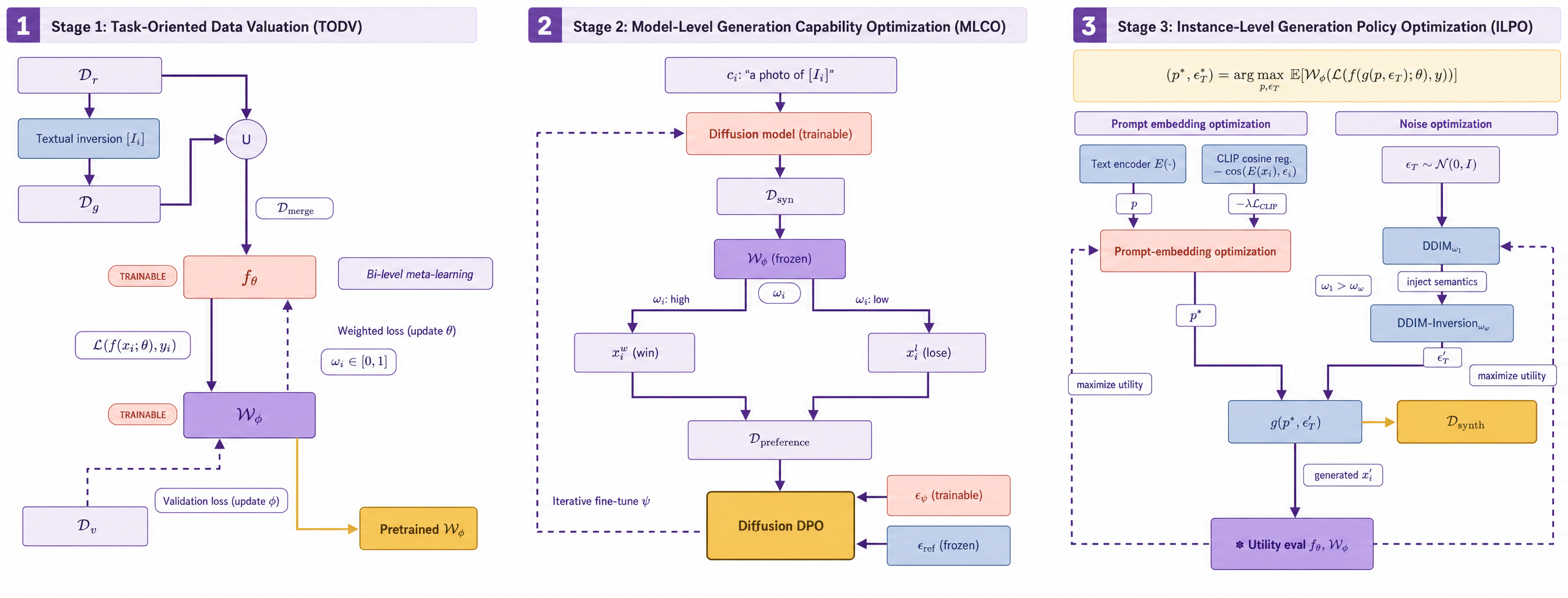}%
}
&
\textbf{Insufficient visual abstraction.} Raw equations are transcribed into the figure rather than distilled, most visibly the full argmax objective placed in a box at the top of Stage 3, and modules are labelled with bare symbols ($f_\theta$, $W_\phi$) instead of their function, reducing the diagram to a box-ified restatement of the text. \\
\hline
\end{tabular}
\end{table}

\newpage
\section{Additional Details on \agentscientist}
\label{app:agentic-scientist-details}

This section provides additional details about the \agentscientist system, including the prompts used by the ideation, execution, ablation-study, and task-drafting agents.

\subsection{Bibliography Retrieval}
\label{app:bibliography-retrieval}

The paper-writing phase (\Cref{sec:paper_writing}) never lets the writer agent produce bibliography entries from memory.
Before drafting, a reference-planning agent reads the staged run artifacts and writes a small query plan (6--10 search queries by default) covering the central topic, the core method families, the comparator literatures a reviewer would expect, and the benchmark context; this agent does not retrieve anything itself.
A deterministic script then executes the plan against the Semantic Scholar Graph API~\citep{kinney2023semantic}: for each query it searches up to 25 candidate papers, ranks them by citation count, fetches the BibTeX entry served by the API, and keeps only candidates with a well-formed entry and citation key, deduplicating by paper identifier and key until a target of 8--15 references (configurable) is met.
If Semantic Scholar is unavailable, rate-limited, or returns too few references, the script falls back to OpenAlex~\citep{priem2022openalex}, resolving each work's BibTeX through DOI content negotiation and filling in author names from OpenAlex metadata; Google Scholar via the \texttt{scholarly} package is used as a last resort.
The script writes a \texttt{references.bib} file together with a provenance record listing, for every entry, the originating query, source URL, and retrieval timestamp.
The writer agent may only cite keys present in this file, and the validation checks of the repair loop reject any citation key or claim-to-source link that does not resolve to a retrieved entry.
After a first draft exists, a gap-analysis agent inspects the manuscript for under-cited passages (e.g., comparator families named in the related work without a citation), a post-draft planner turns these gaps into a few additional queries, the same retrieval script enriches the bibliography incrementally, and the writer revises only the flagged paragraphs.

\subsection{Ideation Agent}
\label{app:prompt-ideation}

\begin{promptbox}{Ideation Agent Prompt}
\appendixfileinput{sections/08_appendix/agsci_prompts/prompts/ideation_agent.txt}
\end{promptbox}

\subsection{Execution Agent}
\label{app:prompt-execution}

\begin{promptbox}{Execution Agent Prompt}
\appendixfileinput{sections/08_appendix/agsci_prompts/prompts/execution_agent.txt}
\end{promptbox}

\subsection{Ablation-Study Agent}
\label{app:prompt-ablation}

\begin{promptbox}{Ablation-Study Agent Prompt}
\appendixfileinput{sections/08_appendix/agsci_prompts/prompts/ablation_agent.txt}
\end{promptbox}

\subsection{Task-Drafting Agent}
\label{app:prompt-task-drafting}

\begin{promptbox}{Task-Drafting Agent Prompt}
\appendixfileinput{sections/08_appendix/agsci_prompts/prompts/task_drafting_agent.txt}
\end{promptbox}

\newpage
\section{Task Properties and Hyperparameters}

\subsection{Aggregate Sweep}
\begin{table}[htbp]
    \centering
    \small
    \caption{Harness and model configurations for the aggregate sweep. All runs use high reasoning effort.}
    \label{tab:aggregate_harness_models}
    \begin{tabularx}{\linewidth}{@{}ll>{\raggedright\arraybackslash}X@{}}
        \toprule
        Harness & Version & Assigned models \\
        \midrule
        Codex & 0.149.0 & GPT-6 Astra; GPT-5.6-Sol; GPT-5.5; GPT-5.4 \\
        Claude Code & 2.1.240 & Fable 5.1; Opus 5; Opus 4.7; Opus 4.6 \\
        Pi & 0.84.2 & GPT-5.6-Sol; Opus 5; DeepSeek V4 Flash \\
        OpenCode & 1.18.21 & GPT-5.6-Sol; Opus 5; DeepSeek V4 Flash \\
        \bottomrule
    \end{tabularx}
\end{table}

\subsubsection{Learned Optimization}
\begin{taskinstructionsbox}{Learned Optimization}
\appendixfileinput{sections/08_appendix/task_prompts/task_learned_optimizer.md}
\end{taskinstructionsbox}

\subsubsection{Hyperparameter Optimization}
\begin{taskinstructionsbox}{Hyperparameter Optimization}
\appendixfileinput{sections/08_appendix/task_prompts/hyperopt_task.md}
\end{taskinstructionsbox}

\subsubsection{World Modelling}
The boxed text records the initial research direction. The manual aggregate sweep uses a fixed TinyWorlds Pong protocol with four context frames and prediction horizons of 1, 4, 8, and 16 frames. Its error metric equally weights pixel and foreground reconstruction; observed actions and action controllability are not evaluated.
\begin{taskinstructionsbox}{World Modelling}
\appendixfileinput{sections/08_appendix/task_prompts/wm_task.md}
\end{taskinstructionsbox}

\subsubsection{Sparse Autoencoders}
\begin{taskinstructionsbox}{Sparse Autoencoders}
\appendixfileinput{sections/08_appendix/task_prompts/sae_task.md}
\end{taskinstructionsbox}

\subsubsection{Tabular Models}
\begin{taskinstructionsbox}{Tabular Models}
\appendixfileinput{sections/08_appendix/task_prompts/tds_task.md}
\end{taskinstructionsbox}

\newpage
\subsection{Details of Doc-to-LoRA Inductive Bias}

\begin{table}[H]
    \centering
    \small
    \caption{Doc-to-LoRA multi-needle NIAH agent-search and evaluation protocol.}
    \label{tab:doc_to_lora_params}
    \begin{tabularx}{\linewidth}{@{}ll>{\raggedright\arraybackslash}X@{}}
        \toprule
        Stage & Setting & Value \\
        \midrule
        Agent search & Idea budget & 40 ideation cycles \\
        & Model / harness & \texttt{gpt-5.6-sol} / Codex; \texttt{claude-opus-5} / Claude; \texttt{deepseek-v4-pro} / Pi \\
        & Hardware & One H100 GPU with 16 CPU cores \\
        & Training time limit & 3 hours; candidates above 3 hours are invalid \\
        \addlinespace
        Dataset & Training datasets & NIAH samples range from 32 to 256 tokens \\
        & Evaluation datasets & NIAH samples range from 32 tokens to 64K tokens \\
        & Example contract & Each context contains 1-4 needles \\
        & Benchmark scale & 512K training samples; 128 samples per length bucket in \{128, 8K, 16K, 32K, 64K\}, totaling 640 samples. \\
        \addlinespace
        Baseline model / training & Base model & Fixed \texttt{google/gemma-2-2b-it} target model and tokenizer \\
        & Adapter setup & \texttt{down\_proj} target module and LoRA rank 4\\
        \bottomrule
    \end{tabularx}
\end{table}

\begin{taskinstructionsbox}{Doc-to-LoRA Inductive Bias}
    \appendixfileinput{sections/08_appendix/task_prompts/doc_to_lora_multi_needle_task.md}
\end{taskinstructionsbox}

\newpage
\subsection{Details of Learned Covariance Matrix Adaptation}

\begin{table}[H]
    \centering
    \small
    \caption{MetaEvo training, search-validation, and transfer protocol. BBOB tasks meta-train neural weights; Ant subsequently supplies the score used by the agentic program search.}
    \label{tab:metaevo_protocol}
    \begin{tabularx}{\linewidth}{@{}ll>{\raggedright\arraybackslash}X@{}}
        \toprule
        Stage & Setting & Value \\
        \midrule
        Agent search & Idea iterations & 30 per agent \\
        & Models / agents & \texttt{gpt-5.6-sol} / Codex; \texttt{claude-opus-5} / Claude;\newline
            \texttt{deepseek-v4-flash} / Pi; \texttt{kimi-k3} / OpenCode \\
        & Valid evaluations & 28 Codex; 30 Claude;\newline 28 DeepSeek + Pi; 30 Kimi + OpenCode \\
        \addlinespace
        Outer meta-evolution & Optimizer & Full CMA-ES \\
        & Population / generations & 256 / 1,500 \\
        & Selected candidates' initial std. & 0.10 (Codex, Claude, Kimi); 0.01 (DeepSeek + Pi) \\
        \addlinespace
        BBOB inner loop & Tasks per outer generation & 128 newly sampled, shared across candidates \\
        & Population / generations & 16 / 50 \\
        & Dimension / initial std. & Uniform 2--10 / 1.0 \\
        & Functions & Sphere, Rosenbrock, Discus, Rastrigin, Schwefel, \\
        & & Bueche--Rastrigin, Attractive Sector, Weierstrass, \\
        & & Schaffers F7, Griewank--Rosenbrock \\
        & Noise & Additive, standard deviation uniform in $[0,0.1]$ \\
        & Aggregate loss & Per-task standardized loss; median across tasks \\
        \addlinespace
        Ant search-validation & Population / generations & 256 / 2,000 \\
        & Seeds / episode length & 8 fixed seeds / 500 steps \\
        & Initial standard deviation & 0.05 \\
        & Score & AUC normalized to initial policy and CMA-ES \\
        \addlinespace
        Brax transfer & Environments / seeds & 8 / 16 new shared seeds each \\
        & Population / generations & 256 / 2,000 \\
        & Initial std. / episode length & 0.05 / 500 steps \\
        \bottomrule
    \end{tabularx}
\end{table}

\newpage
\begin{taskinstructionsbox}{Learned Covariance Matrix Adaptation}
    \appendixfileinput{sections/08_appendix/task_prompts/metaevo_bbob_task.md}
\end{taskinstructionsbox}

\newpage
\subsection{Details of Error-Diffusion experiments}

\begin{table*}[h]
\centering
\small
\caption{Error-Diffusion task and evaluation parameters.}
\label{tab:error_diffusion_protocol}
\begin{tabularx}{\textwidth}{@{}lXXXX@{}}
\toprule
Setting & MNIST & CIFAR-10 & MuJoCo / Brax & Craftax \\
\midrule
Model / Harness
& Sonnet 4.6 / Claude Code 
& Opus 4.6 / Claude Code
& Composer 2.5 Fast / Cursor
& GPT-5.5 / Codex \\

Agent search budget
& 20 ideation cycles 
& 60 ideation cycles
& 60 ideation cycles
& 30 ideation cycles \\

Evaluator resources
& 1 GPU, 4 CPUs
& 1 GPU, 8 CPUs
& 1 GPU, 4 CPUs
& 1 GPU, 4 CPUs \\

Dataset / environment
& MNIST: 55k train / 5k validation / 10k test
& CIFAR-10: 45k train / 5k validation / 10k test
& Ant, HalfCheetah, Humanoid
& Craftax-Symbolic-v1 \\

Final training budget
& 250 epochs
& 250 epochs
& 400M requested steps per seed
& 1B requested steps per seed \\
\bottomrule
\end{tabularx}

\medskip
\begin{minipage}{\textwidth}
\footnotesize.
\end{minipage}
\end{table*}

\begin{taskinstructionsbox}{Error-Diffusion Classification}
    \appendixfileinput{sections/08_appendix/task_prompts/ed_mnist_task.md}
\end{taskinstructionsbox}

\begin{taskinstructionsbox}{Error-Diffusion Convolution}
    \appendixfileinput{sections/08_appendix/task_prompts/ed_conv_task.md}
\end{taskinstructionsbox}

\newpage
\subsection{Details of RL-based LLM fine-tuning}

\label{app:rlvr-details}

\begin{table}[H]
    \centering
    \small
    \caption{GRPO agent-search and evaluation protocol. A fixed Qwen2.5-Math-1.5B model is trained
on the MATH dataset with GRPO; final validation accuracy (mean reward) supplies the score used by
the agentic parallel investigation phase, and the best program of each run is re-scored on a held-out test set.}
    \label{tab:grpo_protocol}
    \begin{tabularx}{\linewidth}{@{}ll>{\raggedright\arraybackslash}X@{}}
        \toprule
        Stage & Setting & Value \\
        \midrule
        Agent search & Idea budget & 50 ideation cycles \\
        & Model / harness & \texttt{claude-opus-4-7} / Claude \\
        & Program evaluations & 1 per idea \\
        & Pruning & Every 20 cycles \\
        & Hardware & Two H100 GPUs with 16 CPU cores \\
        & Training time limit & 2.5 hours \\
        & Execution & Containerized (Docker, CUDA 12.8), Slurm-submitted \\
        \addlinespace
        Dataset & Base model / tokenizer & Fixed \texttt{Qwen/Qwen2.5-Math-1.5B} \\
        & Task & MATH reasoning \\
        & Train / validation split & 6{,}000 / 1{,}500 examples (stratified by subject and level) \\
        & Test split & MATH-500 problems (secret, outside the task mount) \\
        \bottomrule
    \end{tabularx}
\end{table}

\begin{taskinstructionsbox}{RL-based LLM fine-tuning}
    \appendixfileinput{sections/08_appendix/task_prompts/grpo_task.md}
\end{taskinstructionsbox}

\end{document}